\documentclass{CVM}
\usepackage{times}
\usepackage{color}

\usepackage{epsfig}
\usepackage{graphicx}
\usepackage{amsmath}
\usepackage{amssymb}
\usepackage{mathtools}
\usepackage{multirow}
\usepackage{float} 
\usepackage{subfigure}
\usepackage{caption}

\CVMsetup{
type      = {Research Article},
doi       = {s41095-0xx-xxxx-x},
title     = {Learning Accurate Template Matching with Differentiable Coarse-to-Fine Correspondence Refinement},
author    = {Zhirui Gao$^{1}$, Renjiao Yi$^{1}$, Zheng Qin$^{1}$, Yunfan Ye$^{1}$, Chenyang Zhu$^{1}$, and Kai Xu$^{1}$\cor{}\\
},
runauthor = {Gao et al.},
abstract  = {
Template matching is a fundamental task in computer vision and has been studied for decades. It plays an essential role in manufacturing industry for estimating the poses of different parts, facilitating downstream tasks such as robotic grasping. Existing methods fail when the template and source images have different modalities, cluttered backgrounds or weak textures. They also rarely consider geometric transformations via homographies, which commonly exist even for planar industrial parts. To tackle the challenges, we propose an accurate template matching method based on differentiable coarse-to-fine correspondence refinement. We use an edge-aware module to overcome the domain gap between the mask template and the grayscale image, allowing robust matching. An initial warp is estimated using coarse correspondences based on novel structure-aware information provided by transformers. This initial alignment is passed to a refinement network using references and aligned images to obtain sub-pixel level correspondences which are used to give the final geometric transformation.  Extensive evaluation shows that our method to be significantly better than state-of-the-art methods and baselines, providing good generalization ability and visually plausible results even on unseen real data.},
keywords  = {template matching, differentiable homography, structure-awareness, transformers},
copyright = {The Author(s)},
}

\begin{document}

\maketitle
    \begin{figure}[b] \vskip -4mm
    \small\renewcommand\arraystretch{1.3}
        \begin{tabular}{p{80.5mm}} \toprule\\ \end{tabular}
        \vskip -4.5mm \noindent \setlength{\tabcolsep}{1pt}
        \begin{tabular}{p{3.5mm}p{80mm}}
    $1\quad $ & College of Computer, National University of Defense Technology, Changsha 410073, China. E-mail: Z. Rui, gzrer2018@gmail.com; R. Yi, yirenjiao@nudt.edu.cn; Z. Qin, qinzheng12@nudt.edu.cn; Y.Ye yunfan951202@gmail.com; C. Zhu, zhuchenyang07@nudt.edu.cn; K. Xu, kevin.kai.xu@gmail.com\\
    
&\hspace{-5mm} Manuscript received: 2022-01-01; accepted: 2022-01-01\vspace{-2mm}
    \end{tabular} \vspace {-3mm}
    \end{figure}

\begin{figure}[!t]
\begin{center}
   \includegraphics[width=1.0\linewidth]{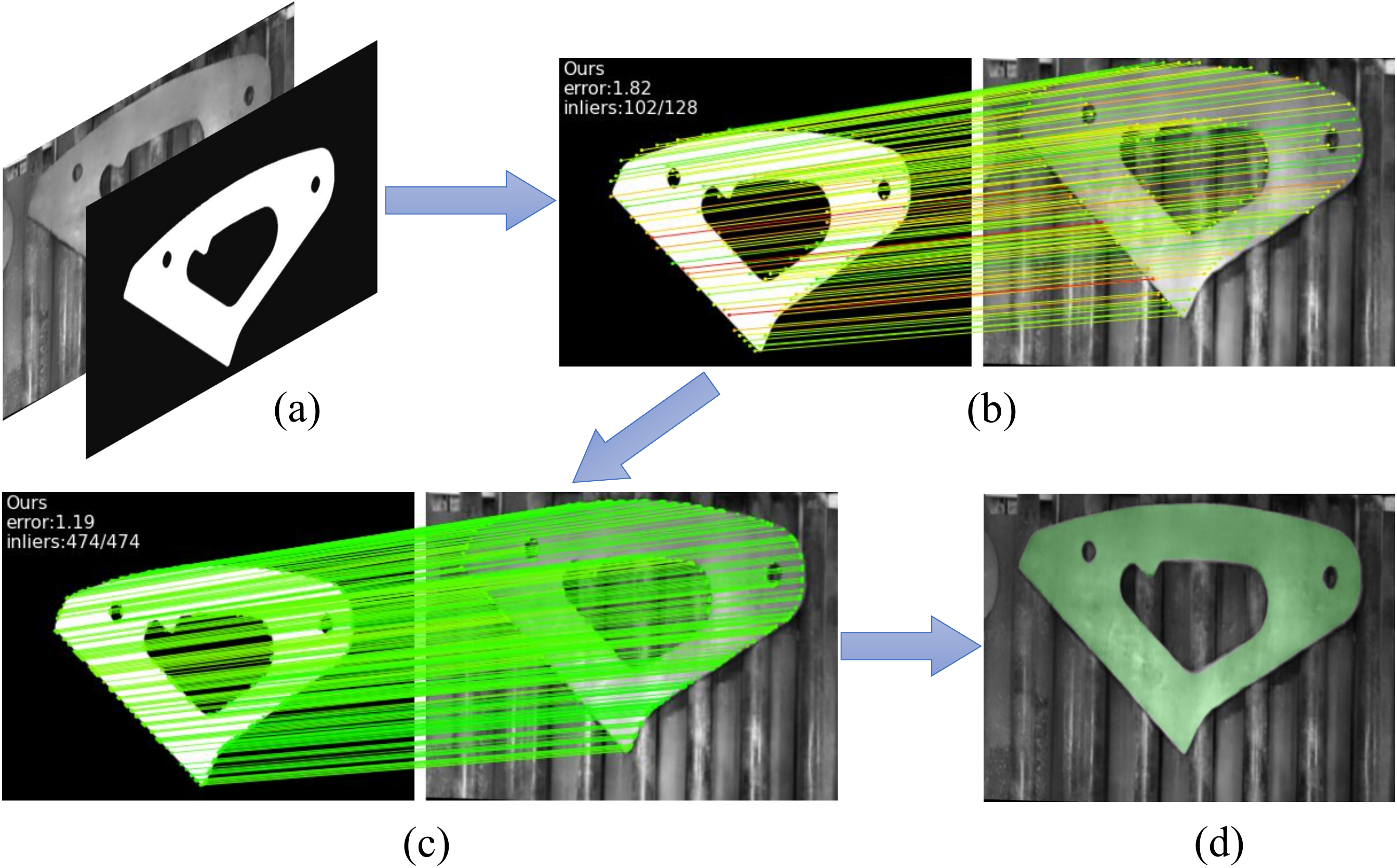}
\end{center}
   \caption{Our template matching method. (a) Template $T$ and image $I$. (b) Coarse matching. (c) Matching refinement. (d) Template warped to the image using the estimated geometric transformation. }
\label{fig:teaser}
\end{figure}
\begin{figure*}[h!t]
\begin{center}
   \includegraphics[width=1.0\linewidth]{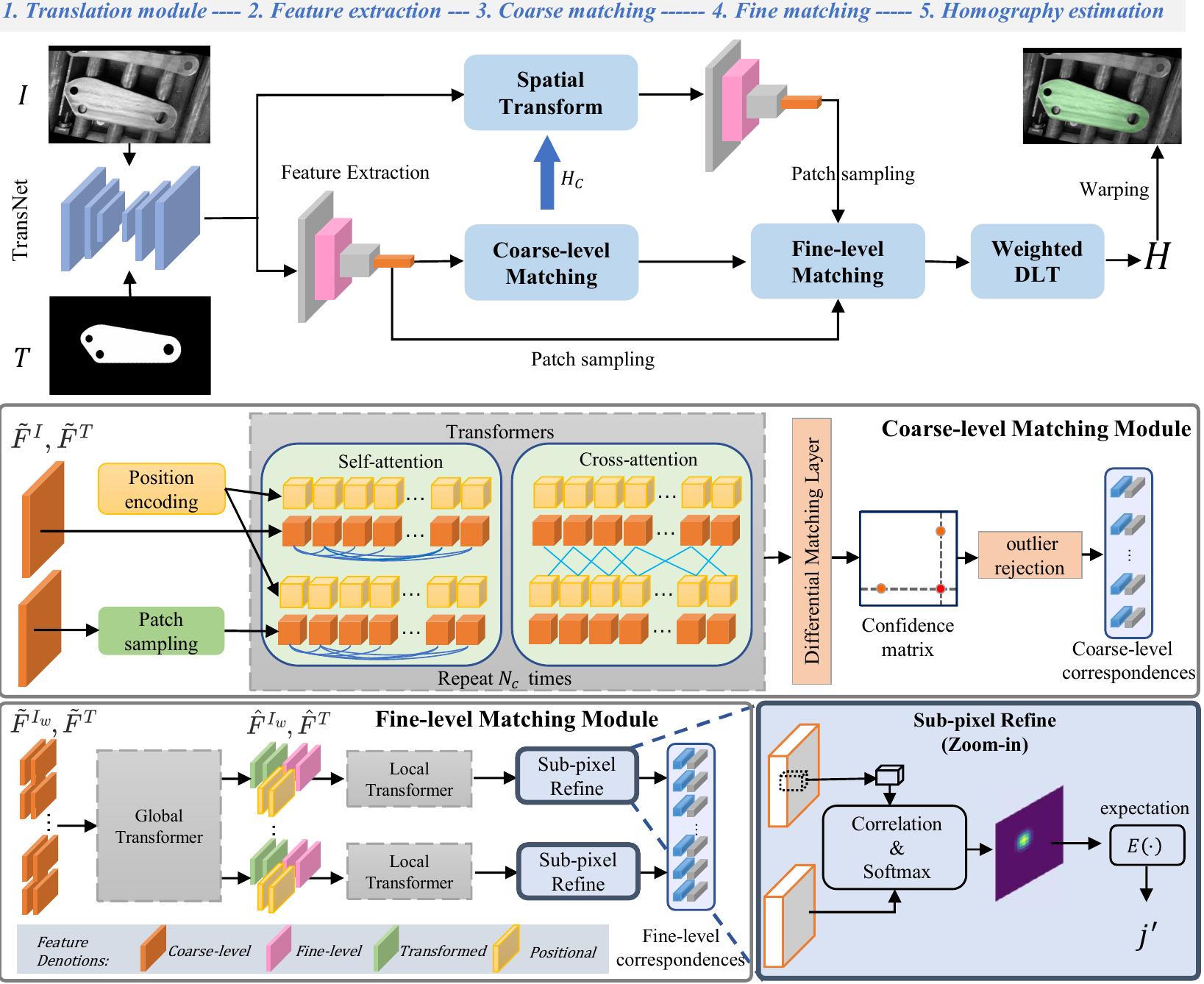}
\end{center}
    \caption{Pipeline: the proposed method has five steps. (1) Translation module: convert the source image $I$ and template mask $T$ into edge maps (Sec. \ref{sec:edge}).  (2) Feature extraction: extract coarse-level feature maps and fine-level feature maps (Sec. \ref{sec:extraction}).  (3) Coarse matching: two sets of coarse-level features are aggregated by interleaving self and cross attention layers to provide the initial homography transformation $H_c$ (Sec. \ref{sec:coarselevel}). 
    (4.) Fine-level matching:  global and local features are fused to give the set of sub-pixel level matches $\mathcal{M}_f$ (Sec.~\ref{sec:finelevelmatching}). (5.) Homography estimation (Sec.~\ref{sec:homography estimation}). }
\label{fig:pipeline}
\end{figure*}

\section{Introduction}

Template matching aims to find given templates in captured images (source images), and is a fundamental technique for many computer vision tasks, including object detection, visual localization, pose estimation, etc. 
Numerous approaches have been proposed to overcome the difficulties and challenges in template matching, and the problem may seem to be solved. Nevertheless, this problem is a critical step in automatic processing on industrial lines, and in real scenarios, various challenges remain, including domain gap, size variance, and pose differences between template and source image. The above challenges motivate our approach of \emph{accurate template matching based on differentiable correspondence refinement}.

Classic methods of template matching~\cite{hinterstoisser2011gradient,ballard1981generalizing,muja2011rein,hinterstoisser2010dominant} generally calculate a similarity score between the template and a candidate image patch.  Linemod-2D~\cite{hinterstoisser2011gradient}  utilizes gradient spreading and gradient orientation similarity measures, achieving real-time detection with high accuracy and robustness for untextured objects and is widely used in industry. However, its performance degrades significantly in the presence of cluttered backgrounds, image blurring or  non-rigid deformation between template and source; these are all common in real world applications. 

Deep learning has shown great potential to overcome such distractors, providing significant improvement for many similar tasks. Related work~\cite{cheng2019qatm,gao2022robust,9486912,wu2017deep} has designed new network structures or similarity metrics based on deep features to improve the robustness of template matching. However, these works all aim to determine the bounding box of the target object, rather than accurate pixel-level matching and pose of the template. Bounding boxes are insufficient for tasks requiring high precision: for example,  robot manipulators need a precise object pose to decide the best grasping direction. Rocco et al.~\cite{rocco2017convolutional} estimate such transformations, but their method fails in cross-modal scenarios where templates are mask images and observed images are in color or grayscale. 

Thus, our work considers the design of an automatic pipeline for determining a high-quality transformation between the template mask and  source image. To allow for the domain difference between a template mask and a grayscale image, an edge translation module is used to convert them to the same modality. To achieve a high-quality transformation estimate, we propose a novel structure-aware outlier rejection approach based on  coarse-to-fine correspondence refinement. As a result, the proposed method not only tolerates different modalities in matching, but also deals with occlusion to some degree as well as complex deformations.

In feature correspondence matching, many recent works~\cite{efe2021dfm,jiang2021cotr,sarlin2020superglue,sun2021loftr} have made remarkable progress in deep feature matching; e.g.\  LoFTR~\cite{sun2021loftr} uses transformers for this task and omits the feature detection step. However, there are limitations when applying LoFTR directly to template matching problems. Firstly, it tends to fail on cross-modal images, when mask images and grayscale images lie in very different in feature spaces. Secondly, the structural consistency of templates and images is not exploited, yet it is critical for accurate matching. More importantly, LoFTR cannot provide sufficiently accurate and reliable correspondences for untextured regions, or when large deformations exist, as in the cross-modal template matching problem.

Motivated by the challenges, we propose a differentiable structure-aware template matching pipeline. To address the modality difference between the template and the source image, we use a translation module to convert both of them to edge maps. 
We believe structural information is particularly important for fast and robust template matching: \emph{the template mask has a specific structure (shape) and correct correspondences between the template and the image should satisfy a specific transformation relationship.} Therefore, we fully exploit template contour information and consider compatibility of angles and distances between correspondences. Specifically, we apply three strategies in our model to better use the structural information of templates and images. Firstly, in order to focus the network  on valid areas, we only sample contour regions of the template as the input. Then the transformer~\cite{vaswani2017attention} using relative positional encoding~\cite{wu2021rethinking} is used to explicitly capture  relative distance information. A method based on distance-and-angle consistency rejects soft outliers.  

In pursuit of high-quality template matches, the transformation between the template and the source image is estimated in a coarse-to-fine style. In the coarse-level stage, we use transformers~\cite{vaswani2017attention}  to encode local features extracted by the convolutional backbone and then establish feature correspondences using a differentiable matching layer. By assigning confidences to these coarse-level matches based on feature similarity and spatial consistency, we obtain a coarse estimate of the geometric transformation, a homography. This coarse matching overcomes differences in scale and large deformations between the source and template image, which is critical for accurate matching at the fine level. We apply the coarse spatial transform~\cite{jaderberg2015spatial} to coarsely align the source image, which then provides an updated source image for the fine level. A refinement module is used at the fine-level to obtain global semantic features and to aggregate features at different scales. We then adopt a correlation-based approach to determine accurate dense matches at the sub-pixel level. These final correspondences are more accurate, and no outlier rejection is needed. All correspondences are used to calculate the final homography. Compared to other recent matching methods~\cite{efe2021dfm,jiang2021cotr,sarlin2020superglue,sun2021loftr}, our correspondences have many fewer outliers, allowing our method to provide robust and accurate template matching without relying on RANSAC~\cite{fischler1981random}. 

We use a linear transformer~\cite{katharopoulos2020transformers} in our pipeline to reduce computational complexity. Farthest point sampling (FPS) is applied to the template image to reduce the input data while retaining its structure. To solve the problem of insufficient training data, GauGAN~\cite{park2019gaugan} is adopted to generate synthetic images of industrial parts for network training.

We have evaluated the proposed method on three datasets, including two newly-collected industrial datasets and a dataset based on COCO~\cite{lin2014microsoft}. Our approach provides significantly improved  homography estimates compared to the best baseline, as we show later. 

Our main contributions can be summarized as:  
\begin{itemize}
	\item An accurate template matching method, robust in challenging scenarios including cross-modality images, cluttered backgrounds, and untextured objects. 
	\item A structure-aware, fully differentiable, template matching pipeline, avoiding the use of RANSAC found in other feature matching approaches; it achieves state-of-the-art accuracy. 
		\item Two new datasets with accurate ground truth, of potential benefit to future research on learning-based template matching. 
\end{itemize}

\section{Related Work}

\subsection{Template Matching}
Traditional methods of template matching mostly rely on comparing  similarities and distances between the template and candidate image patch, using such approaches as sum of squared differences (SSD), normalized cross-correlation (NCC), sum of absolute differences (SAD), gradient-based measures, and so on. Linemod-2d  and the generalized Hough transform (GHT)~\cite{ballard1981generalizing} are widely applied in  industry. Such approaches 
degrade significantly in the presence of cluttered backgrounds, image blurring or large deformations. Deep learning-based template matching algorithms~\cite{cheng2019qatm,gao2022robust,rocco2017convolutional,9486912,wu2017deep} can handle more complex deformations between the template and source image. They usually adopt trainable layers with parameters to mimic the functionality of template matching. Feature encoding layers are assumed to extract the features from both inputs; these deep feature encoders dramatically improve template matching results. While these methods still rely on the rich textures of input images. However, deep learning methods are prone to fail with cross-modal input and are typically unable to provide an accurate pose for the target object.

Motivated by these challenges, our method predicts a homography transformation, and uses an edge-aware module to eliminate the domain gap between the mask template and the grayscale image for robust matching.

\subsection{Homography Estimation}
Classical homography estimation methods usually comprise three steps: keypoint detection (using e.g.\ SIFT~\cite{lowe2004distinctive}, SURF~\cite{bay2006surf}, or ORB~\cite{rublee2011orb}), feature matching (feature correlation), and robust homography estimation (using e.g.\  RANSAC~\cite{fischler1981random} or MAGSAC~\cite{barath2019magsac}). However, RANSAC-like approaches are non-differentiable. Furthermore, differentiable RANSAC algorithms~\cite{Brachmann_2019_ICCV,brachmann2017dsac}  hinder  generalization to other datasets. Other methods, such as the seminal Lucas-Kanade algorithm~\cite{lucas1981iterative}, can directly estimate the homography matrix without detecting features. 
The first deep learning-based homography estimation model was proposed in~\cite{detone2016deep}. Its network regresses the four corner displacement vectors of the source image in a supervised manner and yields the homography using  a direct linear transform (DLT)~\cite{hartley2003multiple}.  Many unsupervised approaches~\cite{nguyen2018unsupervised,zhang2020content,koguciuk2021perceptual} have been proposed to minimize the pixel-wise photometric error or feature difference between the template and source image. 
 
These methods are likely to fail under large viewpoint change, when   textures are lacking, and for differing input modalities. Our work uses the template's structural (shape) properties and samples valid region features in the template to learn the correlation with the source image. An edge-aware module is used to translate the source image and template mask to bypass the effect of modality differences between two inputs. 

\subsection{Feature Matching}
Before the era of deep learning, hand-crafted local features such as SIFT, SURF, and ORB were widely adopted. Deep learning-based methods~\cite{detone2018superpoint, dusmanu2019d2,yi2016lift} significantly improve the feature representation, especially in  cases of significant viewpoint and illumination changes.
SuperPoint~\cite{detone2018superpoint}, D2-Net~\cite{dusmanu2019d2}  and ASLFeat~\cite{luo2020aslfeat}  propose joint learning of feature descriptors and detectors; most computations of the two tasks are shared for fast inferencing using a unified framework. A significant improvement in feature matching was achieved by SuperGlue~\cite{sarlin2020superglue}, which accepts two sets of keypoints with their descriptors, and updates their representations with an attentional graph neural network (GNN).
Drawing inspiration from GNN, more  methods~\cite{chen2021learning,jiang2022glmnet,shi2022clustergnn,roessle2022end2end}  further improve the accuracy and efficiency of graph-based feature matching.
Recently, several works~\cite{jiang2021cotr,suwanwimolkul2022efficient,sun2021loftr} have attempted to adopt transformers to model the relationship between features and provide impressive results.
In this work, we build on the success of transformers and  learn accurate template matching with coarse-to-fine correspondence refinement.

\subsection{Vision Transformers}
Transformers~\cite{vaswani2017attention} were initially proposed in natural language processing (NLP). Vision transformers~\cite{carion2020end} have attracted attention due to their simplicity and computational efficiency for image sequence modeling. Many variants~\cite{katharopoulos2020transformers,kitaev2020reformer,tay2020efficient,lan2022arm3d} have been proposed for more efficient message passing. In our work, we utilize self and cross attention to establish larger receptive fields and capture structural information from the inputs. In particular,  linear transformers~\cite{katharopoulos2020transformers} with relative positional encoding are adopted 
to ensure low computational costs and more efficient message passing. 

\section{Overview}

In industrial template matching, it is usual for the template to be represented as a binary mask indicating only the shape of the source object. In contrast, the source image is often grayscale. Thus, we first use an edge-aware translation module before feature extraction to eliminate the domain difference between these two images: see Sec.~\ref{sec:edge}. We propose a differentiable feature extraction and aggregation network with transformers in Sec.~\ref{sec:extraction}-\ref{sec:transformer}. The whole matching pipeline is performed in a coarse-to-fine style. At the coarse level, to estimate the homography from matched features, we combine spatial compatibility and feature similarity for soft outlier filtering: see Sec. \ref{sec:coarselevel}; this is RANSAC-free and differentiable. A coarse homography is obtained from the coarse correspondences. Then, we apply the spatial transform~\cite{jaderberg2015spatial} to the source image to provide a coarsely-aligned image. At the fine-level, we combine global semantics and local features to achieve sub-pixel dense matching and obtain an accurate homography estimate, as explained in Sec. \ref{sec:finelevel}. The final correspondences are precise between the template mask and source image, ensuring a plausible template matching result. 
Inspired by LoFTR, we adopt a coarse-to-fine matching pipeline, as shown in Fig.~\ref{fig:pipeline}.
Note that unlike LoFTR, our approach takes full advantage of the geometric properties of the template and  spatial consistency between the template and the object. In addition, our coarse-to-fine matching process is fully differentiable via a spatial transform connection, while LoFTR's coarse-to-fine strategy only enhances correspondence accuracy and  is not fully differentiable.  

\section{Method}
\subsection{Task}
Given a binary template image $T$ and a source search image $I$, our method aims to estimate a homography transformation between $T$ and $I$ to provide the precise position and pose of the object in the image $I$. For applications whose scenes have multiple objects and multiple candidate templates, the coarse stage of our method may be performed first to estimate the initial homography for selecting the correct template for each object. We then use the refinement stage to obtain the precise position and pose. 

\subsection{Feature Extraction and Aggregations}\label{sec:feature}
\subsubsection{Edge translation}\label{sec:edge} 
Unlike some other template matching and homography estimation tasks, the case considered here has a domain difference between the template $T$ and source image $I$. The former is a binary mask, and the latter is a grayscale image; their features are too different to use common image matching approaches. Grayscale images may furthermore exhibit strong reflections if the material is glossy. Matching based on photometric similarity are not applicable in such cases. 
Firstly, to ensure domain consistency of the template and source image, and to avoid complications from reflections, 
we adopt a translation network to convert both into edge maps. 
In this step, we adopt PiDiNet~\cite{su2021pixel}, a lightweight and robust edge detector, to compute the edge maps. This conversion is crucial to permit later feature matching. 

\subsubsection{Feature extraction}\label{sec:extraction}
We use a standard convolutional architecture similar to SuperPoint  to extract features at different scales from both images after translation. SuperPoint  has a  VGG-style~\cite{simonyan2014very} encoder trained by self-supervision and shows leading performance in many vision tasks~\cite{jau2020deep,sarlin2020superglue,wang2020learning}. We only retain the encoder architecture of SuperPoint   as our local feature extraction network. Given an input image of size $H \times W$, our feature extraction networks produce feature maps at four resolutions; we save the second layer feature map ($\hat{F}\in \mathbb{R}^{{H}/{2} \times {W}/{2} \times D}$) and the last layer feature map ($\tilde{F}\in \mathbb{R}^{{H}/{8} \times {W}/{8}  \times C}$). Thus, $\tilde{F}^{T}$ and $\tilde{F}^{I}$ are the coarse-level features, $\hat{F}^{T}$ and $\hat{F}^{I}$ are the fine-level features.

\subsubsection{Feature aggregation with transformers}\label{sec:transformer}
Since edge images are not richly textured, the features extracted by the local convolutional neural network are inadequate for robust feature matching. Structural and geometric features are more significant~\cite{zhou2022geometry}. 
Therefore, we adopt transformer blocks~\cite{vaswani2017attention} to encode  $\tilde{F}^{T}$ and $\tilde{F}^{I}$ to produce more global, position-aware features denoted  $\tilde{F}_\mathrm{tr}^{T}$ and $\tilde{F}_\mathrm{tr}^{I}$. A transformer block consists of a self-attention layer to aggregate the global context and a cross-attention layer to exchange information between two feature sets.     

\emph{Patch sampling.} Unlike previous work~\cite{jiang2021cotr,sun2021loftr} which passes all patches of the image into the transformer module, we only keep meaningful feature map patches in $\tilde{F}^{T}$. Specifically,
we use furthest point sampling~\cite{qi2017pointnet++} to sample $N_p$ patches in which edge pixels exist, both to reduce computational cost and increase the efficiency of message passing. $\tilde{F}^{T}$ henceforth denotes the feature map after sampling. 
We do not drop any patches of the source image $I$: every location in $I$ could be a potential match. We perform experiments to show the effect of patch sampling with various numbers of patches in {Sec.~\ref{sec:ablation}}. 

\emph{Positional encoding.} 
In transformers, all inputs are fed in simultaneously, and furthermore, do not encode any information concerning input ordering (unlike RNNs). We must encode positional information for the  tokens input into transformers in order to make available the order of the sequences.
Previous feature matching work using transformers~\cite{jiang2021cotr,sun2021loftr} uses a 2D extension of the standard absolute positional encoding, following DETR~\cite{carion2020end}. In contrast, \cite{li2022lepard,wu2021rethinking} showed that relative positional encoding is a better way of capturing the positional relationships between input tokens. We employ a rotary position embedding~\cite{su2021roformer} proposed in natural language processing for position encoding which has recently been  successfully adopted for point cloud processing~\cite{li2022lepard}. We apply it to  2D images as it can express a relative position in a form like absolute position encoding. Furthermore, it can be perfectly incorporated in linear attention~\cite{katharopoulos2020transformers} at almost no extra cost. In order to obtain the relative positional relationship of the local features between the template and image, we thus use relative positional encoding in a linear transformer. 
For a given 2D location ${n}=(x,y)\in{\mathbb{R}^2}$, and its feature ${f}_n \in{\mathbb{R}^C}$, the relative positional encoding is defined as:  
\begin{equation*}
\mathcal{P}(n,{f}_n) = \Theta(n){f}_n=
	\begin{pmatrix}
	 M_{1}   & &  \\
	  &   \ddots&\\
	    &  &M_{C/4}
	 \end{pmatrix} {f}_n,
\end{equation*}
where  
\begin{equation*}
M_k = 
\begin{pmatrix}
\cos\,x\theta_k & -\sin\,x\theta_k  & 0&0  \\
\sin\,x\theta_k & \cos\,x\theta_k  & 0&0  \\
0 & 0  & \cos\,y\theta_k & -\sin\,y\theta_k  \\
0 & 0  & \sin\,y\theta_k & \cos\,y\theta_k  \\
 \end{pmatrix},
\end{equation*} 
\[
\theta_{k} = \frac{1}{10000^{4(k-1)/C}},\quad k \in[1,...,C/4],
\] 
and $C$ is the number of feature channels. 

Rotary position embedding satisfies: 
\begin{equation}
(\Theta(m)f_m)^{T}(\Theta(n)f_n)
=f_m^{T}\Theta(n-m)f_n
\end{equation}
and $\Theta(n-m)=\Theta(m)^{T}\Theta(n)$. Thus, relative position information between features $f_n$ and $f_m$ can be explicitly revealed by taking the dot product in the attention layer. This position encoding is more suitable in our application than absolute positional encoding, since relative positional relationships between template $T$ and image $I$ is crucial.  $\Theta(.)$ is an orthogonal operation on features, which means it only changes the directions but not the lengths of feature vectors. Therefore, rotary position embedding stabilizes and accelerates the training process~\cite{li2022lepard}, facilitating downstream feature matching tasks.
An experimental comparison to absolute positional encoding can be found in {Sec.~\ref{sec:ablation}}. 

\begin{figure}[t]
\begin{center}
   \includegraphics[width=1.0\linewidth]{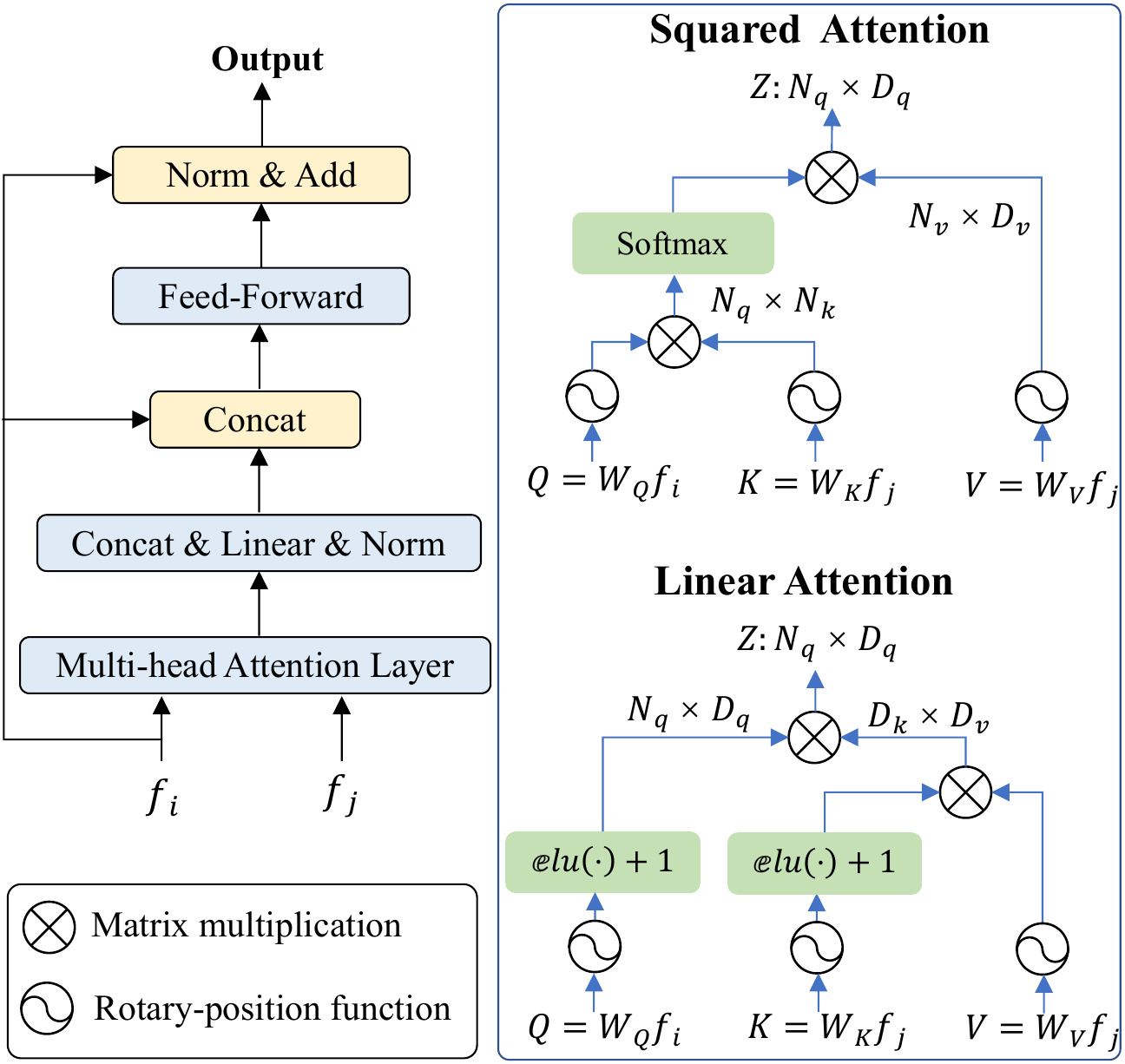}
\end{center}
   \caption{
    {Architecture of encoder and attention layers.}
     Left: encoder. Right: squared ($O(N^2)$ complexity) attention layer  and linear ($O(N)$ complexity) attention layer . 
   }
\label{fig:attention}
\end{figure}

\emph{Self-attention and cross-attention layers.}
The key to the transformer model is attention. We use self and cross attention alternately in our pipeline. The input vectors for an attention layer are query vector $Q$, key vector $K$, and value $V$, and a basic attention layer is given by:  
\[
\mathrm{Attention}(Q,K,V) = \mathrm{Softmax}(QK^T)V.
\] 
Suppose  $Q$ and $K$ have length $N$, and their feature dimensionality is $C$. Then the computational cost of the transformer grows as the square  of the length of the input. The length of the source image $T$'s input token makes a basic version of the transformer impractical for local feature matching. Following~\cite{sun2021loftr}, we adopt a more efficient variant of the attention layer, linear transformer~\cite{katharopoulos2020transformers}. We use a kernel function $\mathrm{sim}(Q,K) = \phi(Q)\phi(K)^{T}$ to replace the softmax calculation, where $\phi(.)=\mathrm{elu}(.)+1$. The computational cost is reduced from $O(N^2)$ to $O(N)$ when $C \ll N$.  Following RoFormer~\cite{su2021roformer}, we do not inject rotary
position embedding in the denominator to avoid the risk of dividing by zero. Differing from~\cite{su2021roformer,li2022lepard} as well as query $Q$ and key $K$, the value $V$ is also multiplied by $\Theta(\cdot)$, since we consider the position information to be important auxiliary information for value $V$. Experiments justifying this approach are described in Sec.~\ref{sec:ablation}. 

Overall, each token in a linear transformer with relative positional encoding is given by:  
\begin{eqnarray*}
&\mathrm{Attention}(Q,K,V)_m=  \\
&\frac{\sum_{n=1}^{N}\big(\Theta(m)\phi(q_m)\big)^{T}\big(\Theta(n)\phi(k_n)\big)\big(\Theta(n)v_n\big)}{\sum_{n=1}^{N}\phi(q_m)^T\phi(k_n)}.
\end{eqnarray*}

\subsection{Coarse Matching}\label{sec:coarselevel}
\subsubsection{Establishing coarse matches} 
We establish coarse matches using the transformed features $\tilde{F}_\mathrm{tr}^{T}$ and $\tilde{F}_\mathrm{tr}^{I}$. An optimal transport (OT) layer  is adopted as our differentiable matching layer. We first calculate a score matrix $S$ using dot-product similarity of the transformed features: 
\[
S(i,j) = \left \langle \tilde{F}_\mathrm{tr}^{T}(i),\tilde{F}_\mathrm{tr}^{I}(j)\right \rangle.
\]
This score matrix $S$ is used as the cost matrix in a partial assignment problem, following~\cite{sun2021loftr,sarlin2020superglue}. This optimization problem can be efficiently solved with the Sinkhorn algorithm~\cite{cuturi2013sinkhorn} to obtain the confidence assignment matrix $C$. 

To obtain more reliable matches, the mutual nearest neighbor (MNN) criterion is enforced, and only  matching pairs with confidence values higher than a threshold $\theta_c$ are preserved. The set of coarse-level matches $\mathcal{M}_c$ is thus:
\[
\mathcal{M}_c = \left\{\,(i,j)\ |\ {\forall}\,(i,j)\in  \mathrm{MNN} (C),\ C(i,j)\geq \theta_c \right\}. 
\]
Another matching layer approach is based on dual-softmax (DS)~\cite{rocco2018neighbourhood,tyszkiewicz2020disk}. It applies softmax
to both dimensions of $S$ to get the probability of a mutual nearest neighbor match.
A comparison of OT and DS methods can be found in Sec.~\ref{sec:ablation}. 

\subsubsection{Confidence weights based on spatial consistency}
The differentiable matching layer provides a tentative match set $\mathcal{M}_c$ based on feature dot-product similarity. In this way, two irrelevant points may be regarded as a matching pair due to similarity of appearance. To prevent this, we add a new constraint, based on the observation that template matching, has an essential property: correct correspondences (inliers) have similar geometric transformations, while transformations of outliers are random.  

 RANSAC  and its variants~\cite{barath2018graph,chum2003locally} are widely adopted for outlier rejection. However, such methods are slow to converge and may fail in cases of high outlier ratios. In contrast, spectral matching (SM)~\cite{leordeanu2005spectral} and its variants~\cite{bai2021pointdsc,chen2022sc2,quan2020compatibility,yang2019performance} significantly improve results for rigid point cloud registration, by constructing a compatibility graph which preserves angle or distance invariance between point pairs. In contrast, our model assumes a non-rigid deformation in which pairwise distances between far points are more likely to vary than between closer ones. We thus extend SM and propose a method based on distance-and-angle consistency for non-rigid deformation outlier rejection.  

\begin{figure}[t]
\begin{center}
   \includegraphics[width=0.9\linewidth]{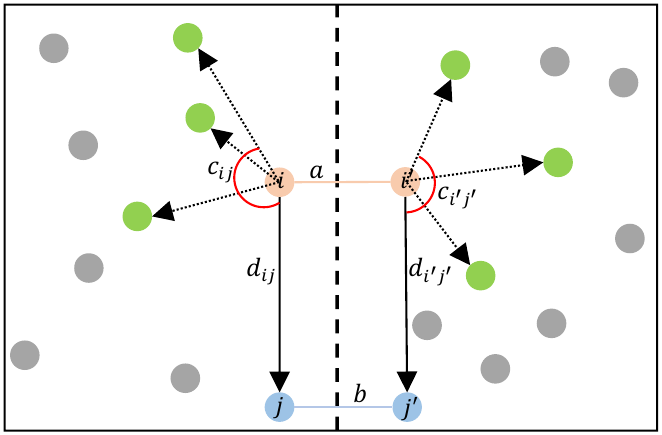}
\end{center}
   \caption{
    Given two matching pairs $a=(i,i^{'})$ and $b=(j,j^{'})$, we calculate both their distance compatibility and their angular compatibility. Green nodes represent $k$-nearest neighbors.
   }
\label{fig:distance-angele-compatibility}
\end{figure}

Let {$ \beta$} denote the distance compatibility term measuring the change in lengths of  matched pairs. To allow for scale differences, we first normalize the distances between matching points on the template and image separately. Then for two coarse matches $a=(i,i^{'})$ and $b=(j,j^{'})$, {$ \beta$} is defined as:
\[
\beta_{(a,b)} = \left[ 1-{{(d_{ij}/d_{i^{'}j^{'}}-1)}^2}/{\sigma_d^{2}}\right]_{+},
\]
where $d_{ij}$ is the normalized pairwise distance between $i$ and $j$, $\left[\cdot\right]_{+}$ means $\max(\cdot,0)$, and $\sigma_d$ is a distance parameter controlling  sensitivity to changes in relative length. Changes in directions are also penalized using a triplet-wise angle. Inspired by~\cite{qin2022geometric}, we compute angular compatibility from triplets of coarse feature points. For a matching pair $a=(i,j)$ with positions $p_i$ and $p_j$,  we first select the $k$ nearest neighbors $\mathcal{N}_i$ of $p_i$. For each $p_x \in \mathcal{N}_i$, the angle $c_{i,j}^{x}=\measuredangle(\triangle_{i,x},\triangle_{i,j})$, where $\triangle_{i,j}=p_i-p_j$. To improve  robustness, we select the maximum  value $c_{ij}$ among the $k$ nearest neighbors  as the angle property for a matching pair $(i,j)$.
As for  distance compatibility {$ \beta$}, we now formulate the angular compatibility $\alpha$ as:
\[
\alpha_{(a,b)} = \left[ 1-{{(c_{ij}-c_{i^{'}j^{'}})}^2}/{\sigma_{\alpha}^{2}}\right]_{+},
\]
where $\sigma_{\alpha}$ is the angular parameter
controlling the sensitivity to changes inn angle. Fig.~\ref{fig:distance-angele-compatibility} illustrates the computation of  distance and angular consistency.

The final spatial compatibility of matches $a$ and $b$ is defined as: 
\[
E(a,b)=\lambda_c \alpha_{(a,b)} + (1-\lambda_c) \beta_{(a,b)}, 
\]
where $\lambda_c$ is a  control weight. $E(a,b)$ is large only if the two correspondences $a$ and $b$ are highly spatially compatible. Following~\cite{bai2021pointdsc,leordeanu2005spectral}, the leading eigenvector ${e}$ of the compatibility matrix ${E}$ is regarded as the inlier probability of the matches. We use the power iteration algorithm~\cite{mises1929praktische} to compute the leading eigenvector $e \in \mathbb{R}^k$ of the matrix $E$.   

\subsubsection{Initial homography estimation} 
Naturally, the inlier probability $e$ together with feature score $s$ must be combined to give the final overall inlier probability, where $s$ is the corresponding element of the feature confidence matrix $C$. We simply compute $w_k = s_k \cdot e_k$: intuitively, $w_k$ takes into account how similar the feature descriptors are ($s_k$) and how much the spatial arrangement is changed ($e_k$) for a matching pair $k$. Finally, we use the confidence $w_k$ as a weight to estimate the homography transformation $H_c$, using the DLT formulation~\cite{hartley2003multiple}. A weighted least squares solution is found to the linear system. The matches-with-confidence make our coarse-to-fine network differentiable and RANSAC-free, enabling end-to-end training. The effectiveness of confidence weights is explored in Sec.~\ref{sec:ablation}.

\subsubsection{Coarse-level training losses} 
Following~\cite{sun2021loftr}, we use  negative log-likelihood loss over the confidence matrix $C$ returned by either the optimal transport layer or the dual-softmax operation to supervise the coarse-level network. The ground-truth coarse matches $\mathcal{M}_c^\mathrm{gt}$ are estimated from the ground-truth relative transformations (homographies). Using an optimal transport layer, the loss is:
\begin{equation*}
    \begin{aligned}
        \mathcal{L}_c =& - \frac{1}{\left|\mathcal{M}_c^\mathrm{gt} \right |}\ \sum_{(i,j)\in\mathcal{M}_c^\mathrm{gt}} \text{log}\ (i,j)\ -\\
           &\frac{1}{\left|\mathcal{I} \right |}\ \sum_{(i,j)\in\mathcal{I}} \text{log}\ (i,j),
    \end{aligned}
\end{equation*}
where $(i,j)\in \mathcal{I}$ means $i$ or $j$ do not have any reprojection in the other image. With the dual-softmax operation, we minimize the negative log-likelihood loss in $\mathcal{M}_c^\mathrm{gt}$ :
$$
\mathcal{L}_c = - \frac{1}{\left|\mathcal{M}_c^\mathrm{gt} \right |}\ \sum_{(i,j)\in\mathcal{M}_c^\mathrm{gt}} \text{log}\ (i,j)\, . 
$$

\subsection{Fine-level Matching}\label{sec:finelevel}

A coarse-to-fine scheme is adopted in our pipeline, a scheme which has been successfully applied in many vision tasks~\cite{efe2021dfm,mok2022affine,parihar2021rord,shen2020ransac,sun2021loftr,truong2020glu}. We apply the obtained coarse homography $H_c$ to the source image $I$ to generate a coarsely-aligned image $I_{w}$. We roughly align the two images, then use a refinement  network to get sub-pixel accurate matches, and finally, a better-estimated transformation matrix is produced from the new matches.

\subsubsection{Fine-level matching network}\label{sec:finelevelmatching}  
For a given pair of coarsely aligned images (warped image $I_w$ and template $T$), sub-pixel level matches are calculated by our fine-level matching network to further enhance the initial alignment.  Although~\cite{efe2021dfm,tyszkiewicz2020disk} claim that local features significantly improve matching accuracy in feature matching when refining, we find that local features are insufficient to achieve robust and accurate matching in untextured cases. Instead, we combine the global transformer and local transformer for feature aggregation to improve fine-level matching, as shown in Fig.\ref{fig:pipeline}.   

The global transformer is first adopted to aggregate coarse-level features as priors. In detail, for every sampled patch pair ($\tilde{i}$, $\tilde{j}$) at the same location on template $T$ and warped image $I_w$, the corresponding coarse-level features are denoted  $\tilde{F}^{T}(\tilde{i})$ and $\tilde{F}^{I_w}(\tilde{j})$, respectively. A global transformer module with $N_f$ self- and cross-attention layers operates on these coarse-level features to produce  transformed feature$(\tilde{F}_\mathrm{tr}^{T}(\tilde{i}),\tilde{F}_\mathrm{tr}^{I_w}(\tilde{j}))$.
Note that, for efficiency we only consider those patches 
which coarse matching sampled. To deeply integrate global and local features,  $\tilde{F}_\mathrm{tr}^{T}(\tilde{i})$ and $\tilde{F}_\mathrm{tr}^{I_w}(\tilde{j})$ are upsampled and concatenated with corresponding local (fine-level) features $\hat{F}^{T}(\tilde{i})$ and $\hat{F}^{I_w}(\tilde{j})$, respectively. Subsequently, the concatenated features are used as inputs to a 2-layer MLP to reduce the channel dimensionality to the same as for the original local features, yielding the fused features. The effectiveness of this module is demonstrated in Sec.~\ref{sec:ablation}.

For every patch pair ($\tilde{i}$, $\tilde{j}$), we then locate their all finer positions ($i$,$j$) where $i$ lies on the edge. As fused feature maps, we crop two sets of local windows of size $w\times w$ centered at $({i},{j})$ respectively. A local transformer module operates $N_f$ times  within each window to generate the final features $(F^{T}(i),F^{I_w}(j))$. Following~\cite{wang2020learning,sun2021loftr}, the center vector of $F^T(i)$ is correlated with all vectors in $F^{I_w}(j)$ resulting in a heatmap that represents the matching probability for each pixel centered on $j$ with $i$.  Using 2D softmax to compute expectation over the matching probability distribution, we get the final position $j'$ with sub-pixel accuracy matching $i$. The final set of fine-level matches $\mathcal{M}_f$ aggregates  all matches $(i,j')$.

\subsubsection{Fine-level homography estimation}\label{sec:homography estimation}  
\begin{figure*}[!t] 
 \begin{center}
 SuperGlue~~~~\qquad\qquad\qquad \qquad COTR~~~\qquad\qquad\qquad\qquad~~ LoFTR~~\qquad\qquad\qquad\qquad~~  Ours \\
    \begin{minipage}{0.015\linewidth}
    \rotatebox{90}{{~~~~~~~~Mechanical parts dataset}}

    \rotatebox{90}{{~~~~~~Assembly hole dataset~~~~~~~~~~~~~}}

    \rotatebox{90}{{~~~~COCO dataset~~~~~~~~~~~}}

    \end{minipage}
    \begin{minipage}{0.24\linewidth}
       \centerline{\includegraphics[width=\textwidth]{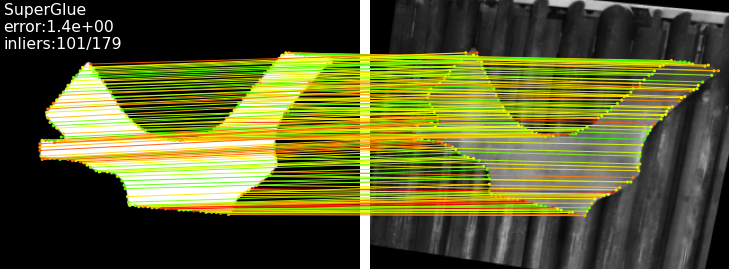}}
        \centerline{\includegraphics[width=\textwidth]{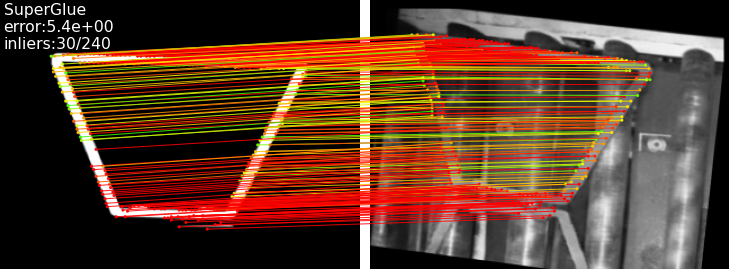}}
        
     \centerline{\includegraphics[width=\textwidth]{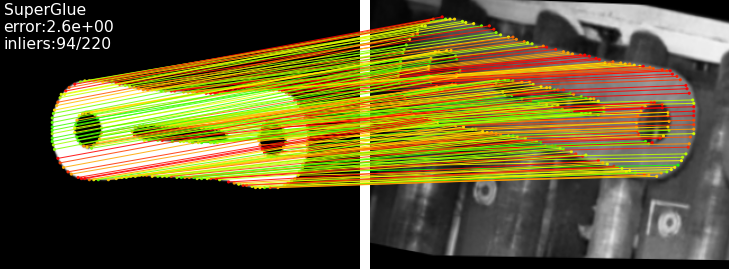}}
    
     \centerline{\includegraphics[width=\textwidth]{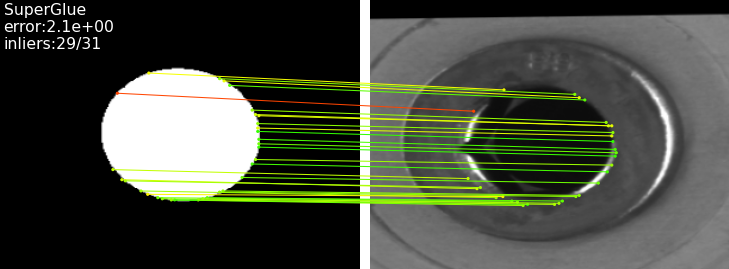}}
        
     \centerline{\includegraphics[width=\textwidth]{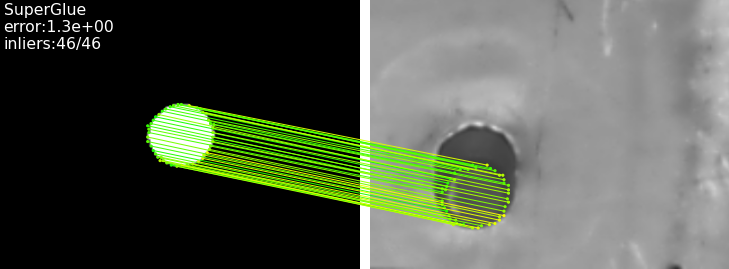}}
        
     \centerline{\includegraphics[width=\textwidth]{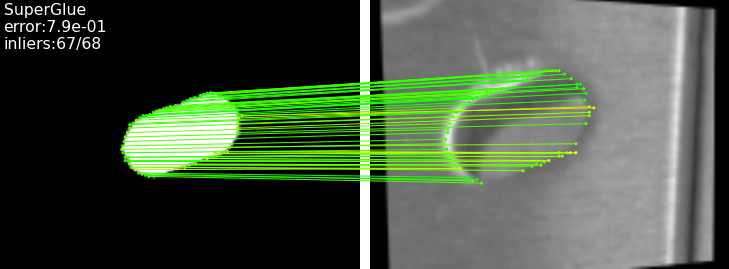}}
       
     \centerline{\includegraphics[width=\textwidth]{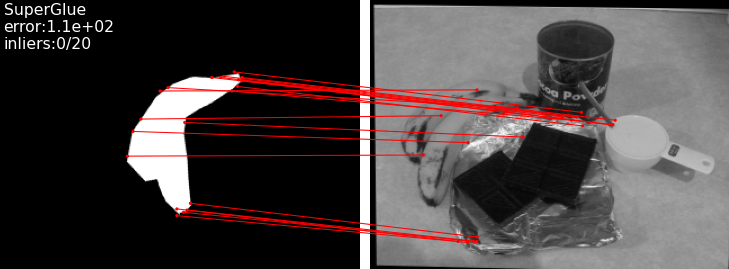}}
        
     \centerline{\includegraphics[width=\textwidth]{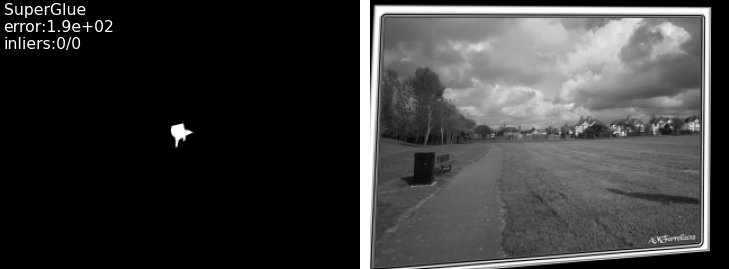}}
        
     \centerline{\includegraphics[width=\textwidth]{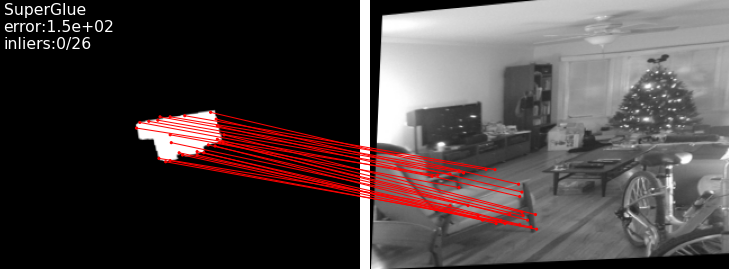}}
 \end{minipage}    
    \begin{minipage}{0.24\linewidth}
        
     \centerline{\includegraphics[width=\textwidth]{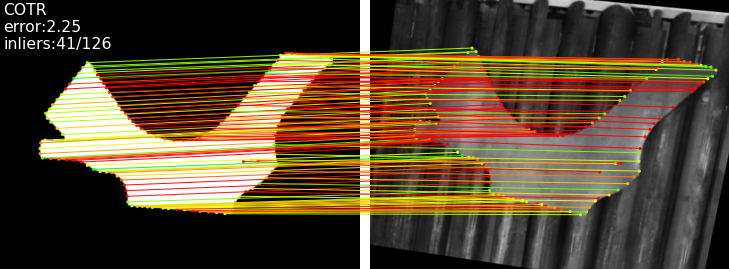}}
        
     \centerline{\includegraphics[width=\textwidth]{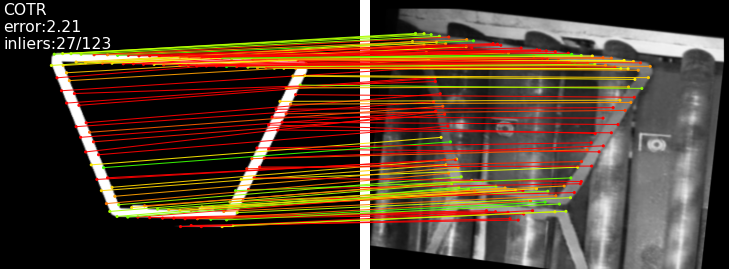}}
        
     \centerline{\includegraphics[width=\textwidth]{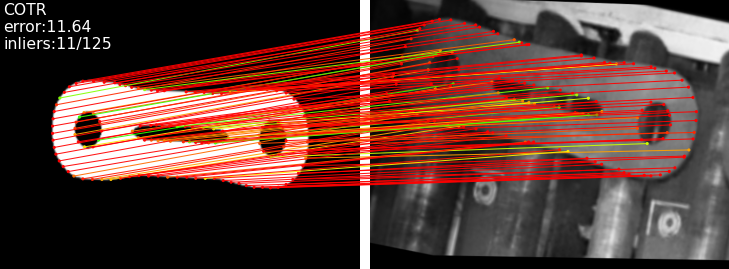}}
    
     \centerline{\includegraphics[width=\textwidth]{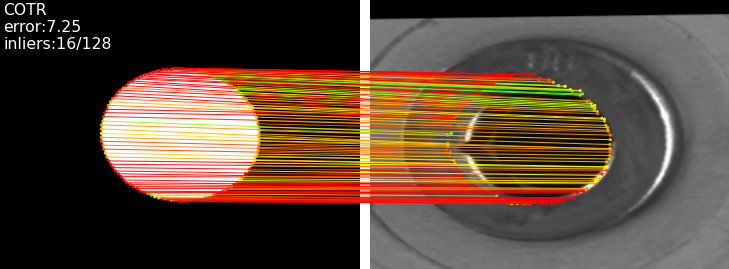}}
        
     \centerline{\includegraphics[width=\textwidth]{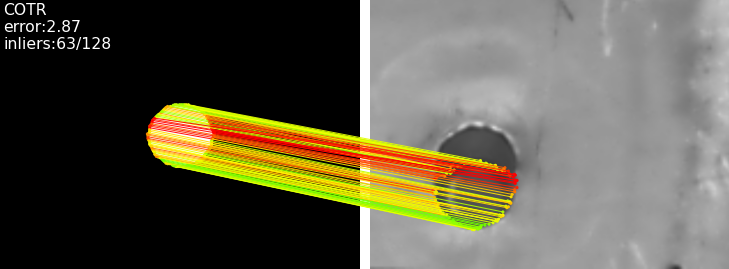}}
        
     \centerline{\includegraphics[width=\textwidth]{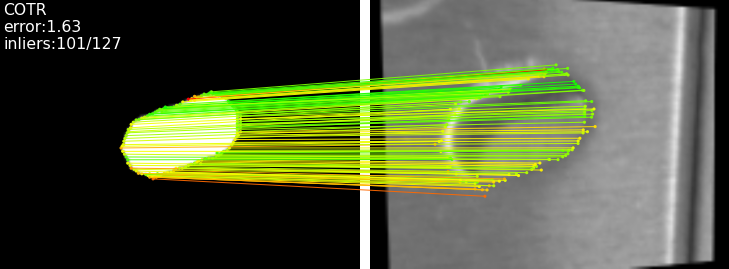}}
         
     \centerline{\includegraphics[width=\textwidth]{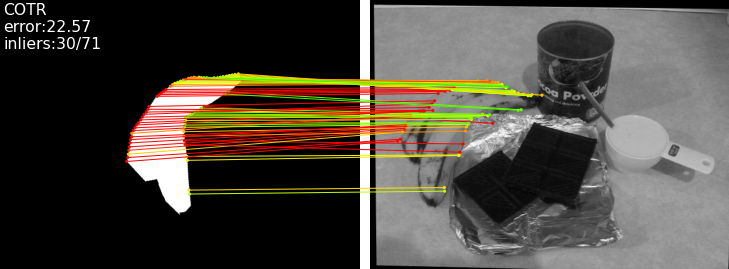}}
        
     \centerline{\includegraphics[width=\textwidth]{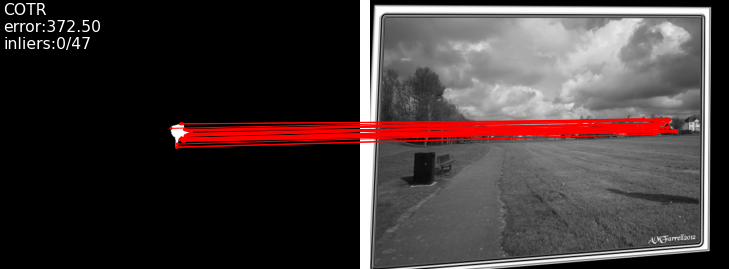}}
        
     \centerline{\includegraphics[width=\textwidth]{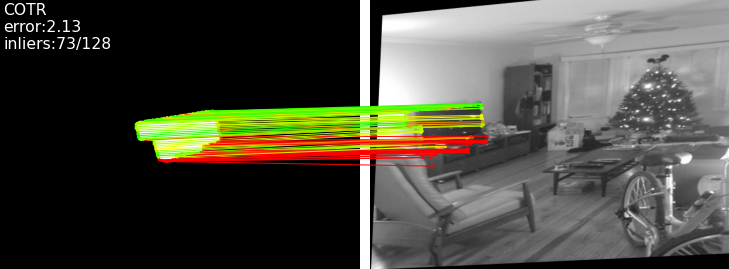}}
 \end{minipage}
    \begin{minipage}{0.24\linewidth}

     \centerline{\includegraphics[width=\textwidth]{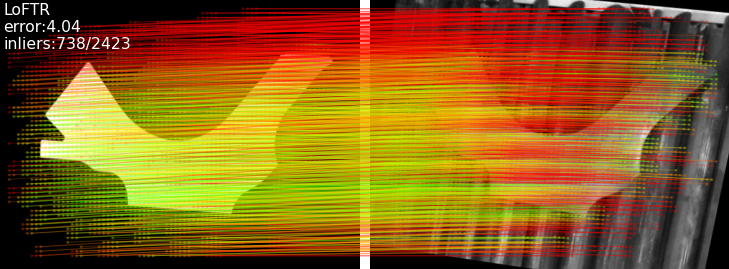}}
        
     \centerline{\includegraphics[width=\textwidth]{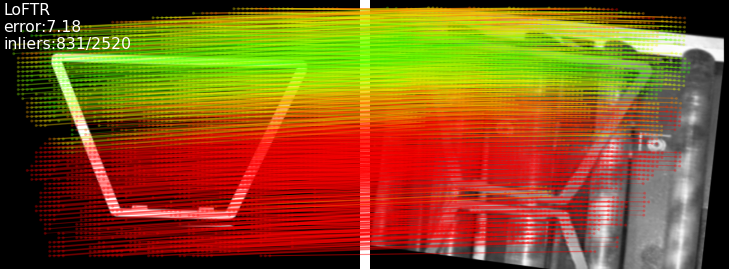}}
        
     \centerline{\includegraphics[width=\textwidth]{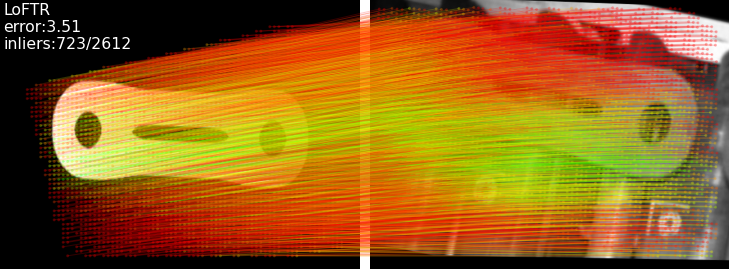}}
     
     \centerline{\includegraphics[width=\textwidth]{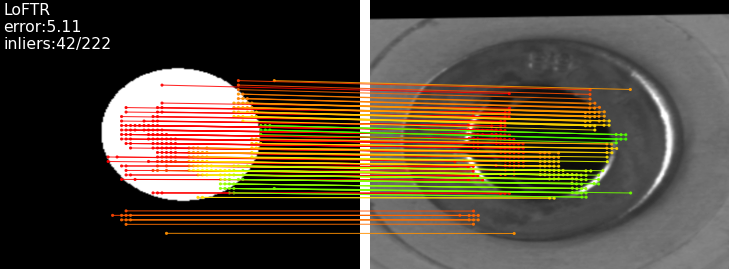}}
        
     \includegraphics[width=\textwidth]{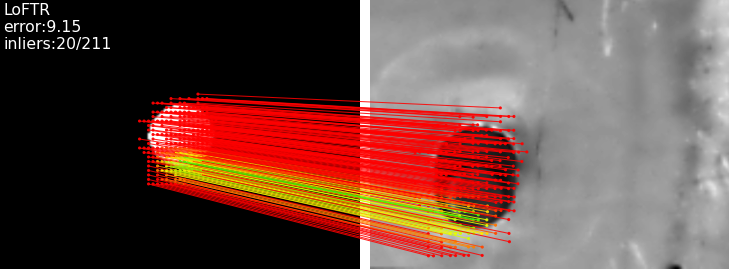}
        
     \centerline{\includegraphics[width=\textwidth]{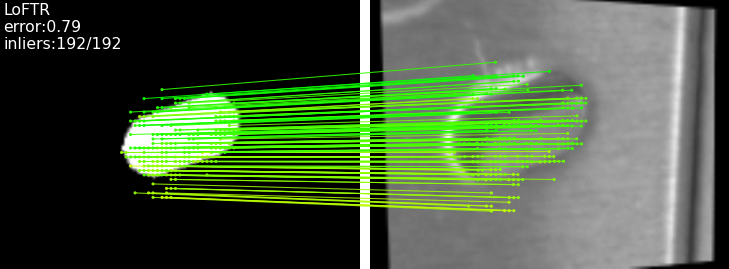}}

     \centerline{\includegraphics[width=\textwidth]{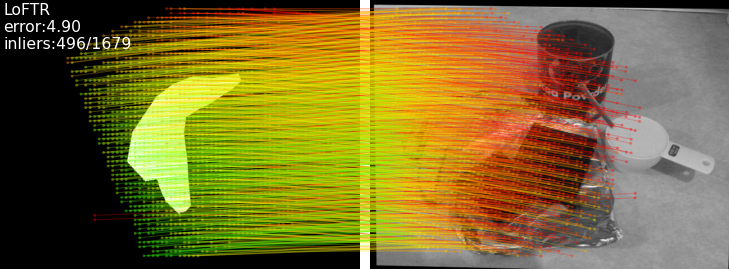}}
        
     \centerline{\includegraphics[width=\textwidth]{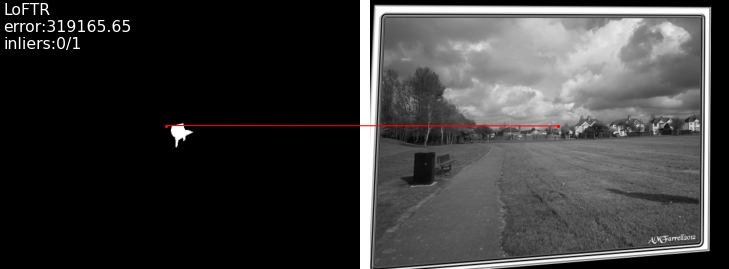}}
        
     \centerline{\includegraphics[width=\textwidth]{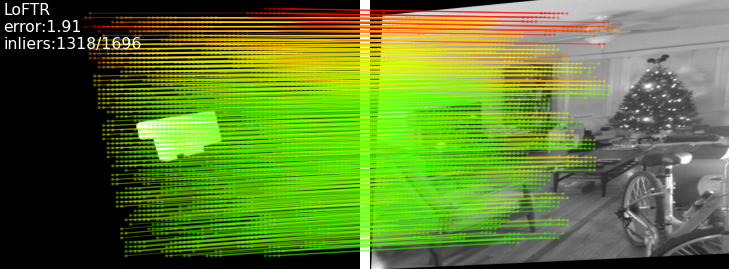}}
 
 \end{minipage}
    \begin{minipage}{0.24\linewidth}

        
     \centerline{\includegraphics[width=\textwidth]{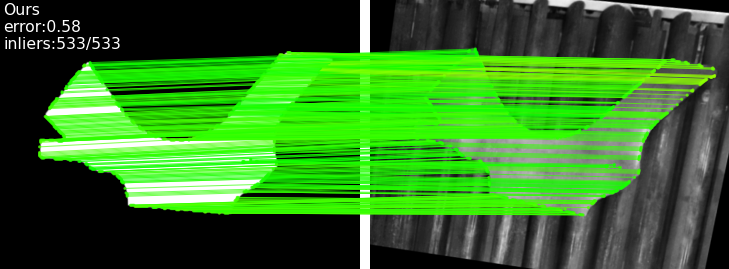}}
        
     \centerline{\includegraphics[width=\textwidth]{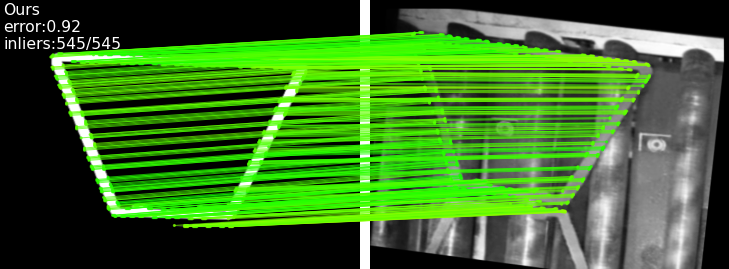}}
        
     \centerline{\includegraphics[width=\textwidth]{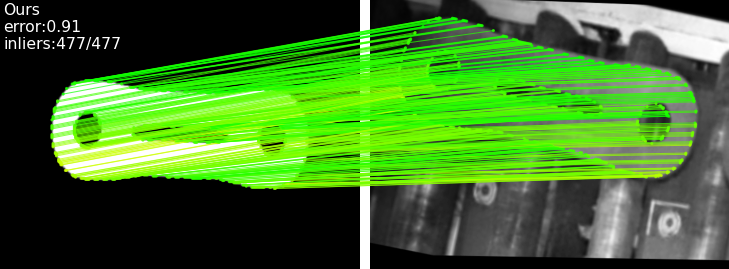}}
        
     \centerline{\includegraphics[width=\textwidth]{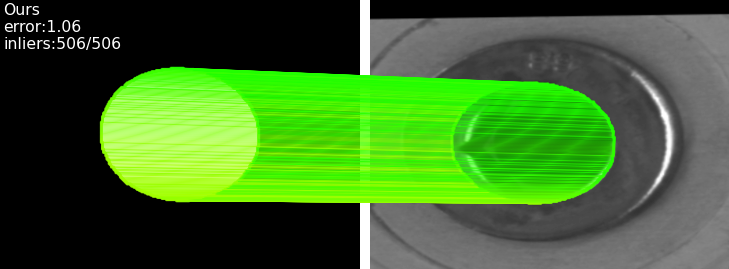}}
        
     \centerline{\includegraphics[width=\textwidth]{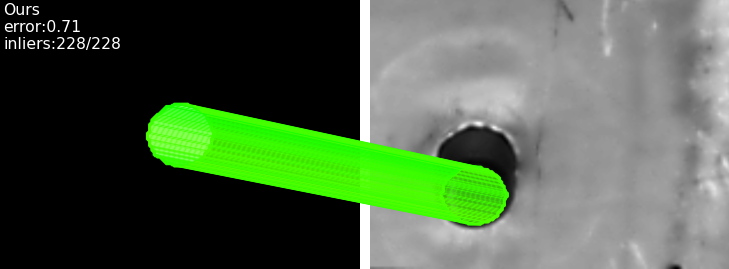}}
        
     \centerline{\includegraphics[width=\textwidth]{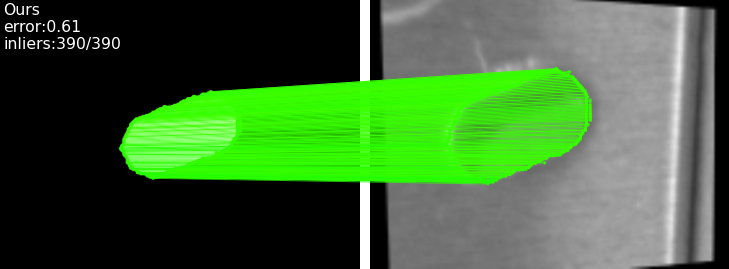}}
       
     \centerline{\includegraphics[width=\textwidth]{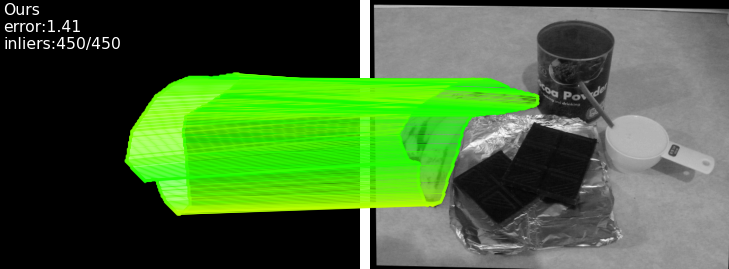}}
        
     \centerline{\includegraphics[width=\textwidth]{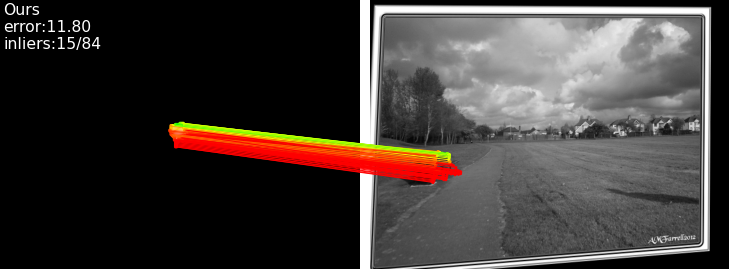}}
        
     \centerline{\includegraphics[width=\textwidth]{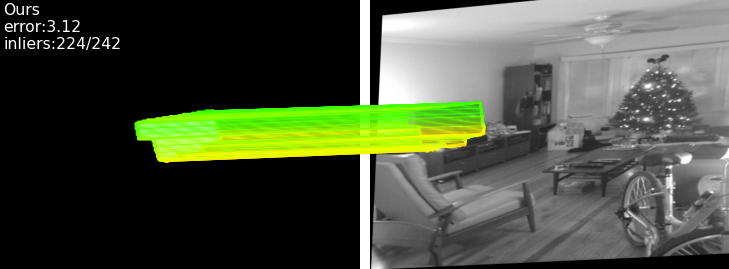}}
     
 \end{minipage}
 \end{center}
	\caption{Qualitative matching results for the three test datasets. Compared to SuperGlue, COTR and LoFTR, our method consistently obtains a higher inlier ratio, successfully coping with large viewpoint change, small objects and non-rigid deformation. Red indicates a reprojection error beyond 3 pixels for the Mechanical Parts and Assembly Holes datasets and 5 pixels for the COCO dataset. Further qualitative results can be found in the declarations~\ref{Declarations}.}
	\label{fig_matching}
\end{figure*}

For each match $(i,j')$ in $\mathcal{M}_f$, we use the inverse transformation of $H_c$ to warp $j'$ to its original position on image $I$. After coarse-to-fine refinement, the correspondences are accurate without obvious outliers (see the last column of Fig.~\ref{fig_matching} later). We obtain the final homography $H$ by wighted least squares using the DLT formulation, based on all matching pairs. The final homography $H$ indicates the transformation from the template $T$ to the source image $I$, precisely locating the template object.

\subsubsection{Fine-level training losses}
While training the fine-level module, the coarse-level module is fine-tuned at the same time. The training loss $\mathcal {L}$ is defined as $\mathcal {L}=\lambda \mathcal {L}_c + \mathcal {L}_f$.
In $ \mathcal {L}_f$, we use ground-truth supervision and self-supervision together for better robustness. 
For ground-truth supervision, we use the weighted loss function from~\cite{wang2020learning}. For self-supervision, we use  $L2$ similarity loss~\cite{lee2019image,shu2018unsupervised} to minimize the differences between local appearances of the warped image $I_w$ and template $T$. $ \mathcal {L}_f$ is formulated as:
\begin{equation*}
    \begin{aligned}
\mathcal{L}_{f} = & \frac{1}{\left|\mathcal{M}_f \right |}\sum_{(i,j^{'})\in\mathcal{M}_f}  \frac{1}{\sigma^{2}(i)} \|j^{'}-j^{'}_{gt}\|\ +\\
& \sum_{(i,j^{'})\in \mathcal{M}_f}(m_{i}* \left\| P_i^{T} -P_{j^{'}}^{I_{w}}  \right\|), 
\end{aligned}
\end{equation*}
where for each query point $i$,  $\sigma^{2}(i)$ is the total variance of the corresponding heatmap, $P_i^{T}$ denotes a local window cropped from template image $T$ with $i$ as the center, $m_i$ is a local area mask indicating presence of an edge pixel. Experiments on $L2$ similarity loss are presented in Sec.~\ref{sec:ablation}.

\section{Experiments}

After introducing the datasets used in our experiments (Sec. \ref{sec 5.1}) and implementation details (Sec. \ref{sec 5.2}), estimated homographies are compared for our proposed method and baselines  (Sec. \ref{sec 5.3}). Applications of our approach in industrial lines  are shown in Sec. \ref{sec:application}), while Sec. \ref{sec 5.5} considers the effectiveness of the components of our strategy. Further experimental details can be found in the {Appendices}.
 
\subsection{Datasets} \label{sec 5.1}

Here we outline the datasets used for testing.  Further details are given  in Appendix~\ref{appx-c}, to ensure reproducibility.

\subsubsection{Mechanical Parts} 
Obtaining poses of industrial parts is essential for robotic manipulator grasping on automated industrial lines. We collected a dataset based on  hundreds of varied planar mechanical parts.  To enrich the dataset while avoiding laborious annotation, we used GauGAN to generate an extra 40k pairs of matching data for training. The test dataset consisting of 800 samples was collected from an industrial workshop with human-labeled ground truth. It was used to quantitatively evaluate our method for single template and single object scenes, and to visually demonstrate the application of our approach to multi-template and multi-object scenes in Sec. \ref{sec:application}.

\begin{figure*}[!t]
 \begin{center}

 ~~~~~~~Template~\qquad ~~Image\qquad~ Linemod-2D ~~~~ SuperGlue\qquad~ COTR\qquad~~~~ LoFTR\qquad~~~~~ Ours \qquad~ Ground-truth \\

\begin{minipage}{0.015\linewidth}
    \rotatebox{90}{{~~~~~~~~~~Mechanical Parts dataset}}

    \rotatebox{90}{{~~~~~~~~~~~Assembly Holes dataset~~~~~~~~~~~}}

    \rotatebox{90}{{~~~~~~COCO dataset~~~~~~~~~~~}}

    \end{minipage}
\begin{minipage}{0.118\linewidth}

     \centerline{\includegraphics[width=\textwidth]{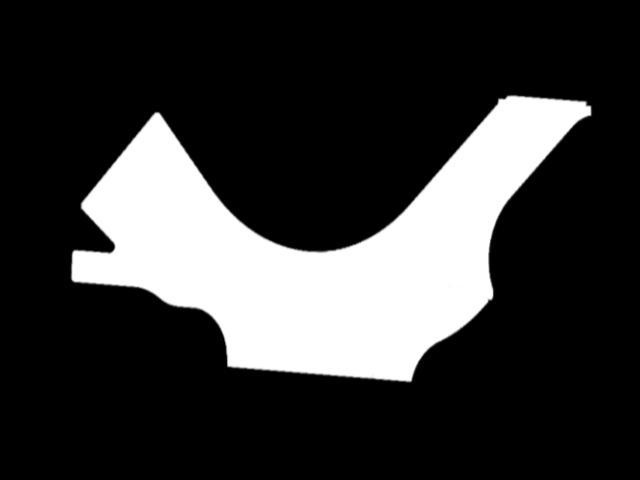}}
      
     \centerline{\includegraphics[width=\textwidth]{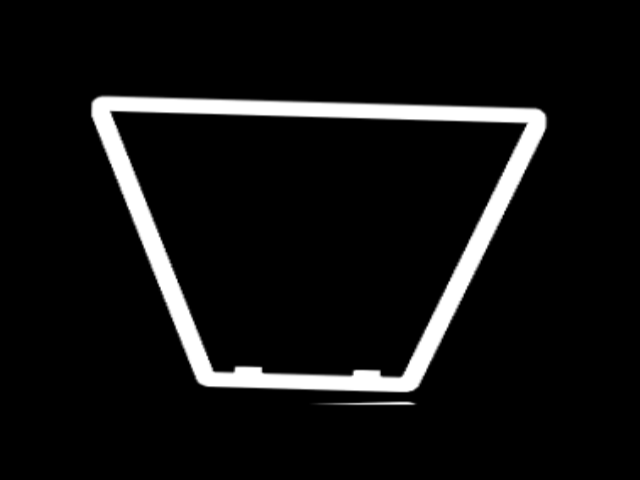}}
     
     \centerline{\includegraphics[width=\textwidth]{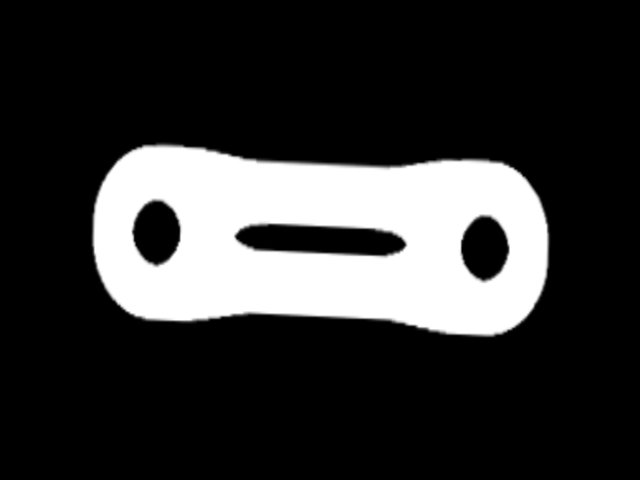}}

     \centerline{\includegraphics[width=\textwidth]{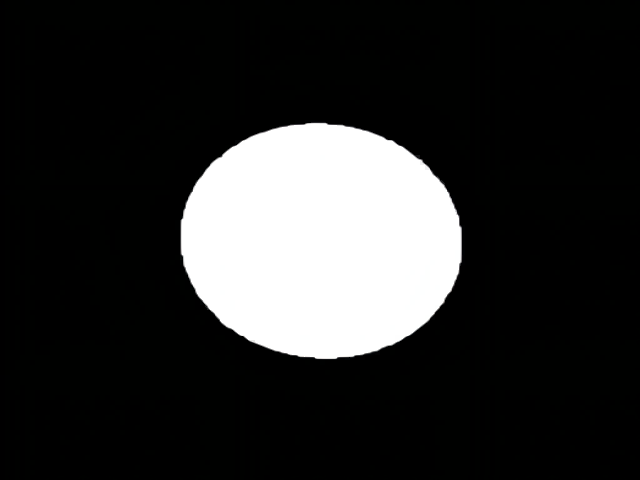}}
    
     \centerline{\includegraphics[width=\textwidth]{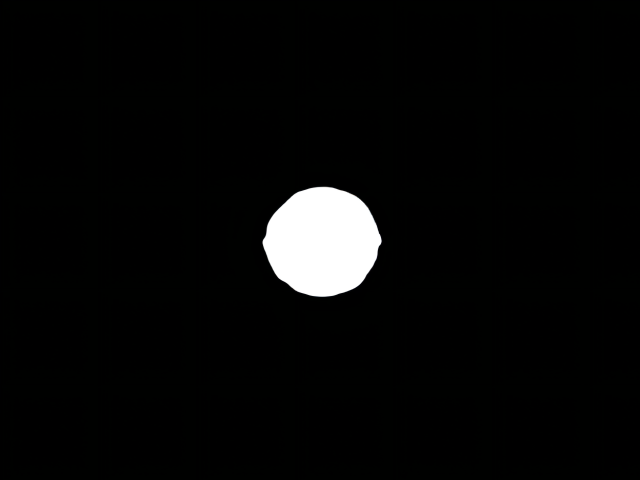}}
     
     \centerline{\includegraphics[width=\textwidth]{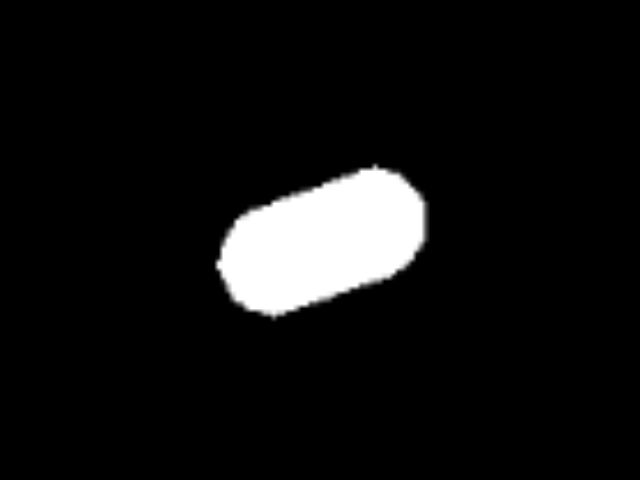}}

     \centerline{\includegraphics[width=\textwidth]{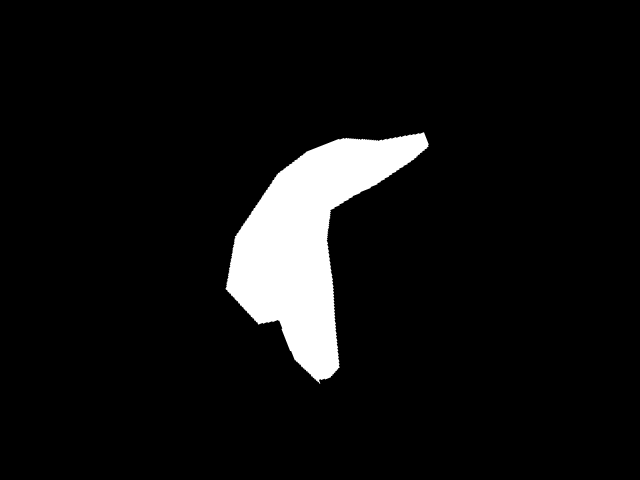}}
      
     \centerline{\includegraphics[width=\textwidth]{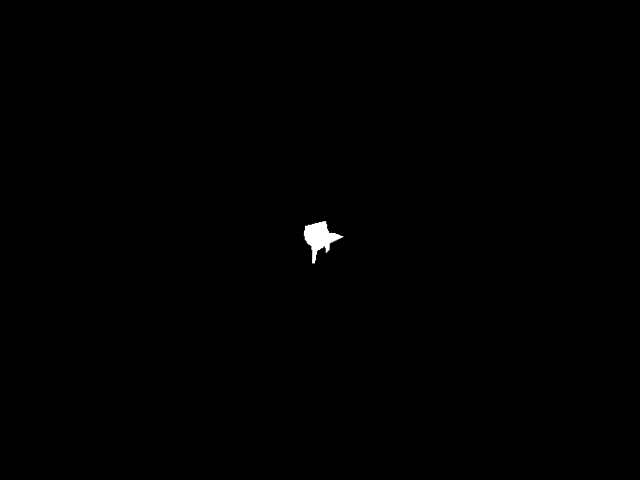}}
      
     \centerline{\includegraphics[width=\textwidth]{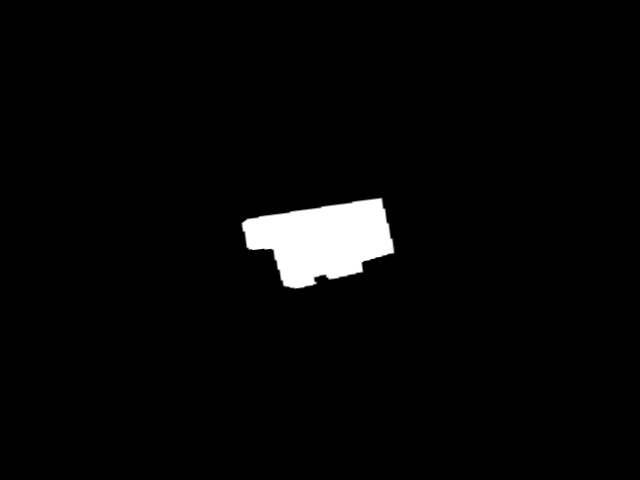}}
     
 \end{minipage}
 \begin{minipage}{0.118\linewidth}
   
    %
     \centerline{\includegraphics[width=\textwidth]{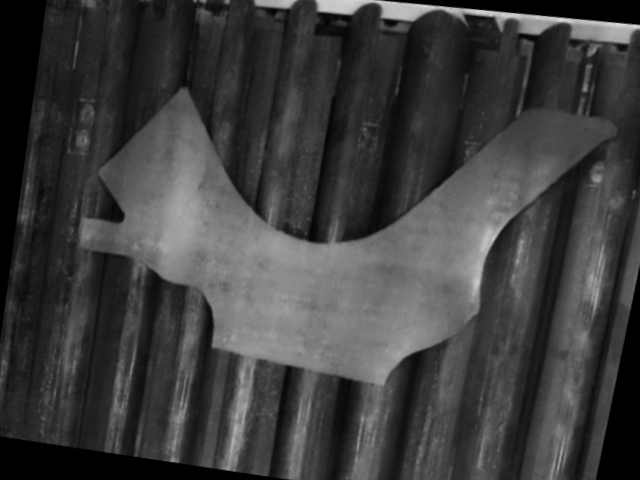}}
        
     \centerline{\includegraphics[width=\textwidth]{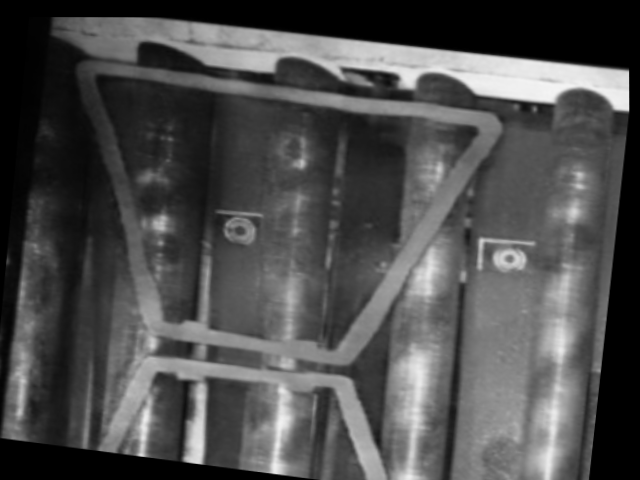}}
        
     \centerline{\includegraphics[width=\textwidth]{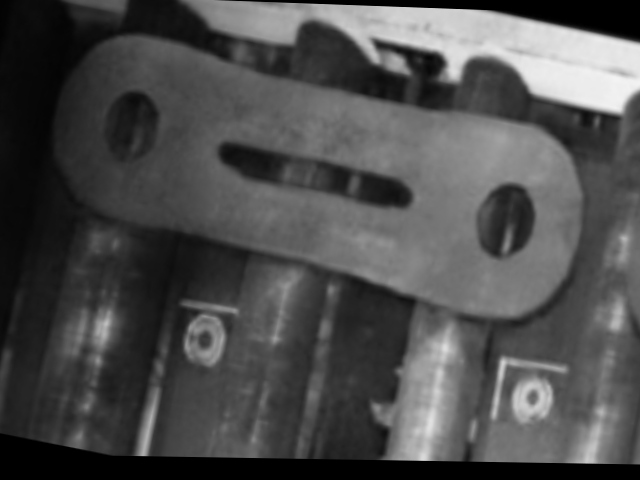}}

     \centerline{\includegraphics[width=\textwidth]{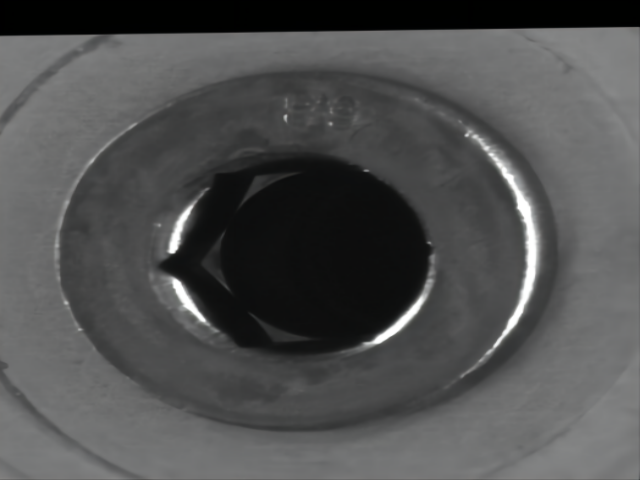}}
        
     \centerline{\includegraphics[width=\textwidth]{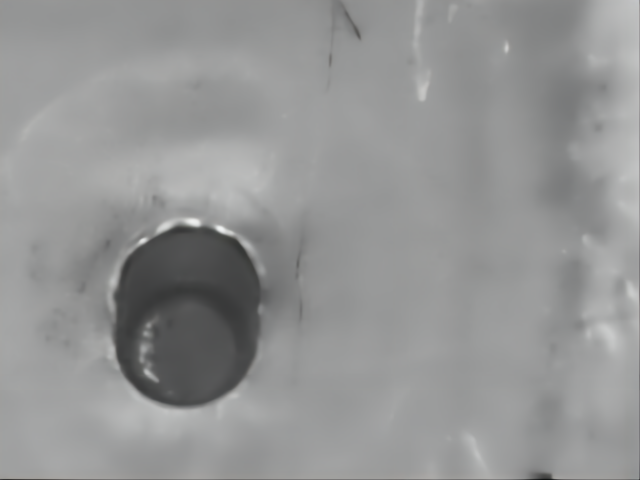}}
        
     \centerline{\includegraphics[width=\textwidth]{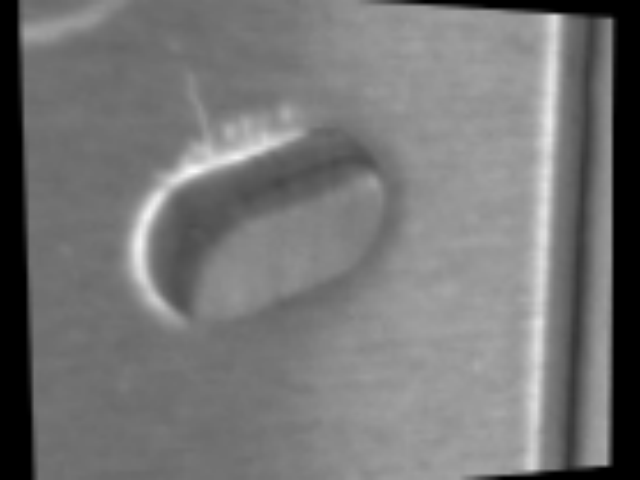}}
       
     \centerline{\includegraphics[width=\textwidth]{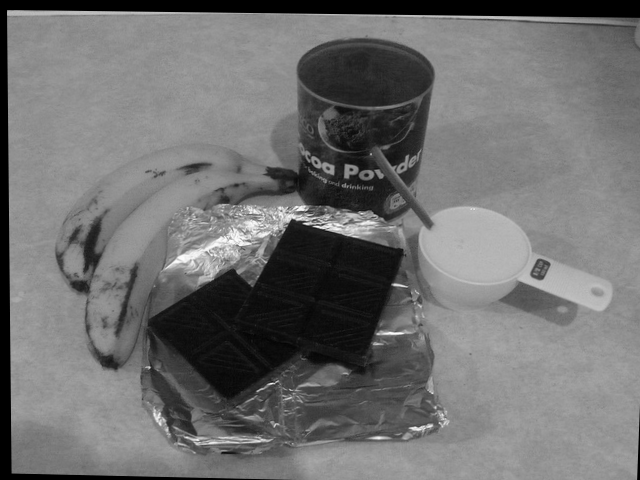}}
        
     \centerline{\includegraphics[width=\textwidth]{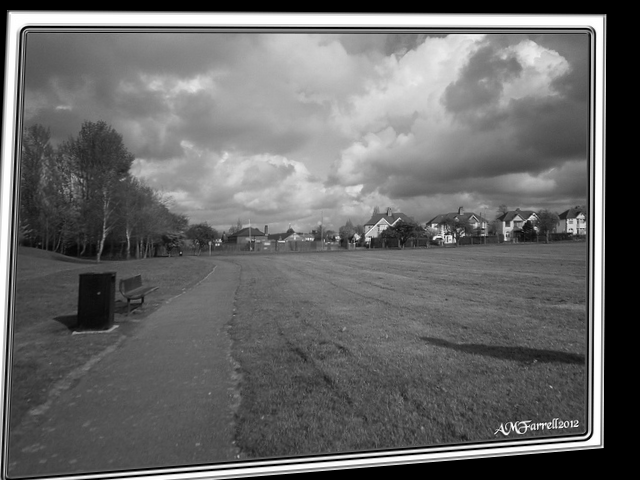}}
        
     \centerline{\includegraphics[width=\textwidth]{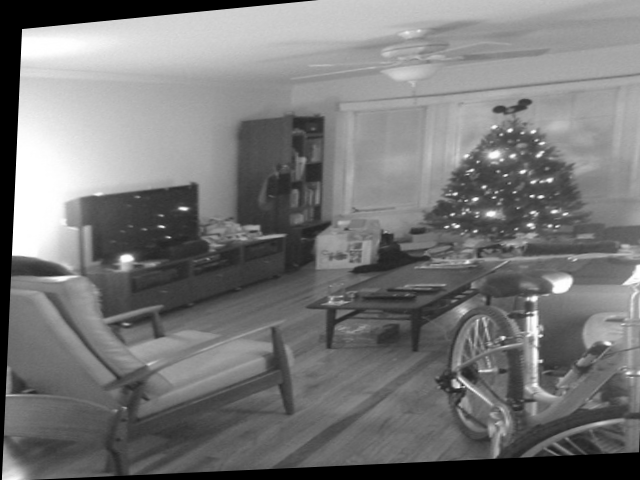}}
     
 \end{minipage}
        \begin{minipage}{0.118\linewidth}
 
     \centerline{\includegraphics[width=\textwidth]{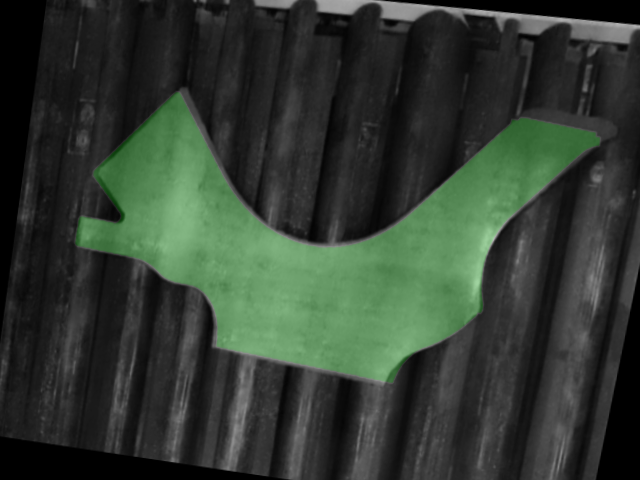}}
       
     \centerline{\includegraphics[width=\textwidth]{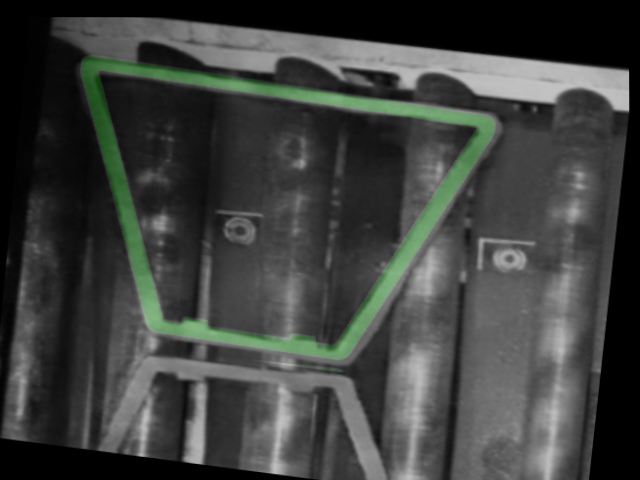}}
       
     \centerline{\includegraphics[width=\textwidth]{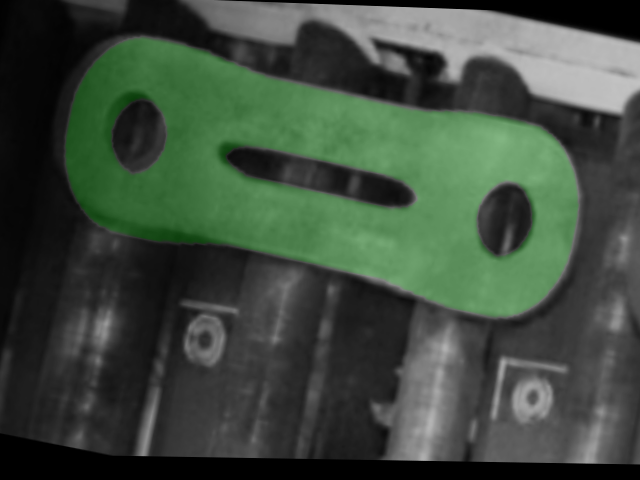}}

     \centerline{\includegraphics[width=\textwidth]{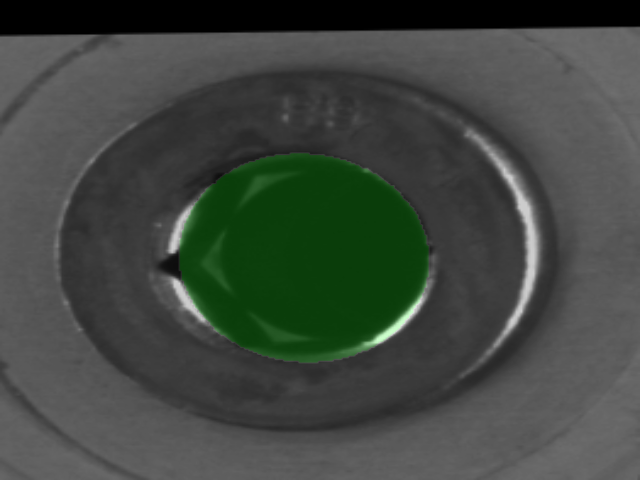}}
       
     \centerline{\includegraphics[width=\textwidth]{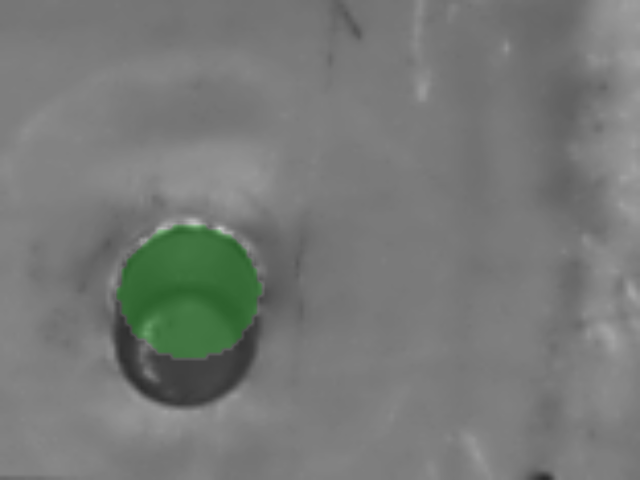}}
       
     \centerline{\includegraphics[width=\textwidth]{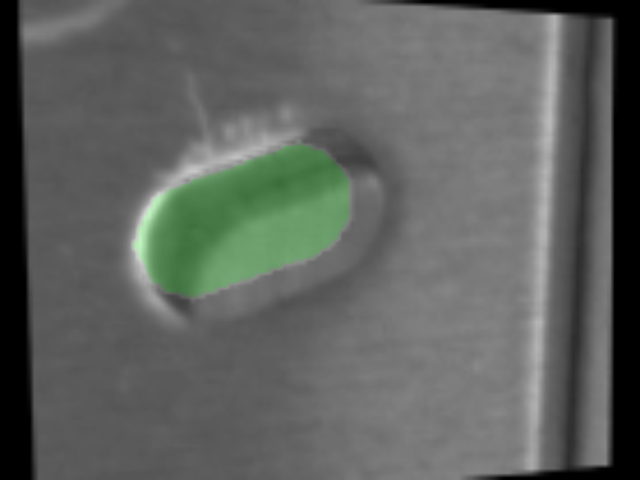}}

     \centerline{\includegraphics[width=\textwidth]{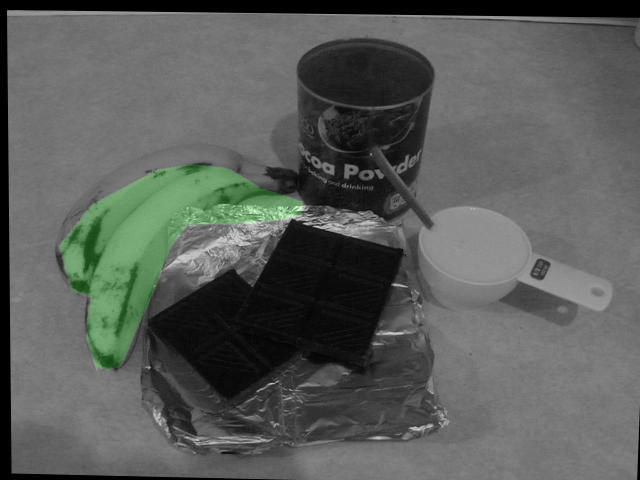}}
       
     \centerline{\includegraphics[width=\textwidth]{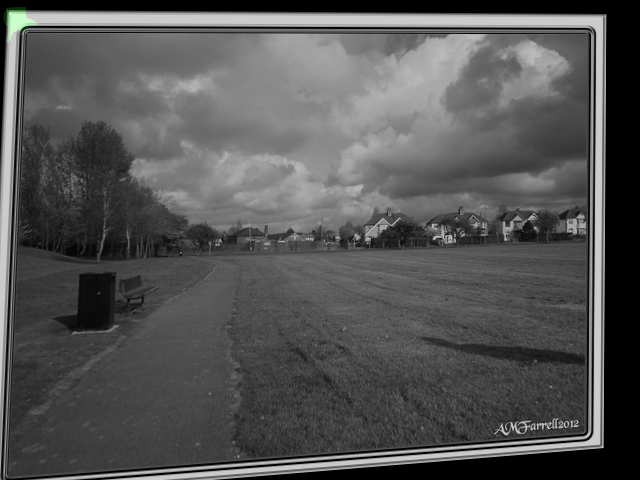}}
       
     \centerline{\includegraphics[width=\textwidth]{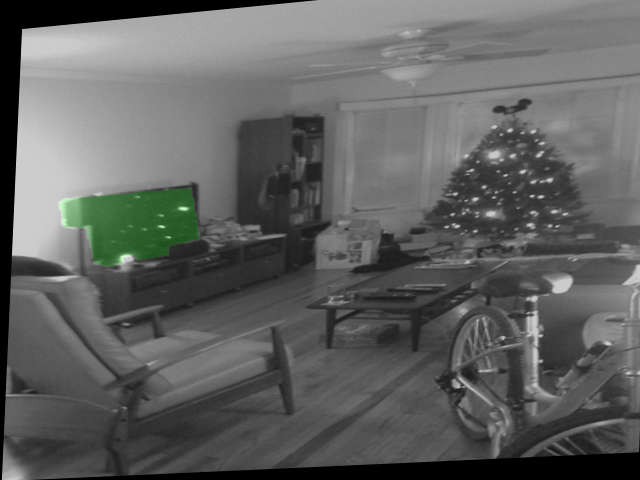}}
     
 \end{minipage}
    \begin{minipage}{0.118\linewidth}

    %
     \centerline{\includegraphics[width=\textwidth]{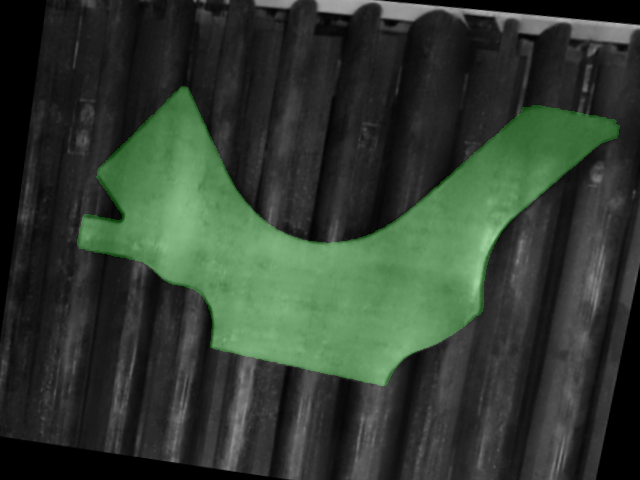}}
       
     \centerline{\includegraphics[width=\textwidth]{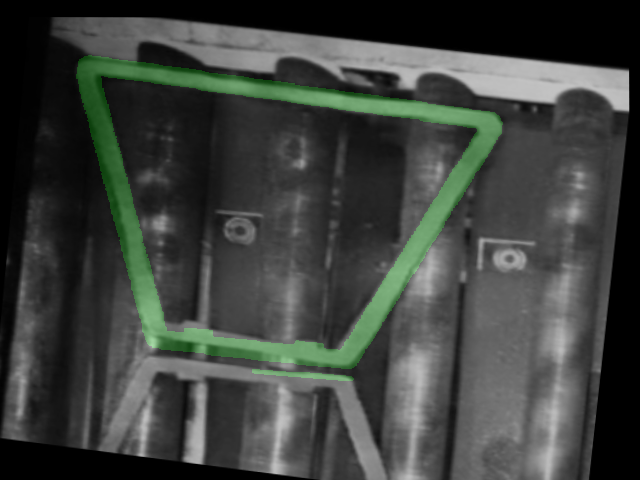}}
       
     \centerline{\includegraphics[width=\textwidth]{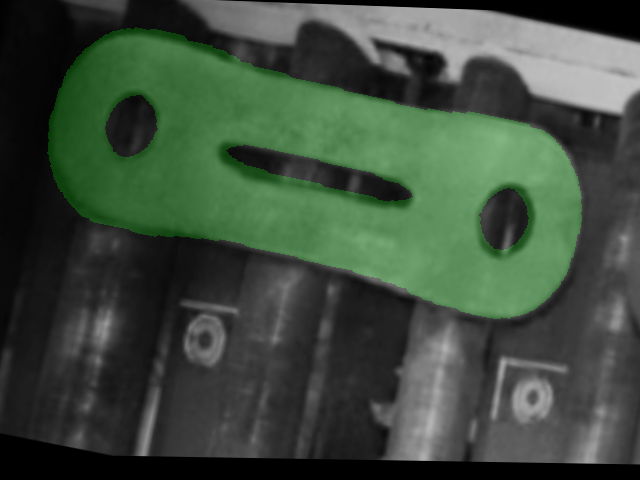}}

     \centerline{\includegraphics[width=\textwidth]{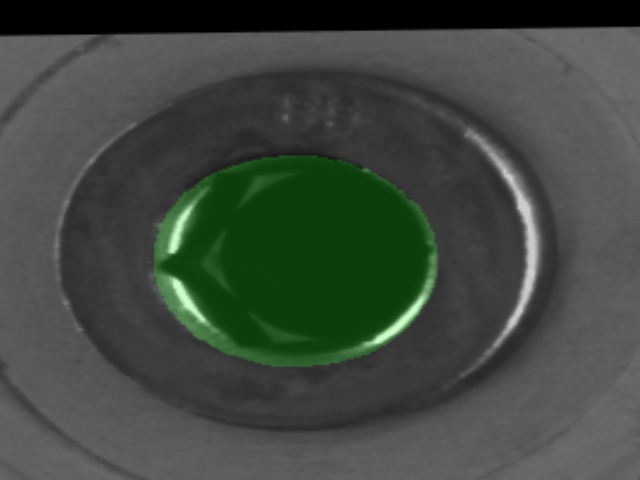}}
       
     \centerline{\includegraphics[width=\textwidth]{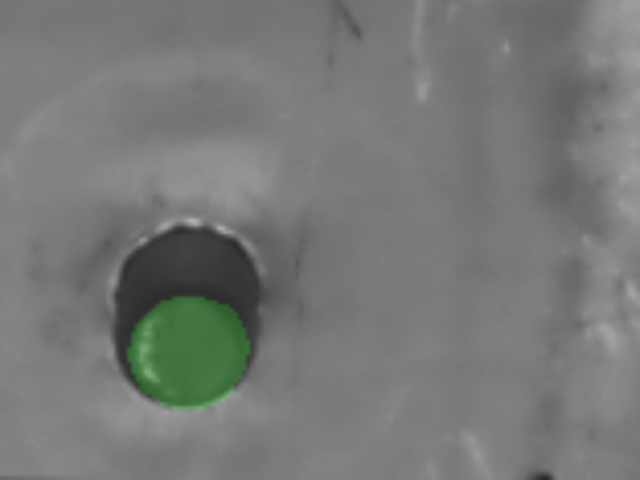}}
       
     \centerline{\includegraphics[width=\textwidth]{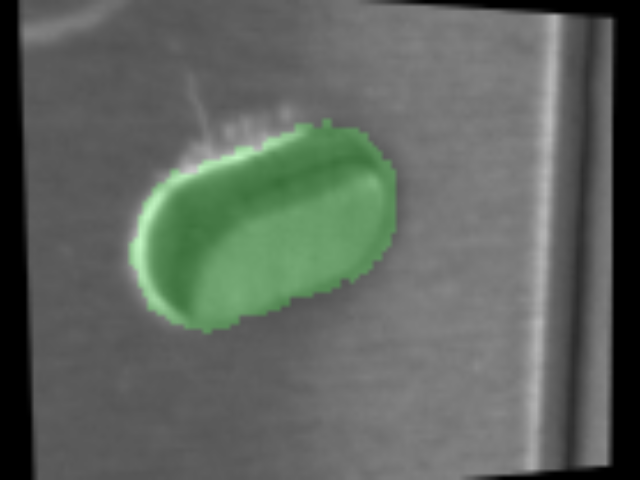}}

     \centerline{\includegraphics[width=\textwidth]{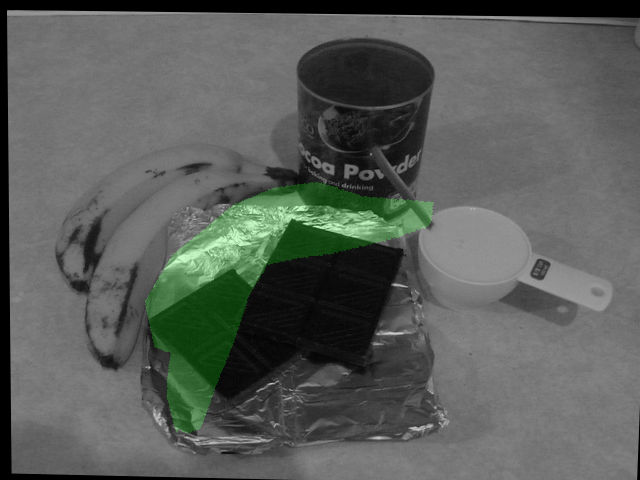}}
       
     \centerline{\includegraphics[width=\textwidth]{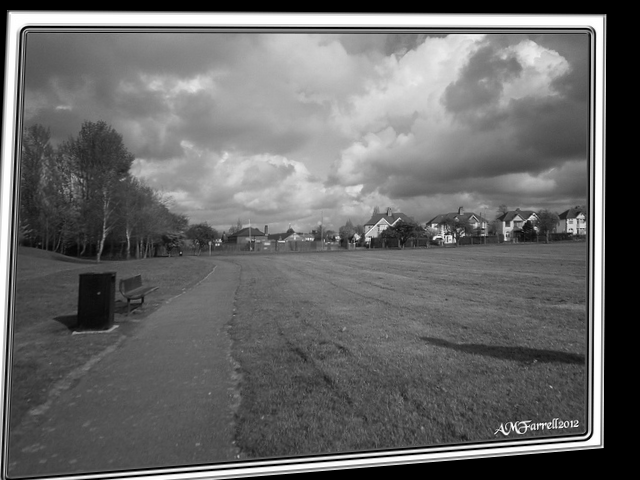}}
       
     \centerline{\includegraphics[width=\textwidth]{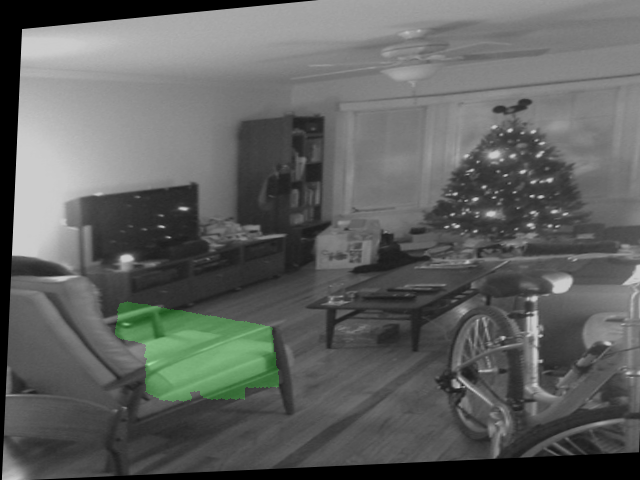}}
 \end{minipage}
  \begin{minipage}{0.118\linewidth}
        
     \centerline{\includegraphics[width=\textwidth]{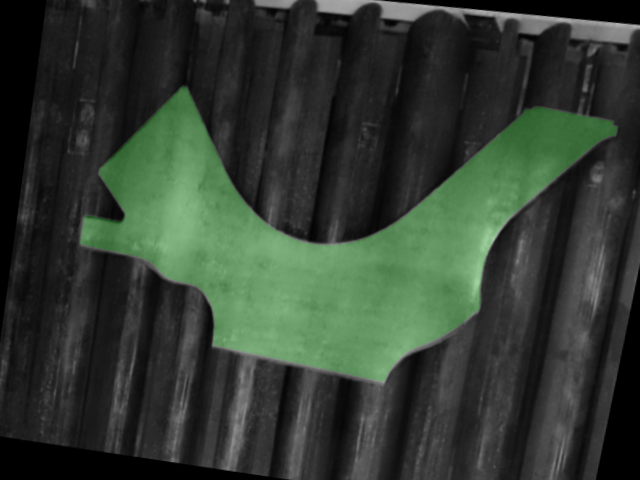}}
       
     \centerline{\includegraphics[width=\textwidth]{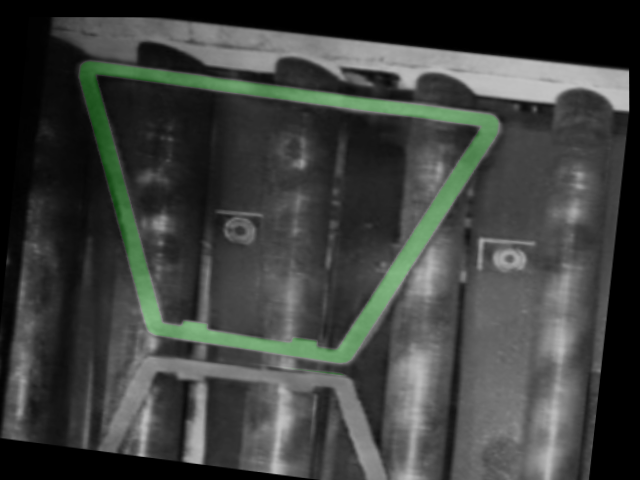}}
       
     \centerline{\includegraphics[width=\textwidth]{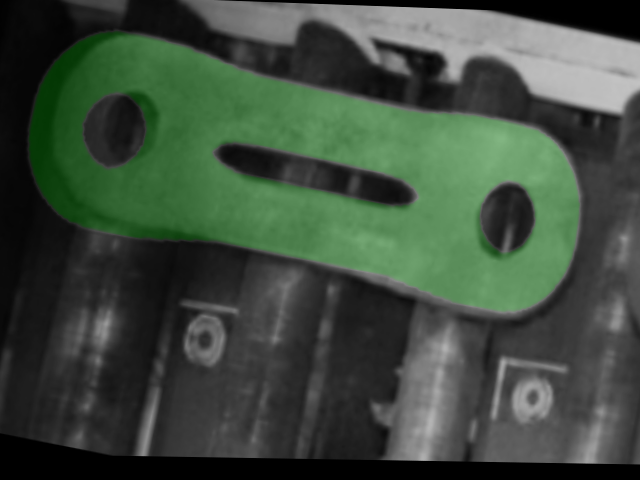}}

     \centerline{\includegraphics[width=\textwidth]{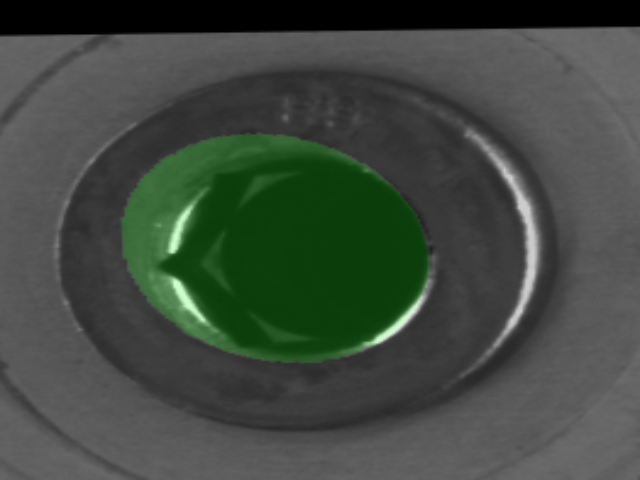}}
       
     \centerline{\includegraphics[width=\textwidth]{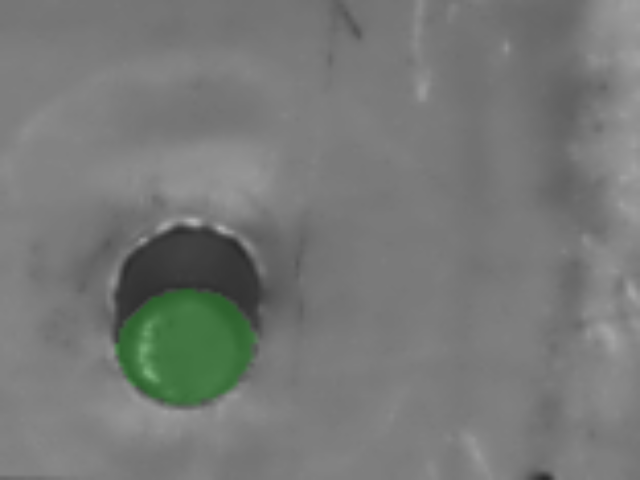}}
       
     \centerline{\includegraphics[width=\textwidth]{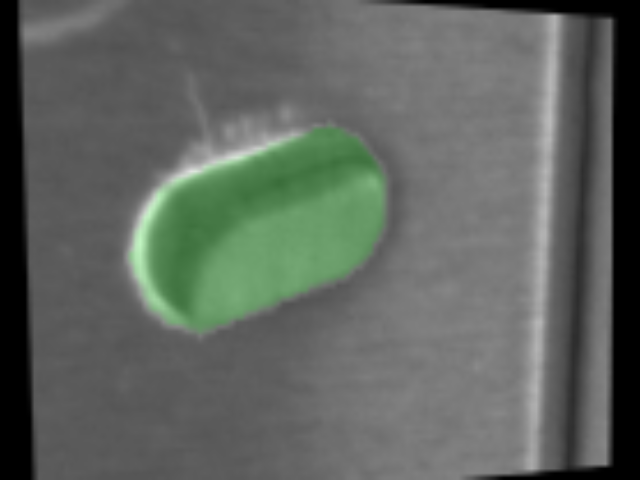}}

     \centerline{\includegraphics[width=\textwidth]{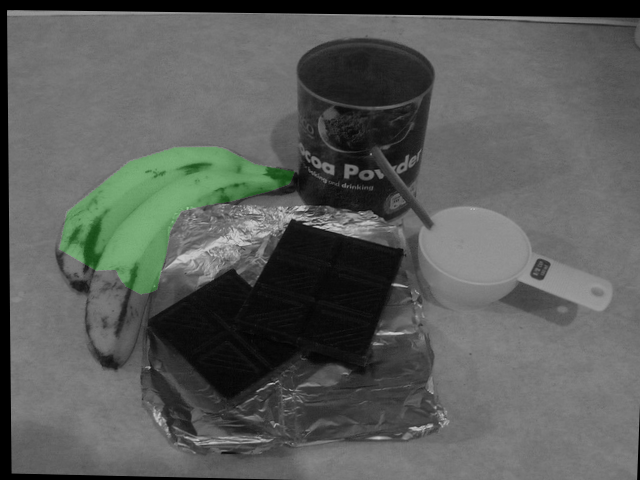}}
       
     \centerline{\includegraphics[width=\textwidth]{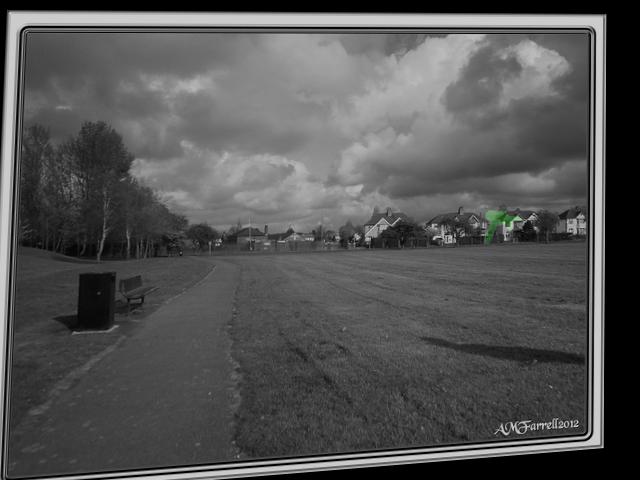}}
       
     \centerline{\includegraphics[width=\textwidth]{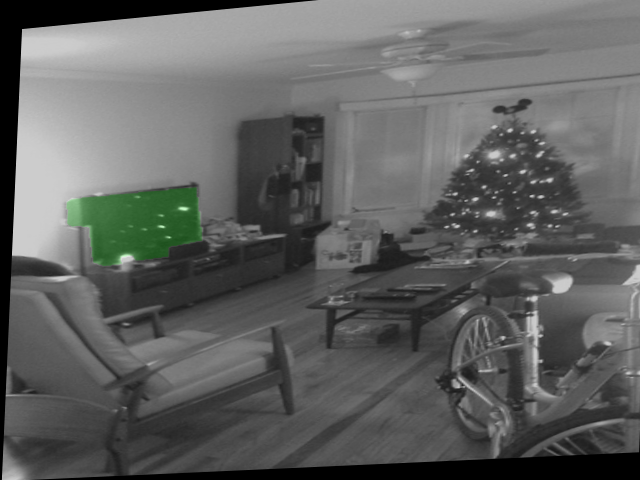}}
     
 \end{minipage}
  \begin{minipage}{0.118\linewidth}
           
     \centerline{\includegraphics[width=\textwidth]{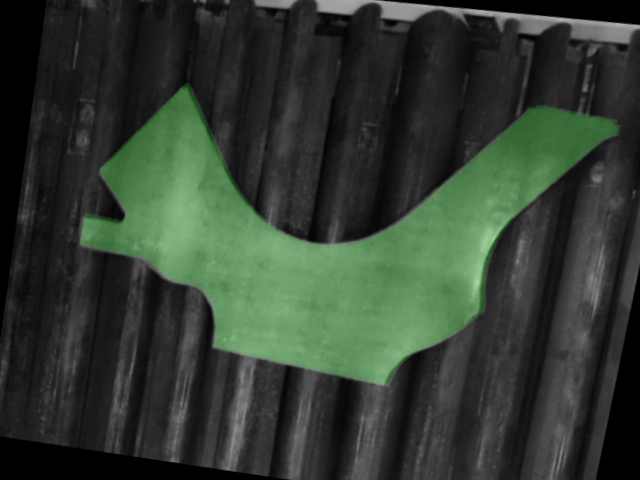}}
       
     \centerline{\includegraphics[width=\textwidth]{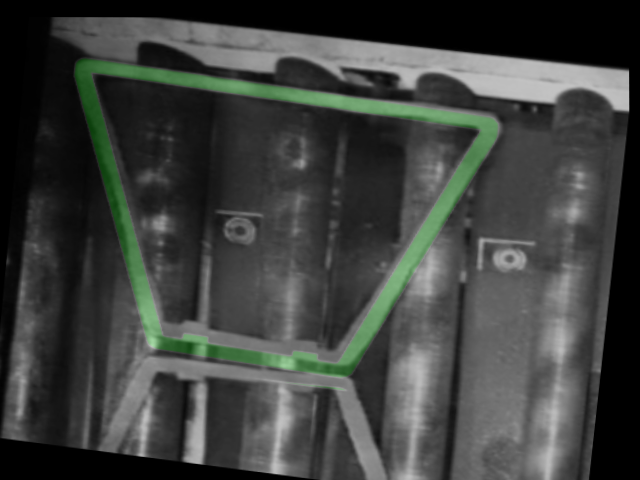}}
       
     \centerline{\includegraphics[width=\textwidth]{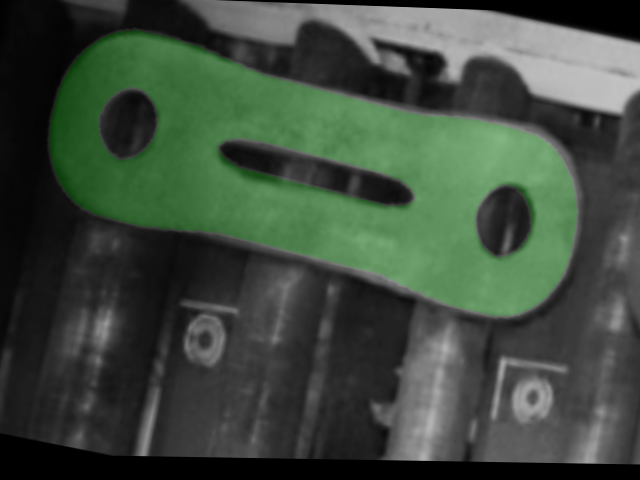}}

     \centerline{\includegraphics[width=\textwidth]{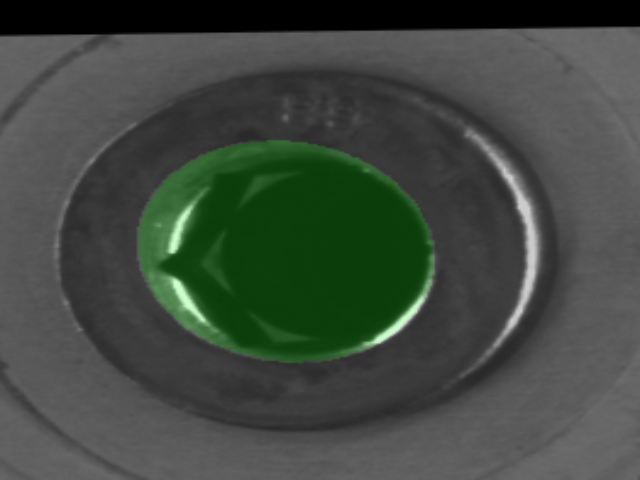}}
       
     \centerline{\includegraphics[width=\textwidth]{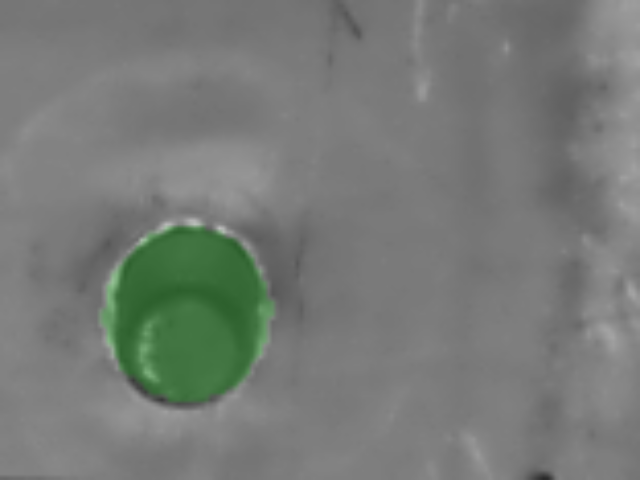}}
      
     \centerline{\includegraphics[width=\textwidth]{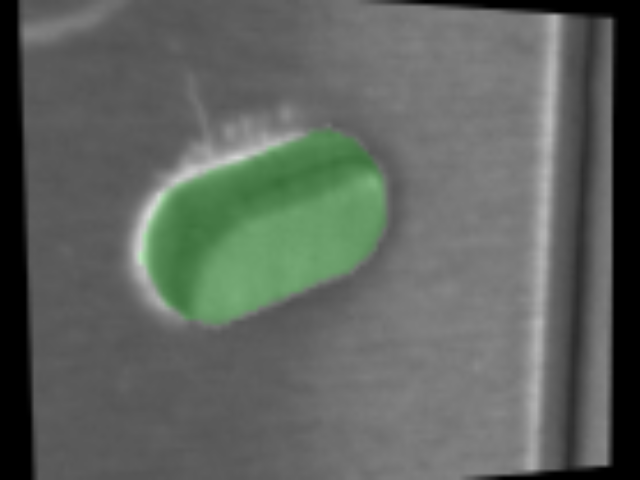}}

     \centerline{\includegraphics[width=\textwidth]{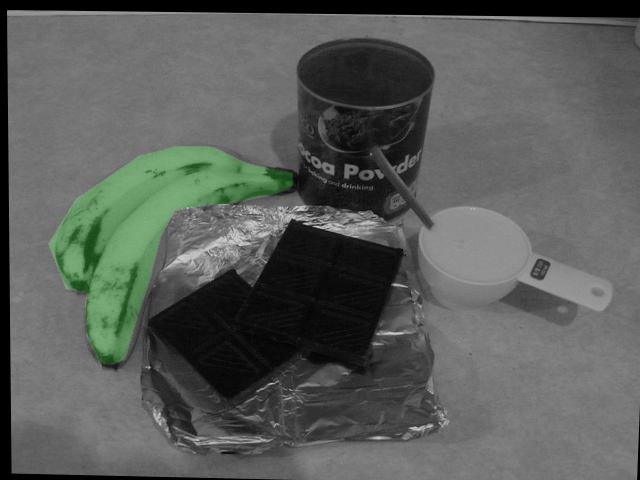}}
       
     \centerline{\includegraphics[width=\textwidth]{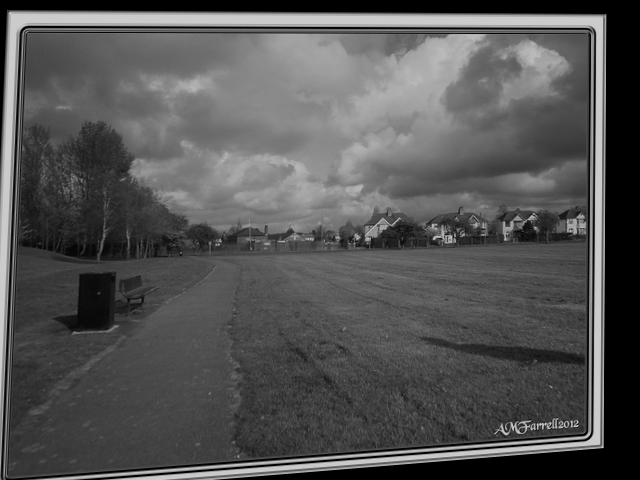}}
       
     \centerline{\includegraphics[width=\textwidth]{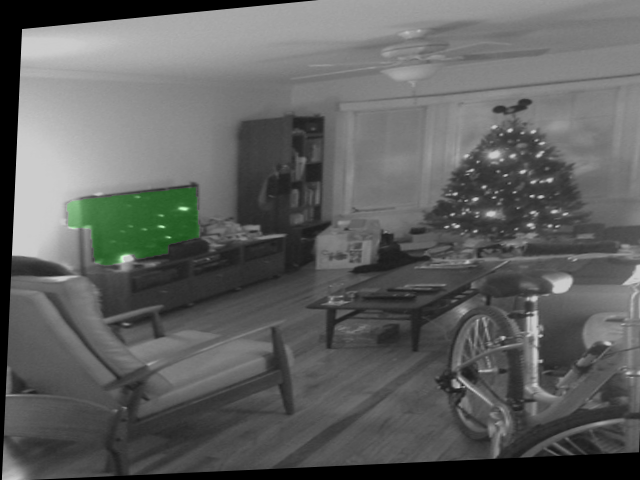}}
     
 \end{minipage}
\begin{minipage}{0.118\linewidth}
             
     \centerline{\includegraphics[width=\textwidth]{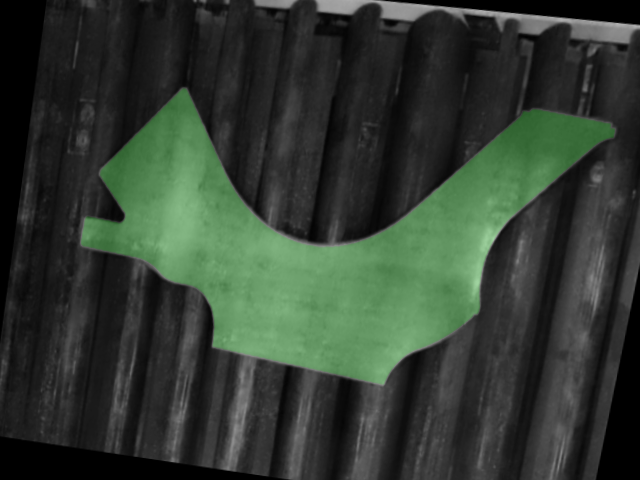}}
       
     \centerline{\includegraphics[width=\textwidth]{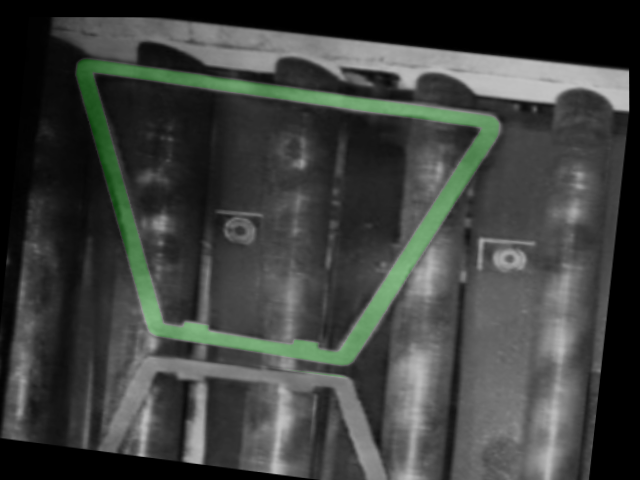}}
       
     \centerline{\includegraphics[width=\textwidth]{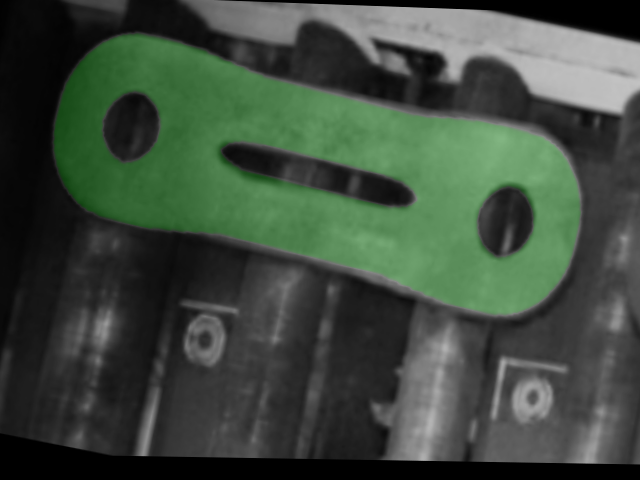}}

     \centerline{\includegraphics[width=\textwidth]{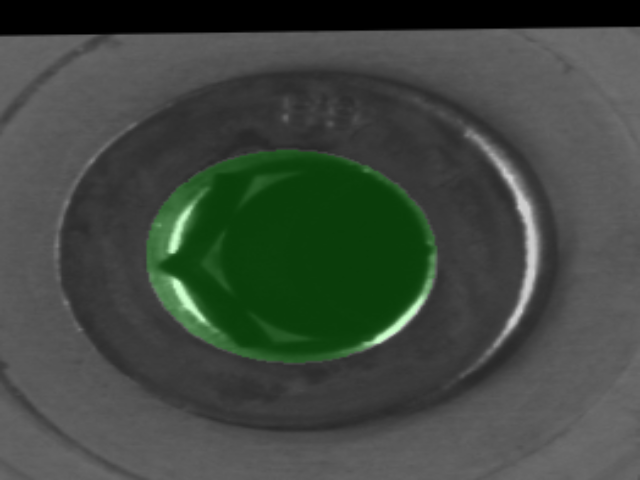}}
       
     \centerline{\includegraphics[width=\textwidth]{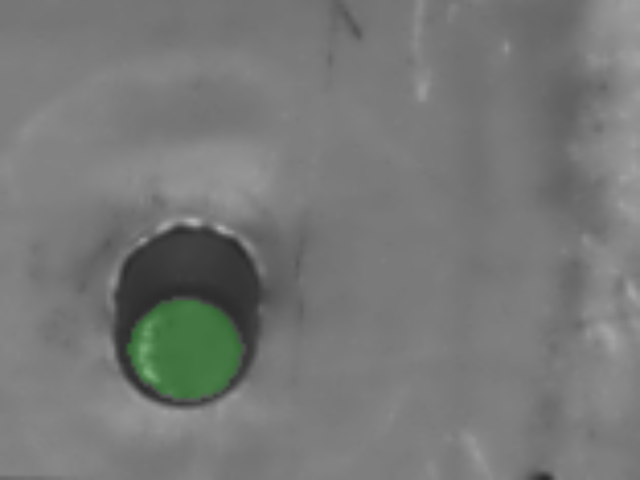}}
      
     \centerline{\includegraphics[width=\textwidth]{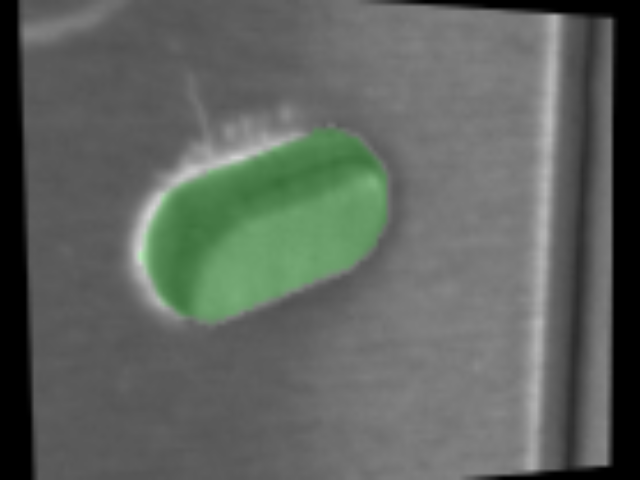}}

     \centerline{\includegraphics[width=\textwidth]{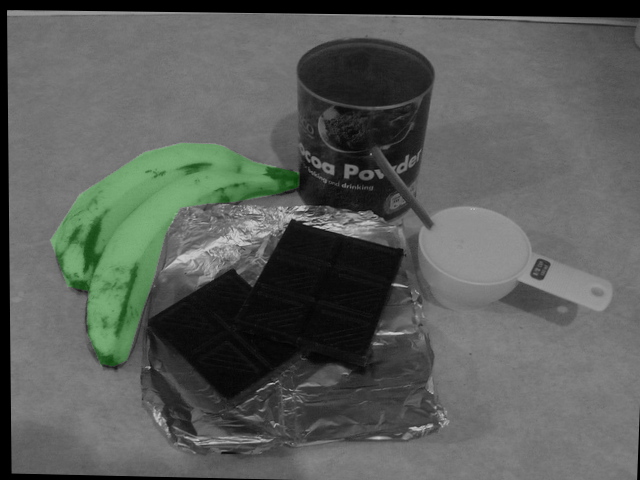}}
       
     \centerline{\includegraphics[width=\textwidth]{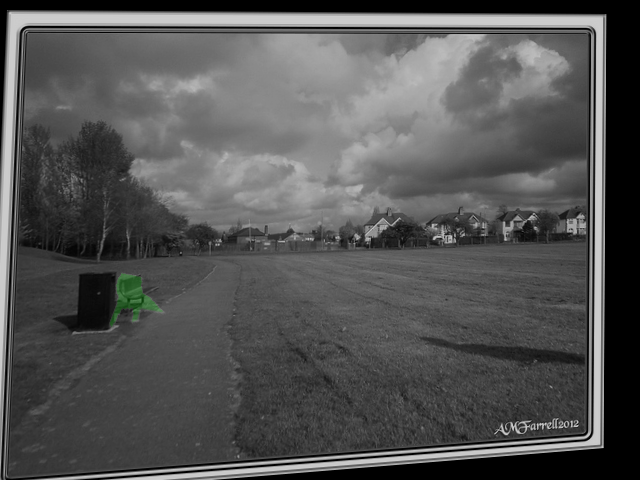}}
       
     \centerline{\includegraphics[width=\textwidth]{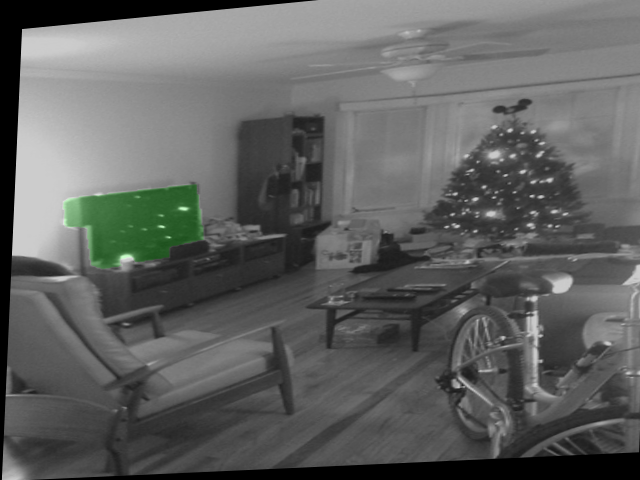}}
 \end{minipage}
\begin{minipage}{0.118\linewidth}
      
     \centerline{\includegraphics[width=\textwidth]{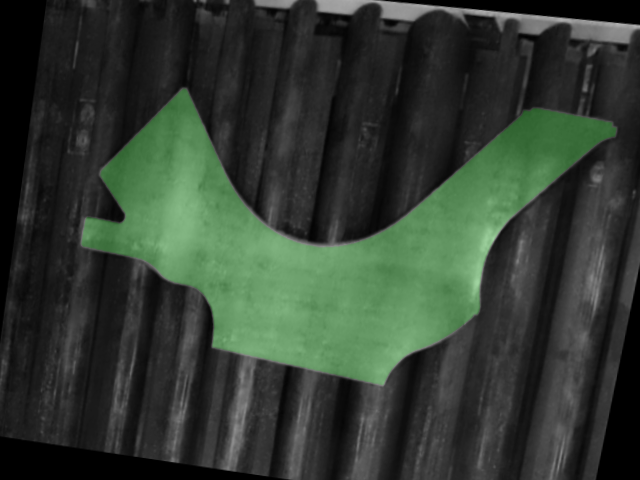}}
       
     \centerline{\includegraphics[width=\textwidth]{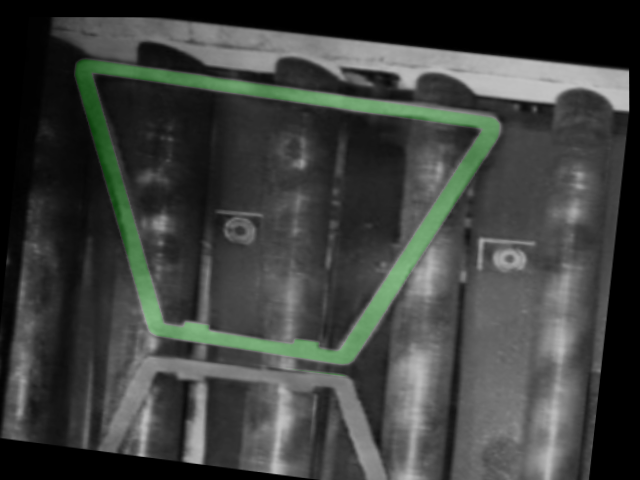}}
       
     \centerline{\includegraphics[width=\textwidth]{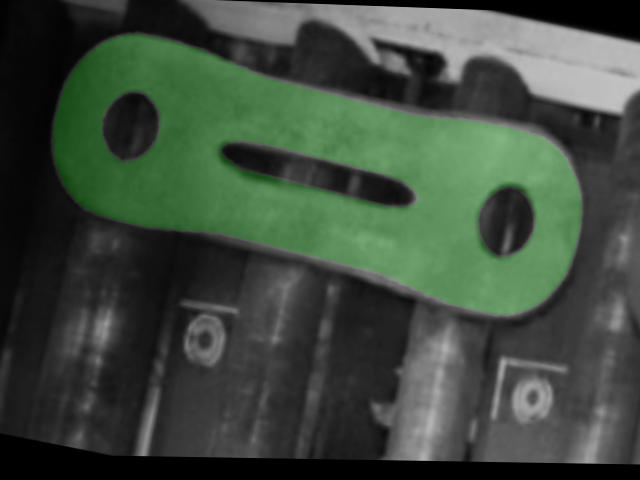}}

     \centerline{\includegraphics[width=\textwidth]{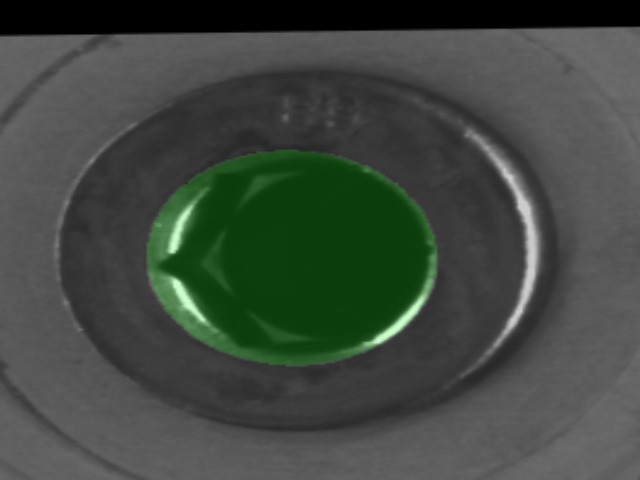}}
       
     \centerline{\includegraphics[width=\textwidth]{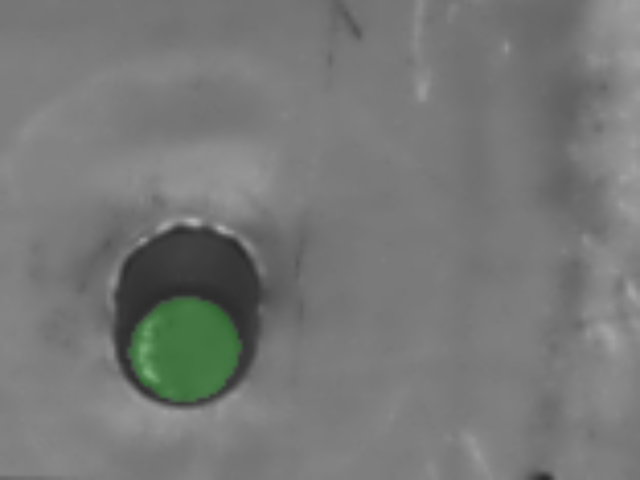}}
       
     \centerline{\includegraphics[width=\textwidth]{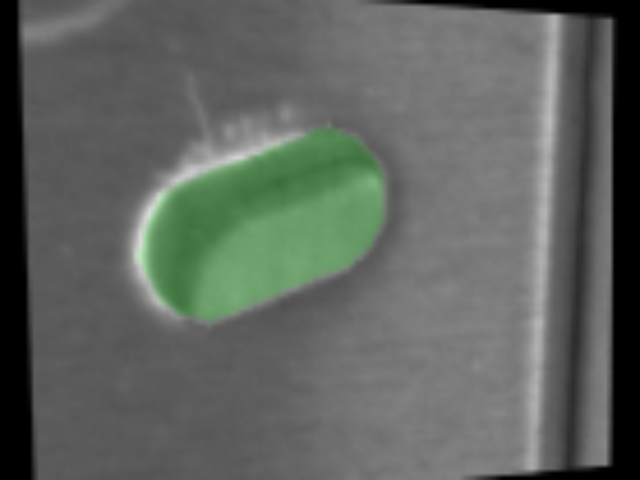}}
        
     \centerline{\includegraphics[width=\textwidth]{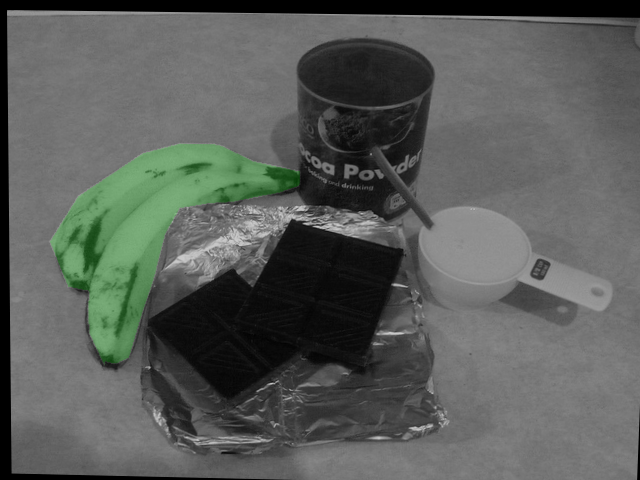}}
       
     \centerline{\includegraphics[width=\textwidth]{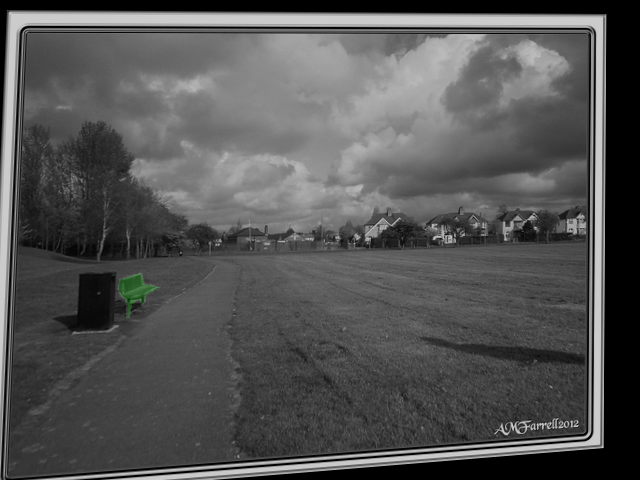}}
       
     \centerline{\includegraphics[width=\textwidth]{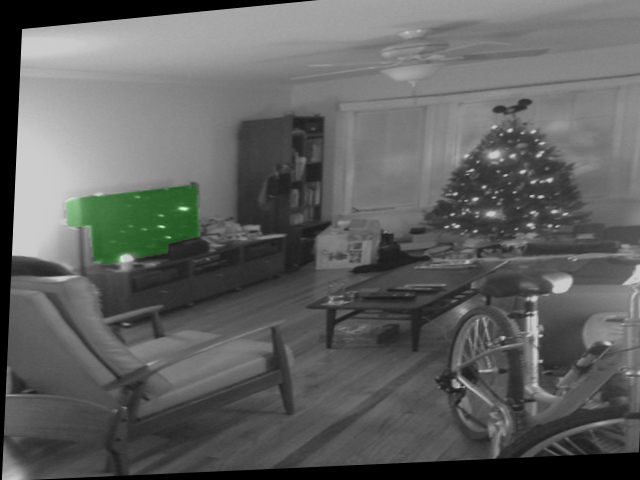}}
     
 \end{minipage}

 \end{center}
	\caption{\textbf{Qualitative registration results for the three test datasets.} The green area represents template mask placed in the input image using the estimated homography. For Linemod-2D, we selected the template with the best match from the set of templates. MAGASAC was used for outlier rejection for SuperGlue,  COTR and LoFTR. }
	\label{fig_align}
\end{figure*}

\subsubsection{Assembly Holes} Locating and matching  assembly holes can help determine whether the product parts have been machined in the correct position. Thus, we collected data for dozens of different assembly holes in vehicle battery boxes, giving about 45k image pairs. Each sample contains a binarized template image, a gray image to be matched, and a human-labeled mask. To simulate a real industry scenario, we randomly scaled the template size and perturbed the image corners to simulate possible hole deformation or camera disturbance. We randomly selected 700 image pairs containing all hole types for testing, and the remainder for training and validation.

\subsubsection{COCO} 
Going beyond industrial scenarios, we also performed tests using the well-known computer vision dataset COCO~\cite{lin2014microsoft} that contains common objects in natural scenes. Since COCO was not devised for template matching, we generated the image and template pair by selecting one instance mask and applying various kinds of transformations, including scaling, rotation and corner perturbation. We randomly selected 50k and 500 images for training and testing from the COCO training and validation set, respectively.

\begin{table*}[t!]
\caption{Homography estimation on the Mechanical Parts dataset. The AUC of the measurement point error  is reported as a percentage.  {SuperGlue$--$ and SuperGlue use the pre-trained SuperPoint and our fine-tuned SuperPoint for keypoint detection, respectively.}
}
\label{tab:tabel_steel}
\begin{center}
\begin{tabular}{l l r r r}
\hline
\multirow{2}{*}{Category} &
\multirow{2}{*}{Method} & \multicolumn{3}{ c }{Homography est. AUC  $\uparrow$}\\
\cline{3-5} 

& & @3px & @5px & @10px \\
\hline
\multirow{2}{*}{Overall similarity measure}
&Linemod-2D &14.7&28.6&52.0 \\
&GHT &1.7&3.5&6.5 \\

\hline
\multirow{6}{*}{Keypoints+ MNN} 
&SURF  RANSAC  & 0.1 & 0.1 & 0.2 \\
&SURF + MAGSAC   & 0.1 & 0.3 &1.0 \\ 
&D2Net  + RANSAC  &{4.5}&{8.3}&{15.1}  \\
&{D2Net  + MAGSAC } &{20.6}&{36.5}&{58.2}  \\

&{ASLFeat  + RANSAC } &{7.5}&{14.3}&{26.5}  \\
&{ASLFeat  + MAGSAC } & {24.8}& {35.9}& {60.7}  \\

& {SuperPoint  + RANSAC } & {1.3}& {3.2}& {9.3}  \\
& {SuperPoint  + MAGSAC } & {12.0}& {25.7}& {50.1}  \\

\hline
\multirow{8}{*}{Learning matchers}

& {SuperGlue$--$ $\ast$ + RANSAC} & {18.4}& {35.2}& {60.3}  \\
& {SuperGlue$--$ $\ast$ + MAGSAC } & {18.7}& {35.7}& {61.2}  \\

& {SuperGlue  + RANSAC} & {34.5}& {55.4}& {76.6}  \\
& {SuperGlue    + MAGSAC } & {32.2}& {53.3}& {75.5}  \\

&COTR  + RANSAC  & 26.1 & 44.3 & 76.1 \\
&COTR  + MAGSAC   & 26.4 & 44.9 & 76.3 \\
&LoFTR  + RANSAC  & 40.0 & 60.9 &80.0 \\
&LoFTR  + MAGSAC   & 40.6 &61.4  & 80.2 \\

&\textbf{Ours} & \textbf{58.8} & \textbf{74.7} &\textbf{87.3}\\
\hline
\end{tabular}
\end{center}
\end{table*}

\subsection {Implementation Details}\label{sec 5.2}
For training and testing, all images were resized to $480\times 640$. We use Kornia~\cite{riba2020kornia} for homography warping in the coarse alignment stage. Parameters were set as follows: window size $w=8$, numbers of transformer layers: $N_c=4$ and $N_f=2$, match selection threshold $\sigma=0.2$, loss weight $\lambda=10$ is set to 10, maximum number of template patches $N_p=128$, spatial consistency distance parameter $\sigma_d=0.4$,  angular consistency parameter $\sigma_{\alpha}=1.0$, weight control parameter = 0.5, and the number of neighbors $k=3$.

\subsection{Evaluation} \label{sec 5.3}

\subsubsection{Evaluation Metrics}
Following~\cite{sarlin2020superglue,zhang2020content,sun2021loftr}, we compute the  reprojection error of specific measurement points between the images warped with the estimated and the ground-truth homography. We then report the area under the cumulative curve (AUC) up to  thresholds of [3, 5, 10] pixels for industrial datasets, and [5, 10, 20] pixels for the COCO dataset. To ensure a fair comparison, we sampled 20 points uniformly on each template boundary as measurement points for use throughout the experiments.

\subsubsection{Baselines} We compared our method to three kinds of methods, based on: (i) overall similarity measure-based template matching, including Linemod-2D and generalized Hough transform (GHT), which are  widely used for industrial scenes, (ii) keypoint detection with MNN search, including SURF, D2Net, ASLFeat  and SuperPoint,and  (iii) matching learning, including SuperGlue, COTR~\cite{jiang2021cotr} and LoFTR  (state-of-the-art feature matching networks). 

For overall similarity measure-based methods which cannot deal with perspective transformation, we apply a more tolerant evaluation strategy. Specifically, we generate templates at multiple scales (step size =  $0.01$) and orientations (step-size = $1^{\circ}$ for matching. We use the centroids of generated templates as measure points and select the template with the best score as the final result. For SURF, we use the PiDiNet edge detector to preprocess  the input images. In SuperGlue, we choose SuperPoint  for keypoint detection and descriptor extraction. All learning-based baselines were fine-tuned  on each dataset until convergence, based on the parameters of the source model. Further details of training setup are provided in Appendix~\ref{appx-b}.

We adopted RANSAC  and MAGSAC for outlier rejection for all correspondence-based baselines when estimating the homography transformation, following~\cite{zhang2020content}. Direct linear transformation (DLT)  is applied directly in a differentiable manner to our method,  assuming matches have high inlier rates and trustworthy confidence weights.

\begin{table*}[tp!]
\caption{Homography estimation on the Assembly Holes dataset. The AUC of the measurement point error  is reported as a percentage.  {SuperGlue$--$ and SuperGlue use the pre-trained SuperPoint and our fine-tuned SuperPoint for keypoint detection, respectively.} 
}
\label{tab:tabel_hole}
\begin{center}
\begin{tabular}{l l r r r}
\hline
\multirow{2}{*}{Category} &
\multirow{2}{*}{Method} & \multicolumn{3}{ c }{Homography est. AUC  $\uparrow$}\\
\cline{3-5} 

& & @3px & @5px & @10px \\
\hline
\multirow{2}{*}{Overall similarity measure}
&Linemod-2D  & 24.7 & 37.1 &53.2  \\
&GHT & 18.7 & 31.2 &49.3 \\

\hline
\multirow{6}{*}{Keypoints+ MNN}
&SURF + RANSAC  & 0.2 & 0.5 & 2.0  \\
&SURF + MAGSAC   & 0.8 & 2.1 & 7.5 \\ 
&ORB + RANSAC  & 0.2 & 0.5 & 2.0 \\
&ORB + MAGSAC & 0.5 & 1.0 & 2.7 \\ 

& {D2Net  + RANSAC } & {7.6}& {13.1}& {24.7}  \\
& {D2Net  + MAGSAC } & {19.9}& {31.8}& {49.1}  \\

& {ASLFeat  + RANSAC } & {16.4}& {28.2}& {40.3}  \\
& {ASLFeat  + MAGSAC } & {23.9}& {35.7}& {53.2}  \\

& {SuperPoint  + RANSAC } & {15.6}& {26.8}& {44.6}  \\
& {SuperPoint  + MAGSAC } & {17.2}& {31.1}& {52.0}  \\

\hline
\multirow{8}{*}{Learning matchers}

& {SuperGlue$--$    + RANSAC } & {15.1}& {26.2}& {43.6}  \\
& {SuperGlue$--$    + MAGSAC } & {16.8}& {27.9}& {44.7}  \\

& {SuperGlue    + RANSAC } & {41.6}& {58.9}& {76.4}  \\
& {SuperGlue    + MAGSAC } & {41.5}& {58.9}& {76.3}  \\

&COTR  + RANSAC  & 31.4 & 50.1 & 71.7 \\
&COTR  + MAGSAC   & 31.5 & 50.1 & 72.0 \\
&LoFTR  + RANSAC  & 54.3 & 68.8 &81.8 \\
&LoFTR  + MAGSAC   & 54.3 & 68.7 & 81.8 \\

&\textbf{Ours} & \textbf{69.1} & \textbf{81.0} &\textbf{90.4}\\
\hline
\end{tabular}
\end{center}
\end{table*}

\begin{figure*}[t!]

 \begin{center}
~~~~~~~~~~~\qquad SuperGlue~~~~~~~~~~~~~~\qquad\qquad COTR \qquad\qquad\qquad\qquad~~ LoFTR \qquad\qquad\qquad\qquad~~ Ours \qquad\qquad~~~~~~~\\
        \begin{minipage}{0.22\linewidth}
     \centerline{\includegraphics[width=\textwidth,height=0.65\textwidth]{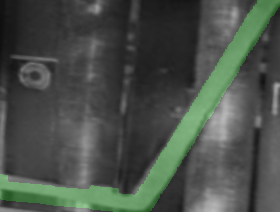}}
           
      \centerline{\includegraphics[width=\textwidth,height=0.65\textwidth]{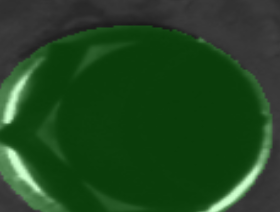}}
            
     \centerline{\includegraphics[width=\textwidth,height=0.65\textwidth]{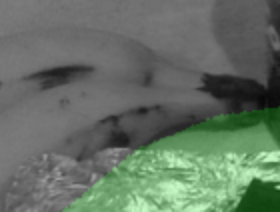}}
       
 \end{minipage}
    \begin{minipage}{0.22\linewidth}

     \centerline{\includegraphics[width=\textwidth,height=0.65\textwidth]{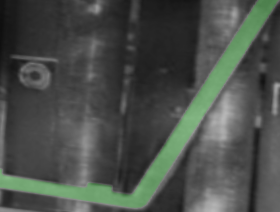}}
         
       \centerline{\includegraphics[width=\textwidth,height=0.65\textwidth]{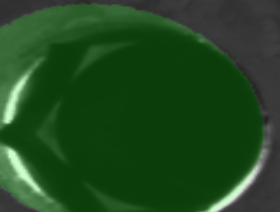}}
           
     \centerline{\includegraphics[width=\textwidth,height=0.65\textwidth]{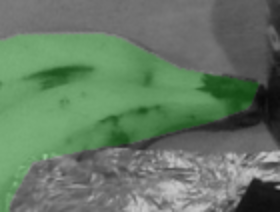}}
       
 \end{minipage}
    \begin{minipage}{0.22\linewidth}

       
     \centerline{\includegraphics[width=\textwidth,height=0.65\textwidth]{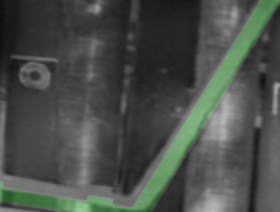}}
         
      \centerline{\includegraphics[width=\textwidth,height=0.65\textwidth]{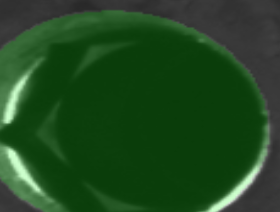}}
          
     \centerline{\includegraphics[width=\textwidth,height=0.65\textwidth]{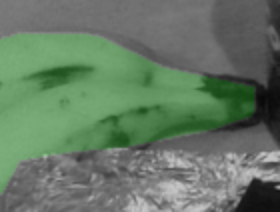}}
       
 \end{minipage}
    \begin{minipage}{0.22\linewidth}

         
     \centerline{\includegraphics[width=\textwidth,height=0.65\textwidth]{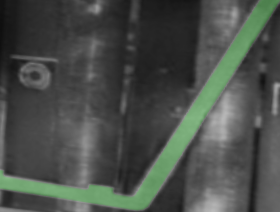}}
         
        \centerline{\includegraphics[width=\textwidth,height=0.65\textwidth]{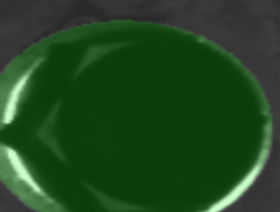}}
         
     \centerline{\includegraphics[width=\textwidth,height=0.65\textwidth]{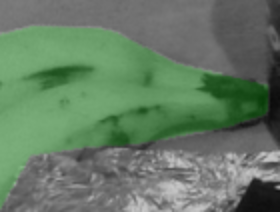}}
       
 \end{minipage}
 
 \end{center}
	\caption{ {Close-ups of registration results from SuperGlue, COTR, LoFTR and our method. Our method accurately focuses on the contours of objects.}}
	\label{zooming}
\end{figure*}

\subsubsection{Qualitative Comparison} 
We provide qualitative results in Figs.~\ref{fig_matching} and~\ref{fig_align}. In both figures, the first three rows use the Mechanical Parts dataset, the next three, the Assembly Holes dataset, and the last three, COCO. {Fig.~\ref{fig_matching} shows that, compared to SuperGlue, COTR  and LoFTR, the correspondences of our method are more accurate and reliable. While the correspondences predicted by SuperGlue  and COTR, like ours, lie on the contour of the object, they contain more outliers.}  LoFTR  yields more correspondences even in the blank area. However, these matching pairs tend to become inaccurate when further from the object. Instead, our method  effectively uses contour information by focusing the matching points on the contour. With more correct matches and fewer mismatches, our approach does not need RANSAC or its variants for post-processing, which are essential for other methods. The second example from the COCO dataset demonstrates our method's superior ability to stably match small target objects.  

 {In Fig.~\ref{fig_align}, we qualitatively compare  our registration results to those of a classic template matching method, Linemod-2D,  and three deep feature matching methods. Linemod-2D  is susceptible to cluttered backgrounds. Learning-based matching baseline methods perform better but are prone to unstable results, especially for small objects.} Our method produces a warped template with more pixels aligned in all these scenarios. Fig.~\ref{zooming} shows that our approach provides much more accurate registration when examined in fine detail.

\begin{table*}[t!]
\caption{Homography estimation on the CoCo dataset. The AUC of the measurement point error  is reported as a percentage.  {SuperGlue$--$ and SuperGlue use the pre-trained SuperPoint and our fine-tuned SuperPoint for keypoint detection, respectively.} 
}
\label{tab:tabel_coco}
\begin{center}
\begin{tabular}{l l r r r}
\hline
\multirow{2}{*}{Category} &
\multirow{2}{*}{Method} & \multicolumn{3}{ c }{Homography est. AUC  $\uparrow$}\\
\cline{3-5} 

& & @3px & @5px & @10px \\
\hline
\multirow{2}{*}{Overall similarity measure}
&Linemod-2D  &26.2  & 47.5 & 64.2 \\
&GHT & 1.8 & 4.5  &10.1 \\

\hline
\multirow{6}{*}{Keypoints+ MNN}
&SURF + RANSAC  & 0.1 & 0.1 &0.3   \\
&SURF + MAGSAC   & 0.1 & 0.2 & 0.8 \\ 

& {D2Net  + RANSAC } & {0.5}& {2.4}& {3.7}  \\
& {D2Net  + MAGSAC } & {1.3}& {3.5}& {7.2}  \\

& {ASLFeat  + RANSAC } & {1.3}& {3.4}& {7.6}  \\
& {ASLFeat  + MAGSAC } & {2.5}& {5.3}& {10.8}  \\

& {SuperPoint  + RANSAC } & {0.1}& {1.4}& {1.2}  \\
& {SuperPoint  + MAGSAC } & {0.5}& {1.8}& {4.4}  \\

\hline
\multirow{8}{*}{Learning matchers}

& {SuperGlue$--$    + RANSAC } & {2.7}& {6.5}& {11.9}  \\
& {SuperGlue$--$ -   + MAGSAC } & {4.8}& {9.4}& {14.6}  \\

& {SuperGlue    + RANSAC } & {14.5}& {21.7}& {31.3}  \\
& {SuperGlue    + MAGSAC } & {14.7}& {22.2}& {32.1}  \\

&COTR  + RANSAC  & 19.1 & 33.5 &47.4  \\
&COTR  + MAGSAC   &22.4  &36.3  & 48.6 \\
&LoFTR  + RANSAC  & 26.9 & 47.2 &62.8 \\
&LoFTR  + MAGSAC   &28.0& 48.5 &64.0 \\

&\textbf{Ours} & \textbf{32.4} & \textbf{51.5} &\textbf{66.2}\\
\hline
\end{tabular}
\end{center}
\end{table*}

\begin{figure}[t!]
\begin{center}
   \includegraphics[width=1.0\linewidth]{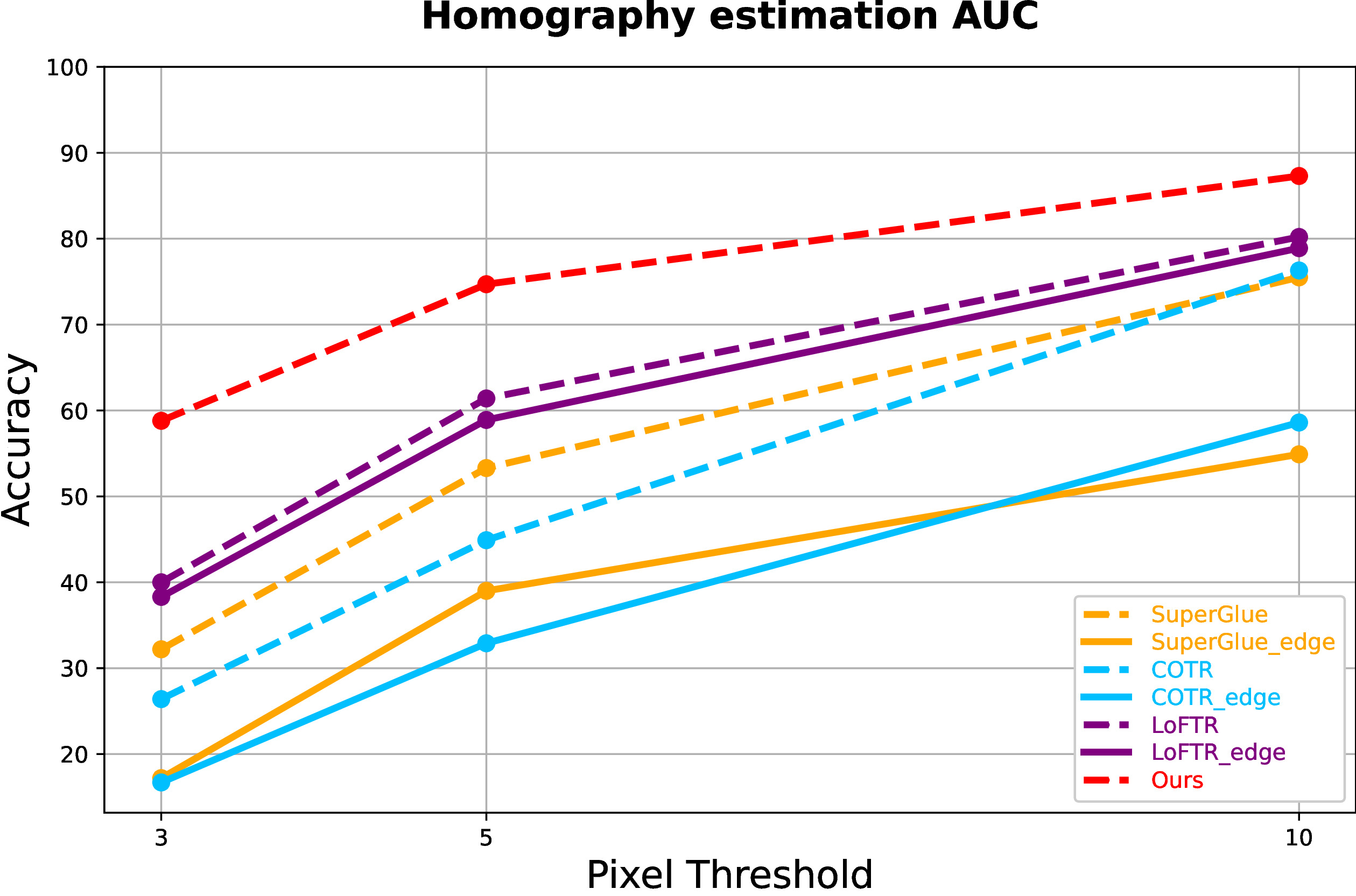}
\end{center}
   \caption{Accuracy of  various methods with (solid lines) and without (dashed lines) edge detection preprocessing of the input images. The homography estimation accuracy is reported for pixel thresholds [3, 5, 10].}
\label{fig:edge_ablation}

\end{figure}

\subsubsection{Baselines using Edge Maps} 
As extracting edges of input images may reduce the impact of modality differences on initial feature extraction, we performed further experiments on the Mechanical Parts dataset to evaluate competitive learning-based baseline methods using edge detection as pre-processing. For a fair comparison, we use PiDiNet to extract edge maps from the template and source images for all methods. Training settings remained the same as for the training process without edge extraction. As  Fig.~\ref{fig:edge_ablation} shows, edge detection preprocessing worsens the results of these baseline methods, especially SuperGlue and COTR. We note that these methods tend to provide correspondences with lower accuracy for low-texture scenes, and edge detection results in images with little texture.

\begin{figure*}[t!]
\begin{center}
   \includegraphics[width=0.9\linewidth]{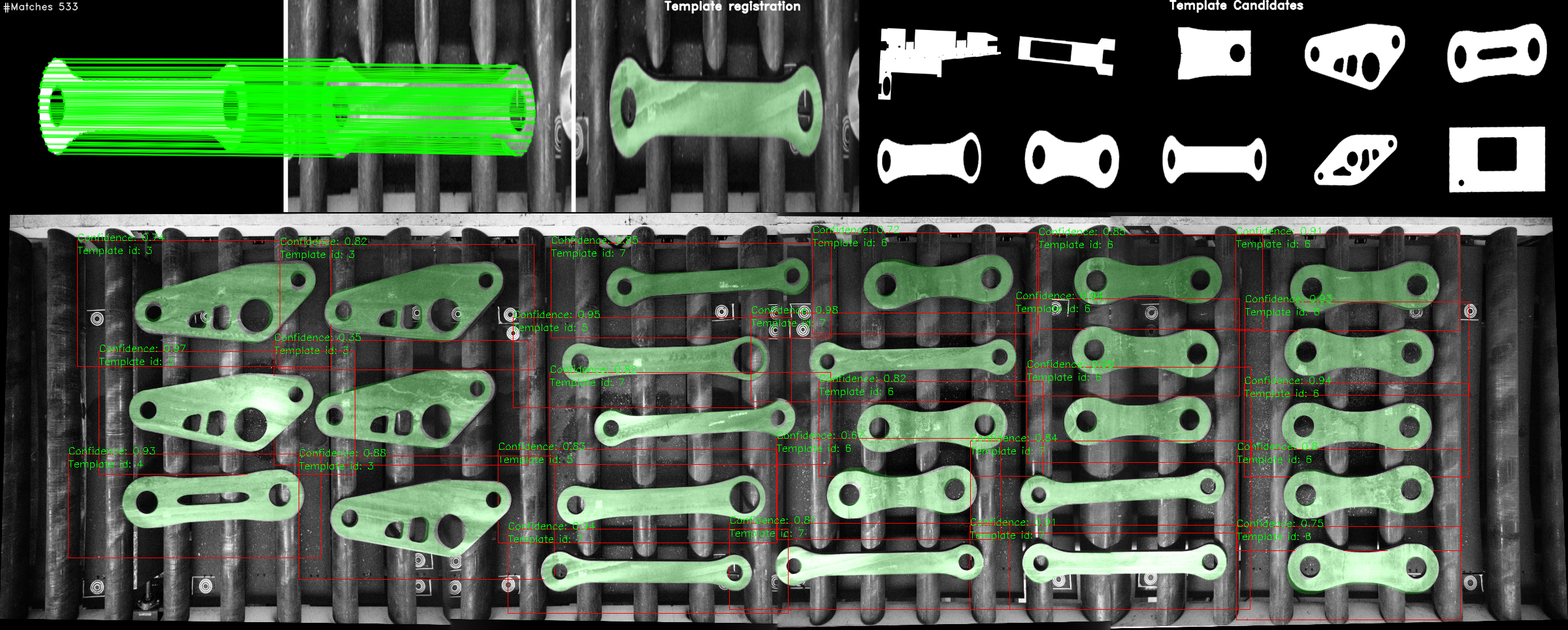}
\end{center}
   \caption{Application of our method in an industrial line. Above left:  best matching template for an object. Above right: set of candidate templates. Below: final matches to selected templates. The coarse matching inlier rate using every template is used as a basis for template selection.}
\label{fig:application}
\end{figure*}

\begin{table*}[t]
\caption{Evaluation of design choices using the Mechanical Parts dataset. Strategies marked with $\star$ are the ones adopted in our method.}
\label{tab:ablation}
\begin{center}
\begin{tabular}{llrrr}
\hline
\multirow{2}{*}{Design aspect} 
 &\multirow{2}{*}{Method} & \multicolumn{3}{ c }{Homography est. AUC  $\uparrow$}\\
\cline{3-5} 
& & @3px & @5px & @10px \\
\hline
\multirow{2}{*}{Matching-layer}
& Dual-Softmax& 58.7 & \textbf{74.7} &\textbf{87.3}\\
& \textbf{Optimal transport} $\star$ &\textbf{58.8} & \textbf{74.7} &\textbf{87.3} \\
\hline
\multirow{2}{*}{Position-encoding}
& Absolute & {53.0} & {70.9} &{85.4} \\
& \textbf{Relative} $\star$ &\textbf{58.8} & \textbf{74.7} &\textbf{87.3}\\
\hline

\multirow{2}{*}{Homography estimation}
& MAGSAC   & 53.8 & 71.3 & 85.6  \\
& \textbf{Consistency} $\star$&\textbf{58.8} & \textbf{74.7} &\textbf{87.3} \\
\hline

\multirow{2}{*}{Value \& position}
& w/o position & 57.9 & 74.1 & 86.9 \\
& \textbf{w position} $\star$ &\textbf{58.8} & \textbf{74.7} &\textbf{87.3}\\
\hline

\multirow{2}{*}{Translation module}
& Canny&57.6  & 73.9 & 86.8  \\
& \textbf{translation network }$\star$ &\textbf{58.8} & \textbf{74.7} &\textbf{87.3} \\
\hline

\multirow{2}{*}{Feature fusion}
& local feature &  51.2& 69.4 & 84.6  \\
& \textbf{local-global feature fusion} $\star$ &\textbf{58.8} & \textbf{74.7} &\textbf{87.3}\\
\hline

\multirow{2}{*}{One stage vs coarse-to-fine}
& one stage & 45.9 & 65.8 & 82.8  \\
& \textbf{coarse-to-fine} $\star$ &\textbf{58.8} & \textbf{74.7} &\textbf{87.3}\\
\hline

\multirow{2}{*}{ {self-supervision loss}}
&  {w/o self-supervision} &  {55.0} &  {72.2} &  {86.0}  \\
& \textbf{ {w self-supervision}} $\star$ & {\textbf{58.8}} &  {\textbf{74.7}} & {\textbf{87.3}}\\
\hline

\end{tabular}

\end{center}
\end{table*}

\begin{table*}[t!]
\caption{Effects on speed and accuracy of varying the number of patches sampled. $\star$ indicates the number used in our method.}
\label{tab:ablation_patch}
\begin{center}
\begin{tabular}{r rrrr}
\hline
\multirow{2}{*}{Number of Patches} 
 & \multicolumn{3}{ c }{Homography est. AUC  $\uparrow$}
 & \multirow{2}{*}{Runtime (ms) $\downarrow$ }
 \\
\cline{2-4} 
& @3px & @5px & @10px &  \\
\hline
 {8} & {22.0} &  {40.9} & {66.0} & {99.3} \\
\hline
 {16} & {37.7} &  {58.0} & {78.6}& {99.8} \\

\hline
 {32} & {50.5} & { 69.0} & {84.2}&  {107.6}\\ 
\hline
 {64} & {57.4} & { 73.7 }& {86.8}&  {112.2} \\
\hline
$\star$ {128}  & {58.8} &  {74.7} & {87.3}& {115.8} \\
\hline
 {256} & {60.0} &  {75.5 }& {87.7}& {125.1} \\
\hline
 {512} & {60.1} &  {75.5} & {87.7}& {175.3} \\
\hline
 {Unsampled} & {52.0} &  {70.2} & {85.0}& {218.4} \\
\hline
\end{tabular}

\end{center}
\end{table*}

\subsection{Application} \label{sec:application}

We now describe a challenging application of our method to real industrial lines, illustrated in Fig.~\ref{fig:application}. For each batch of industrial parts, the task is to select the correct template from a set of candidates templates for each part and to calculate its accurate pose. This is now an $N$-to-$N$ template matching problem. We first pre-process the original scene using a real-time object detection network~\cite{redmon2016you} to roughly locate each part and crop it into a separate image. For each candidate template, we first conduct coarse matching to select the optimal template: we use the correspondences with weighted confidences obtained by coarse matching to get an initial homography. Based on that homography, the template containing the most inlier correspondences is regarded as optimal. We then apply fine matching to accurately  obtain the pose of the object using the selected optimal template.    

To quantitatively evaluate our algorithm in multi-template scenarios, besides the correct template, we randomly add extra 9 noisy ones to the candidate template set. We tested 284 scenarios with 2445 test samples and achieved a recognition accuracy of 98.8\%, when taking an inlier rate of more than 80\% as  correct recognition using the estimation matrix. 
In addition, our method runs at a competitive speed since we adopt the strategy of only using coarse matching for template selection. Further details of runtimes are presented in Appendix~\ref{appx-a}. 

We further note that our model generalizes well to unseen real scenarios  after training only on synthetic data. We provide the link to video demonstrations in the declarations~\ref{Declarations}.

\subsection{Analysis and Discussion} \label{sec 5.5}

\subsubsection{Design study}\label{sec:ablation} 

\begin{figure*}[htb]
 \begin{center}
~~~~~~~~~~~~\qquad\qquad RANSAC~~~~~~~~~~~~~~~~~~~\qquad\qquad\qquad\qquad Consistency \qquad\qquad\qquad\qquad~~~~~~~~~~Ground-truth \qquad\qquad\qquad~~~\\
  \begin{minipage}{0.32\linewidth}
        
     \centerline{\includegraphics[width=\textwidth]{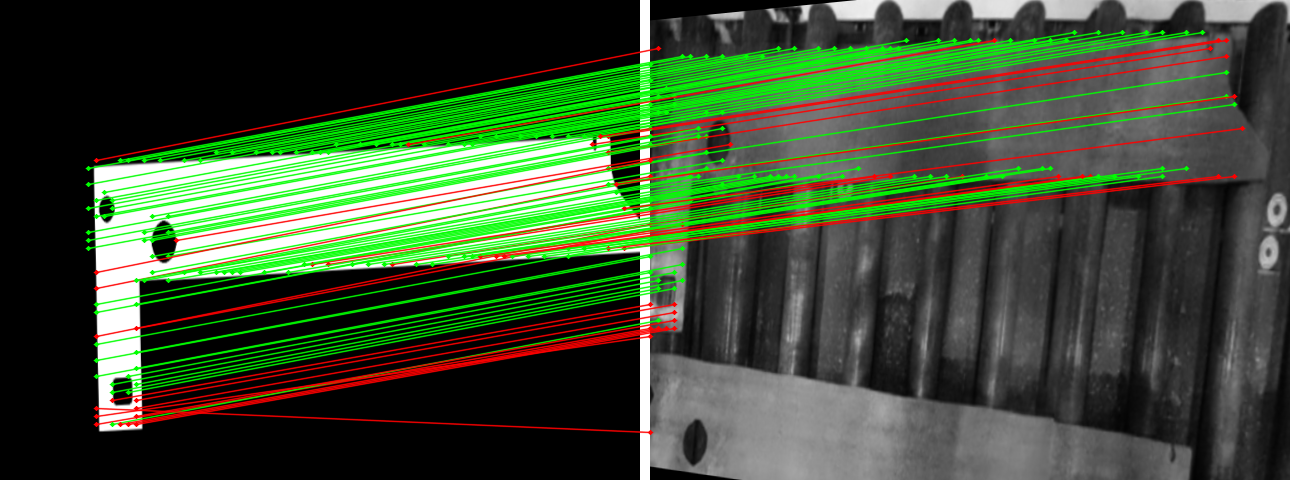}}
     \centerline{\includegraphics[width=\textwidth]{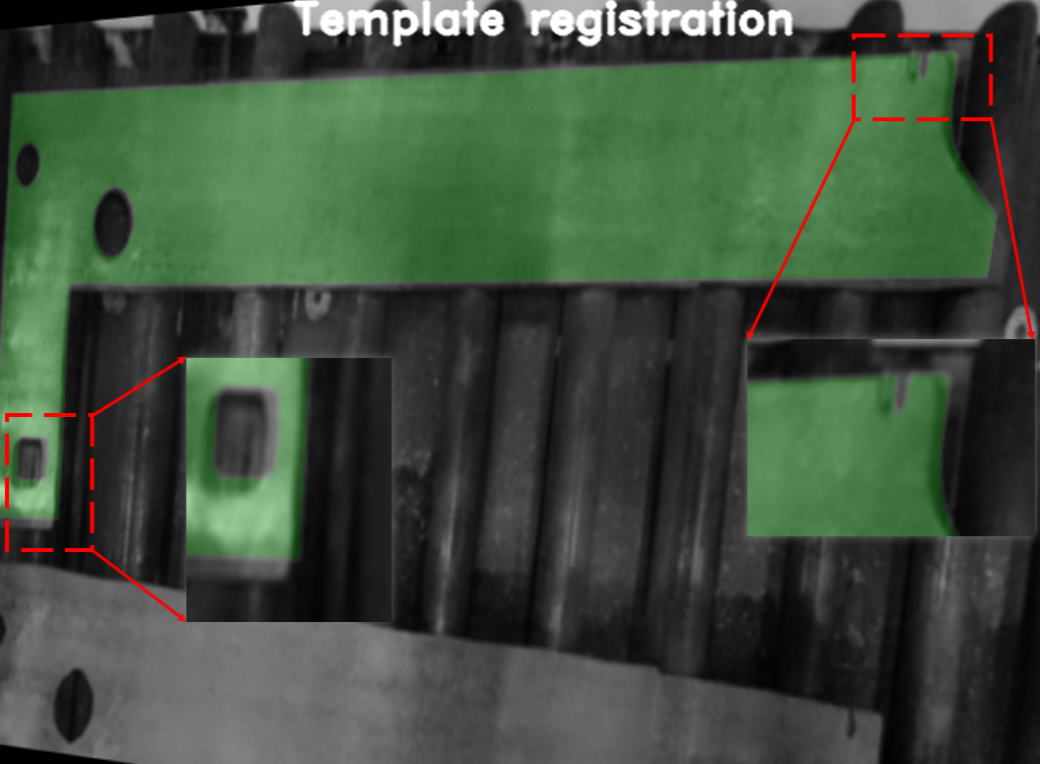}}
       
 \end{minipage}
    \begin{minipage}{0.32\linewidth}
    %
     \centerline{\includegraphics[width=\textwidth]{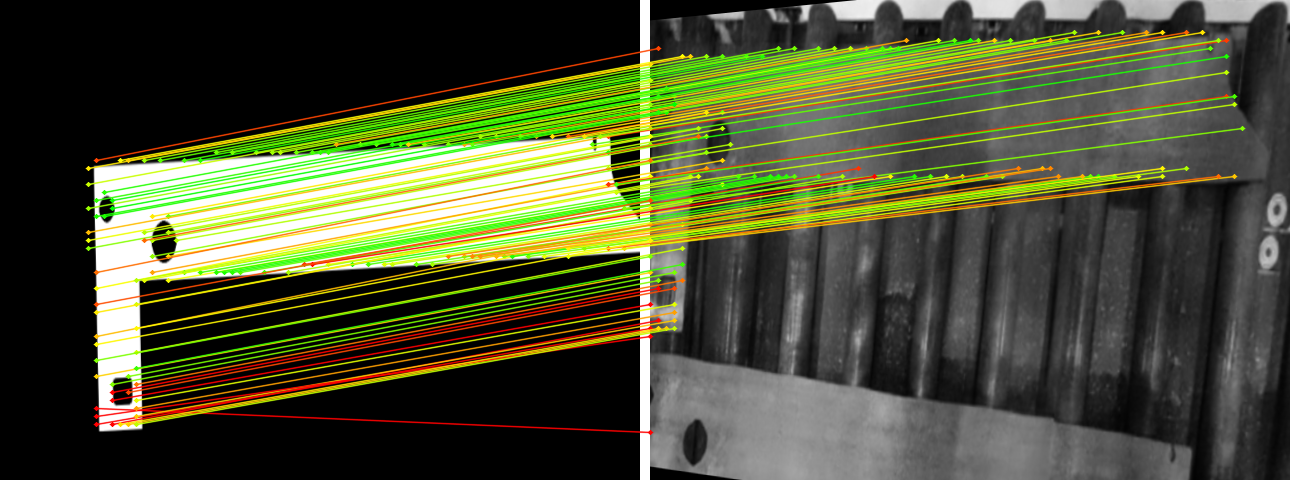}}
     \centerline{\includegraphics[width=\textwidth]{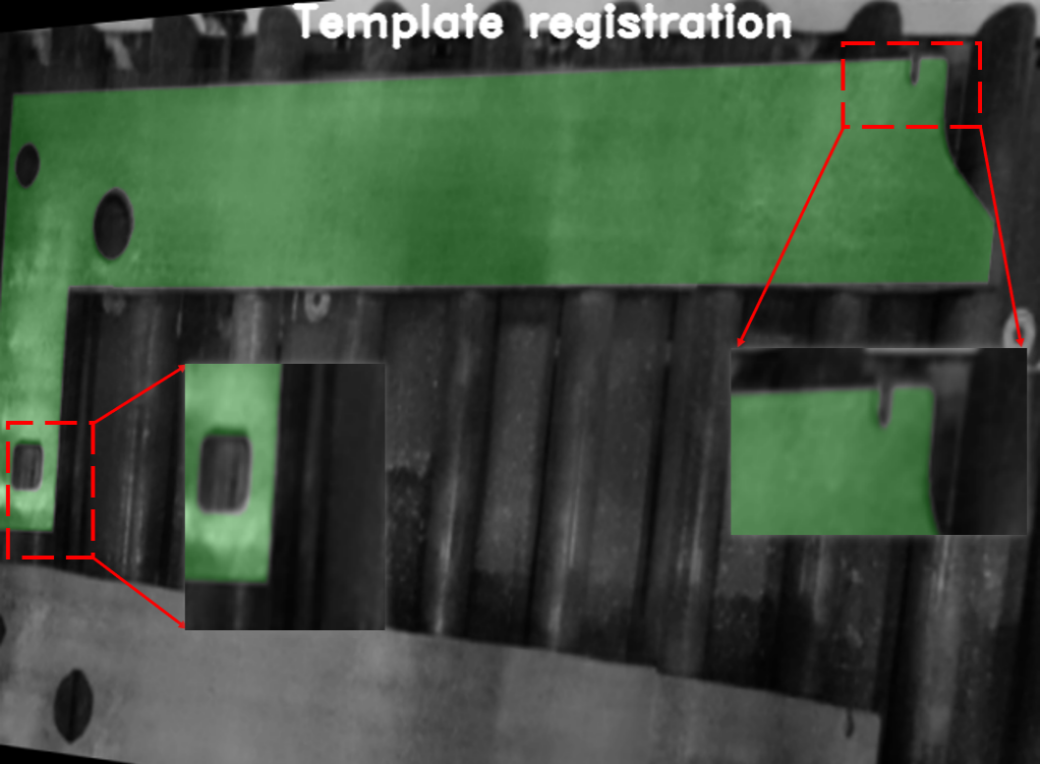}}
       
 \end{minipage}
    \begin{minipage}{0.32\linewidth}
       
     \centerline{\includegraphics[width=\textwidth]{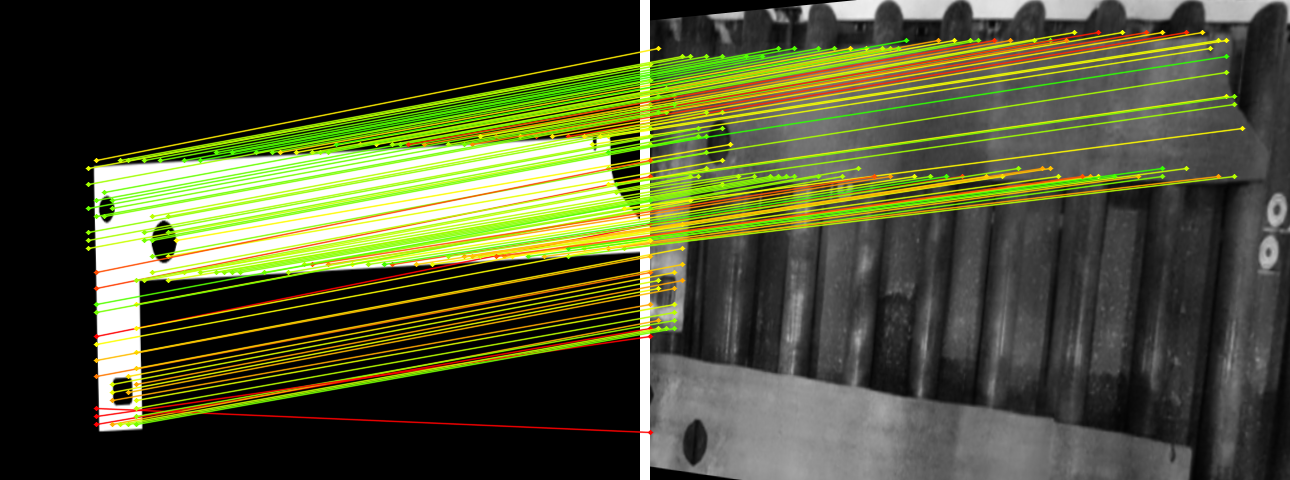}}
     \centerline{\includegraphics[width=\textwidth]{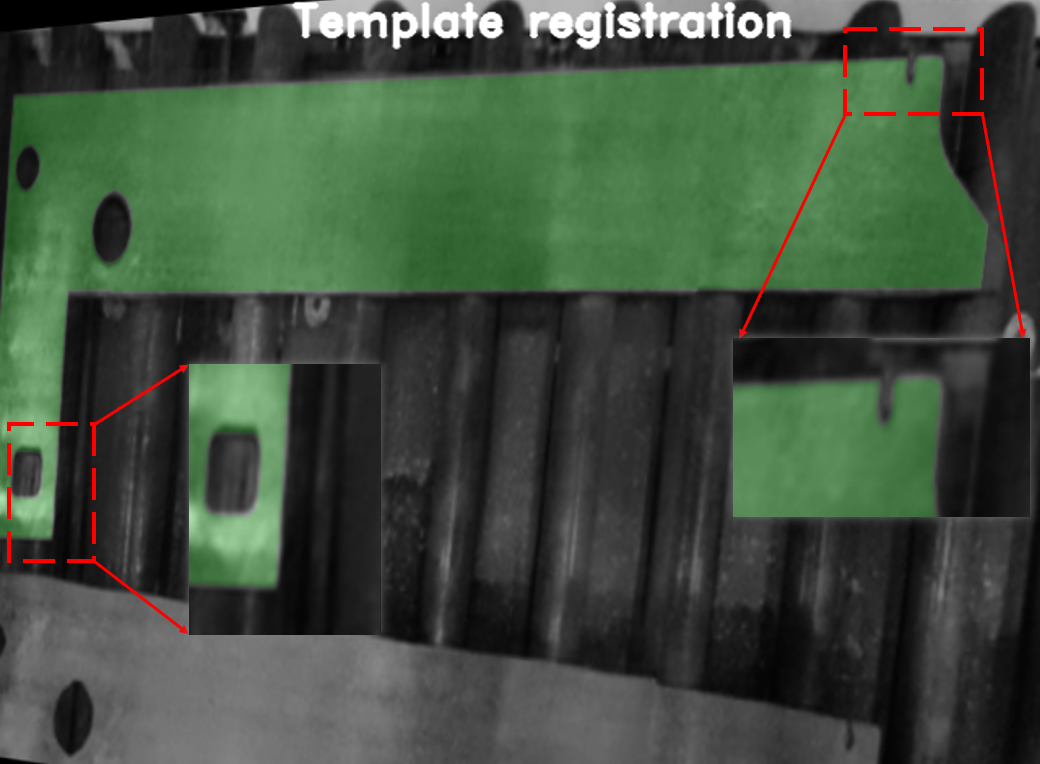}}

 \end{minipage}
 
 \end{center}
	\caption{
	 {Comparison of using  RANSAC,  or  consistency, for homography estimation. Above: correspondences provided by coarse matching. Below: template registration results. Confidence is indicated by line colour from green (1) to red (0.)  In RANSAC, inliers have a  confidence of 1, and outliers, 0.  For the ground-truth, the reprojection error represents confidence. 
	}}
	\label{consistency_plot}
\end{figure*}

To better understand our proposed method, we conducted seven comparative experiments on different modules, using the Mechanical Parts dataset. {The quantitative results in Tabs.~\ref{tab:ablation} and~\ref{tab:ablation_patch} validate our design decisions and show that they have a significant effect on performance.} 
The choices considered are 
\begin{itemize}
\item \emph{Matching-layer: dual-softmax vs. optimal transport} Both the dual-softmax operator and optimal transport achieve similar scores, and either would provide effective matching layers in our method. 
\item \emph{Position-encoding: absolute vs. relative} Replacing  relative positional encoding by absolute positional encoding results in a significant drop in AUC. Relative position information is important in template matching. 
\item \emph{Homography estimation: RANSAC vs. consistency confidence} Since our method provides high-quality correspondences with confidence weights based on consistency, the differentiable DLT outperforms MAGSAC. An example is shown to demonstrate the advantages of DLT with consistency weights over RANSAC in Fig.~\ref{consistency_plot}. Inliers and outliers are explicitly distinguished by the RANSAC estimator, so correspondences with insufficient accuracy are directly discarded or fully adopted to estimate the final transformation matrix. Instead, our consistency module  provides confidence weights, and we observe that the confidence weights estimated by the proposed method are consistent with ground-truth reprojection errors. Our method  effectively assigns higher weights to  more accurate correspondences and suppresses outliers. Therefore, in the case of high-quality matches, our consistency module can efficiently utilize correspondence information and so outperforms RANSAC.
\item \emph{Value \& position} Multiplying the value token by the positional embedding in the transformer module provides better results. 
\item \emph{Translation module: Canny~\cite{canny1986computational} vs. translation network} Accuracy using the translation network is better than using Canny edge detection. 
\item\emph{Feature fusion} In the refinement stage, deep fusion of local features and global features leads to a noticeable performance improvement. 
\item\emph{One stage vs. coarse-to-fine} The coarse-to-fine module contributes to the estimation accuracy significantly by finding more matches and refining them to a sub-pixel level.  
\item\emph{Self-supervision loss} Using self-supervision loss ($L2$ similarity loss) brings a significant performance boost in fine-level training.
\item\emph{Maximum number of sample patches} See Tab.~\ref{tab:ablation_patch}. As the maximum number of samples based on the contour increases, accuracy of our method tends to improve. However, without sampling and using the entire template image  as input,  performance is somewhat lower than achieved  by sampling with the best number of patches. We believe that edge-based sampling allows our method to more efficiently perceive the template structure and aggregate local features. We set the maximum number of patches to 128 as a trade-off between accuracy and runtime.
\end{itemize}

\subsubsection{Understanding Attention}
\begin{figure*}[t!]
 \begin{center}

\begin{minipage}{0.02\linewidth}
    \rotatebox{90}{{
~~~~~~~~~~~~~~~~Features
}}

    \rotatebox{90}{{
~~~~~~Self~~~~~
}}
        \rotatebox{90}{{
~~~~Cross~~~~~~~~~~~~~~~~~~~~
}}

    \end{minipage}
  \begin{minipage}{0.48\linewidth}
     \centerline{\includegraphics[width=\textwidth]{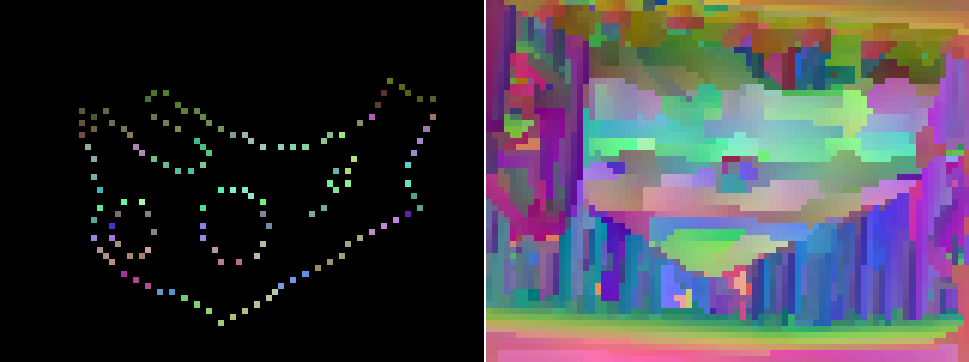}}
        
     \centerline{\includegraphics[width=\textwidth]{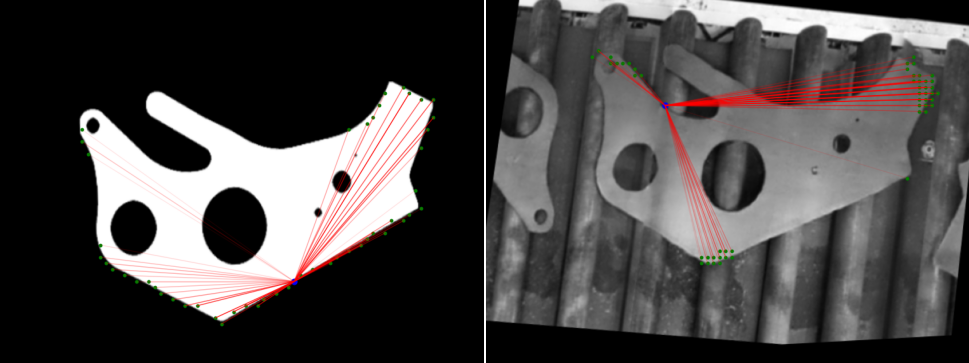}}
        
     \centerline{\includegraphics[width=\textwidth]{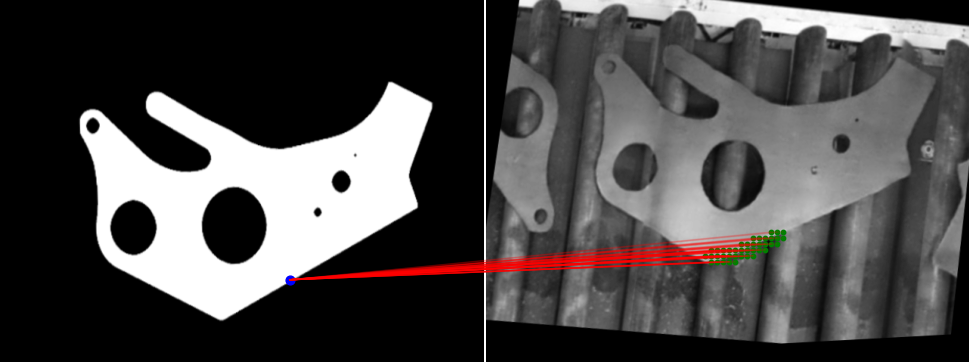}}

 \end{minipage}
   \begin{minipage}{0.48\linewidth}
         \centerline{\includegraphics[width=\textwidth]{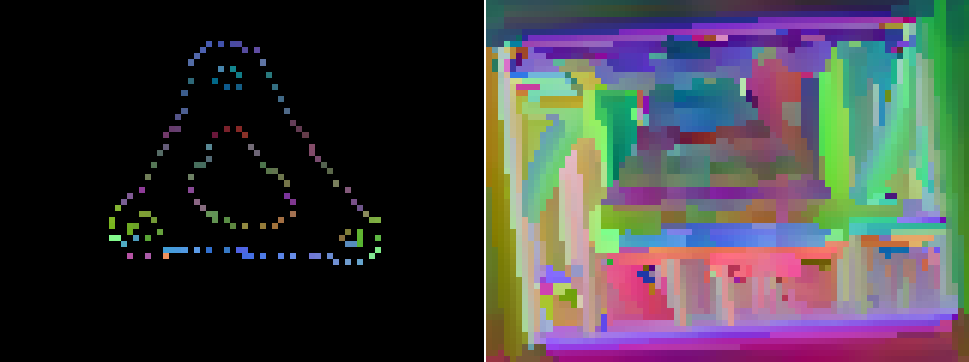}}
        
     \centerline{\includegraphics[width=\textwidth]{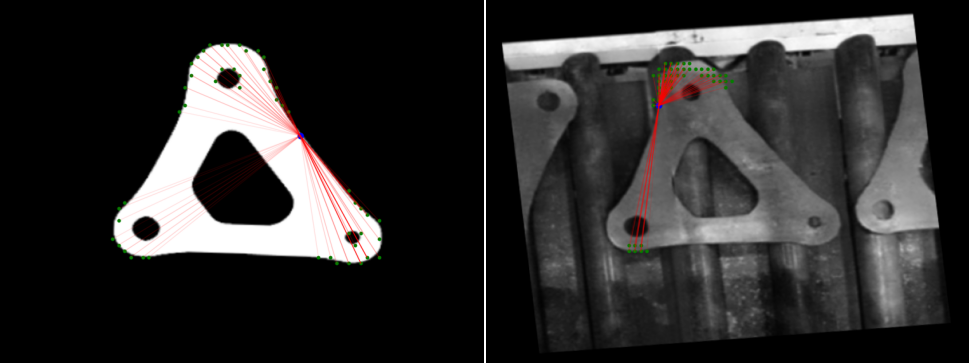}}
        
     \centerline{\includegraphics[width=\textwidth]{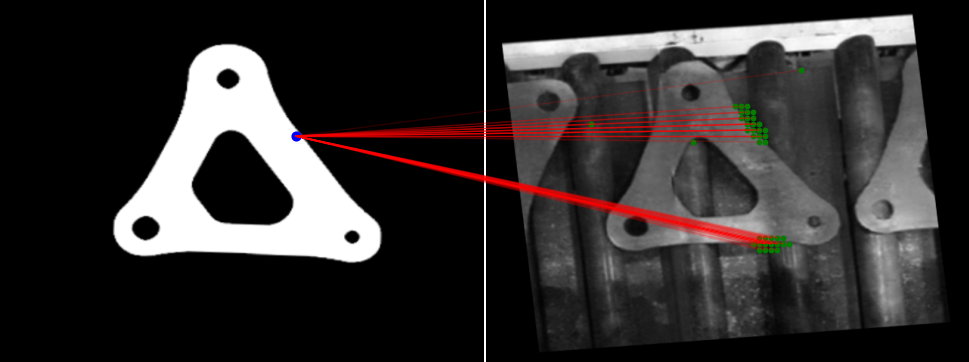}}
 \end{minipage}

 \end{center}
	\caption{Transformed features using t-SNE, and self- and cross-attention weights during coarse matching.}
	\label{weight_vis}
\end{figure*}

To better understand the role of attention in our method, we visualize  transformed features with  t-SNE~\cite{van2008visualizing}, and self- and cross-attention weights in Fig.~\ref{weight_vis}. The visualization shows our method learns a position-aware feature representation. The visualized attention weights  reveals that the query point can aggregate global information dynamically and focus on meaningful locations.  Self-attention may focus  anywhere in the same image, especially regions with obvious differences, while cross-attention focuses on regions with a similar appearance in the other image.

\subsubsection{Limitations and Future Work} 
Our method utilizes an existing edge detection network to eliminate the domain gap between templates and images, which is convenient for our approach. However, we believe that jointly training the translation network is a promising avenue for further improving performance. Another interesting follow-up is to
design a one-to-many template matching algorithm that does not rely on any pre-processing.

\section{Conclusions}
We have presented a differentiable pipeline for accurate correspondence refinement for industrial template matching. With efficient feature extraction and feature aggregation by transformers, we obtain high-quality feature correspondences between the template mask and the grayscale image in a coarse-to-fine manner. The correspondences are then used to get a precise pose or transformation for the target object. To eliminate the domain gap between the template mask and grayscale image, we exploit a translation network. Based on the properties of the cross-modal template matching problem, we design a structure-aware strategy to improve robustness and efficiency. Furthermore, two valuable datasets from industrial scenarios have been collected, which we expect to benefit future work on industrial template matching. Our experiments show that our method significantly improves the accuracy and robustness of template matching relative to multiple state-of-the-art methods and baselines. Video demos of $N$-to-$N$ template matching in real industrial lines show the effectiveness and good generalization of our method.

\section{Declarations}
\subsection{Availability of data and materials}\label{Declarations}
The well-known CoCo dataset is available from \url{https://cocodataset.org/}. Our two industrial datasets can be freely downloaded from \url{https://drive.google.com/drive/folders/1Mu9QdnM5WsLccFp0Ygf7ES7mLV-64wRL?usp=sharing}. The video demos are available at the page: \url{https://github.com/zhirui-gao/Deep-Template-Matching}.

\subsection{Competing interests}
The authors declare that they have no known competing
financial interests or personal relationships that could have appeared to influence the work reported in this paper. 

\subsection{Funding}
This paper was supported in part by the National Key Research
and Development Program of China (2018AAA0102200), the National Natural Science Foundation of China (62132021, 62002375, 62002376), the Natural Science Foundation of Hunan Province of China (2021JJ40696), Huxiang Youth Talent Support Program (2021RC3071,2022RC1104) and NUDT Research Grants (ZK19-30, ZK22-52). 

\subsection{Authors’ contributions}
\textbf{Zhirui Gao:} Methodology, Writing Draft, Visualization, Results Analysis; \textbf{Renjiao Yi:} Methodology, Supervision, Writing Draft, Results Analysis; \textbf{Zheng Qin:} Supervision, Results Analysis; \textbf{Yunfan Ye:} Supervision, Results Analysis; \textbf{Chenyang Zhu:} Methodology, Supervision; \textbf{Kai Xu:}
Methodology, Supervision.

\subsection{Acknowledgements}
We thank Lintao Zheng and Jun Li for their help with dataset preparation and discussions. 

\appendix
\section{Appendices}
\subsection{Speed} \label{appx-a}
We have tested the runtime of our method and other baselines on the Assembly Holes dataset, and report average values using an NVIDIA RTX 3080Ti. Coarse matching in our method takes 63 ms to match one pair; full matching takes 105 ms. LoFTR  takes 87 ms, while COTR  is much slower at 17 s. GHT and Linemod-2D  take  4.9 s and 2.2 s respectively: using multiple templates for different scales and poses is time-consuming. For a scene with 10 objects and 10 candidate templates, our method takes about 6.7 s to locate and identify all objects, and provide accurate poses.  

\subsection{Training Details}\label{appx-b}
Our network was trained on 2 NVIDIA RTX 3090 GPUs using a batch size of 16. Although end-to-end training is feasible, we found that a two-stage training strategy yielded better results. The first stage trained coarse-level matching using the loss term $\mathcal{L}_c$,  until the validation loss converged. The second stage trained the whole pipeline using both $\mathcal{L}_c$ and $\mathcal{L}_f$ until the validation loss converged. Using the Mechanical Parts / Assembly Holes / COCO datasets, we trained our network for 15 / 15 / 30 epochs respectively for the first stage using Adam, with an initial rate of $10^{-3}$, and 18 / 15 / 12 epochs for the second stage using Adam, with an initial rate of $10^{-4}$. We loaded pre-trained weights for the translation network and local feature CNN provided by~\cite{su2021pixel,detone2018superpoint}, and fixed the local feature CNN parameters in the second stage.  

 {We also loaded pre-trained parameters for other learning-based baseline methods and retrained them until the validation loss converged. Numbers of training epochs used for the different learning-based baseline methods  are shown in Tab.~\ref{tab:train_epoch} for each dataset. For better performance, for the keypoints methods (D2Net, ASLFeat  and SuperPoint), we only used the edge points on the template to construct the ground-truth matching pairs when training the network. For COTR, we followed its three-stage training strategy to fine-tune the network. Since there is no recommended training method for SuperGlue, we trained it based on the code at \url{https://github.com/gouthamvgk/SuperGlue_training}.}

\begin{table}[t!]
\caption{Number of training epochs used for different learning-based baseline methods, for the three datasets Mechanical Parts (MP),  Assembly Holes (AH) and COCO. For SuperGlue$--$ and SuperGlue, we respectively used the pre-trained SuperPoint and our fine-tuned SuperPoint in the keypoint detection phase. }
\label{tab:train_epoch}
\begin{center}
\begin{tabular}{llrrr}
\hline
{ Category}                    & { Method}    & { MP} & { AH} & { COCO}  \\ \hline
{ }                            & { D2Net }      & { 20}               & { 20}             & { 20}    \\
{ }                            & { ASLFeat }    & { 15}               & { 15}             & { 15}    \\
\multirow{-3}{*}{{ Keypoint}} & { SuperPoint } & { 10}               & { 10}             & { 10}    \\ \hline
{ }                            & { SuperGlue$--$  }  & { 30}               & { 30}             & { 30}    \\
{ }                            & { SuperGlue  }  & { 10 + 30}            & { 10 + 30}          & { 10 + 30} \\
{ }                            & { COTR }       & { 100}              & { 100}            & { 90}    \\
\multirow{-4}{*}{{ Matching}}  & { LoFTR }      & { 12}               & { 12}             & { 35}    \\ \hline
\end{tabular}
\end{center}
\end{table}

\subsection{Data and Ground-truth Generation}\label{appx-c}
\subsubsection{Mechanical Parts Dataset} 
We used GauGAN to generate further image pairs for various shape parts. The manually adjusted image pairs provided by Linemod-2D  served as training data for GauGAN. We discovered that GauGAN can learn well even from somewhat noisy data, and can provide high-quality ground-truth. Using arbitrarily distributed mask images (templates), we used  GauGAN to generate synthetic industrial part images of size $480 \times 640$ for our network training. For each pair of generated images, we first moved the template mask to the center of the image, then randomly scaled the synthetic image by a factor in the range [0.8, 1.2] and rotated it through an angle in the range [-15$^{\circ}$, 15$^{\circ}$].

\subsubsection{Assembly Holes Dataset} 
The ground-truth of the Assembly Holes dataset was annotated by humans: we segmented the outer circle of the part to give the mask for the image. The scaling range was [0.75, 1.25].

\subsubsection{COCO Dataset} 
We used the instance mask as the template $T$ and the original image as the search image $I$ for the COCO dataset~\cite{lin2014microsoft}. We first filtered out  masks near the image boundary because these masks tend to be the
image background. Among the remaining masks, we chose the mask with the largest area as the template. The selected image pairs were resized to $480\times 640$, the scaling range set to [0.9, 1.1] and the rotation range to [-30$^{\circ}$, 30$^{\circ}$].  

\subsubsection{All Datasets}
To simulate possible object deformation or camera disturbance, we randomly perturbed the four corners of the image by values within the range [-32, 32] pixels for all datasets.

\bibliographystyle{CVMbib}
\bibliography{refs}

\subsection*{Author Biographies}

\begin{biography}
[gaozhirui_w]{Zhirui Gao} received his B.E. degree in Computer Science and Technology from the Chinese University of Geosciences, Wuhan in 2021. He is now a master student at the National University of Defense
Technology, China (NUDT). His research interests include image matching and 3D vision.
\end{biography}
\begin{biography}
[renjiaoYi]{Renjiao Yi} is an Assistant Professor in the
School of Computing, NUDT. She is interested in 3D
vision problems such as inverse rendering
and image-based relighting.
\end{biography}
\vspace*{1.9em}
\begin{biography}
[qinzheng]{Zheng Qin} received B.E. and M.E. degrees in Computer Science and Technology from NUDT in 2016 and 2018, respectively, where he is currently pursuing a Ph.D. degree. His research interests focus on 3D vision, including point cloud registration, pose estimation, and 3D representation learning.
\end{biography}
\vspace*{0.8em}
\begin{biography}
[yunfa]{Yunfan Ye} is a Ph.D. candidate in the
School of Computing, NUDT. His research interests include computer vision and graphics.
\end{biography}
\vspace*{0.8em}
\begin{biography}
[chenyangzhu]{Chenyang Zhu} is an Assistant Professor in the
School of Computing, NUDT. His current directions of interest include data-driven shape analysis and modeling, 3D vision, robot perception and robot navigation.
\end{biography}
\vspace*{0.8em}
\begin{biography}
[kaixu]{Kai Xu} is a Professor in the
School of Computing, NUDT, where he received his Ph.D. in 2011.
He serves on the editorial board of ACM
Transactions on Graphics, Computer Graphics Forum, Computers \& Graphics, etc.
\end{biography}
\vspace*{0.8em}

\subsection*{Graphical abstract}
\begin{figure}[H]
\begin{center}
   \includegraphics[width=1.0\linewidth]{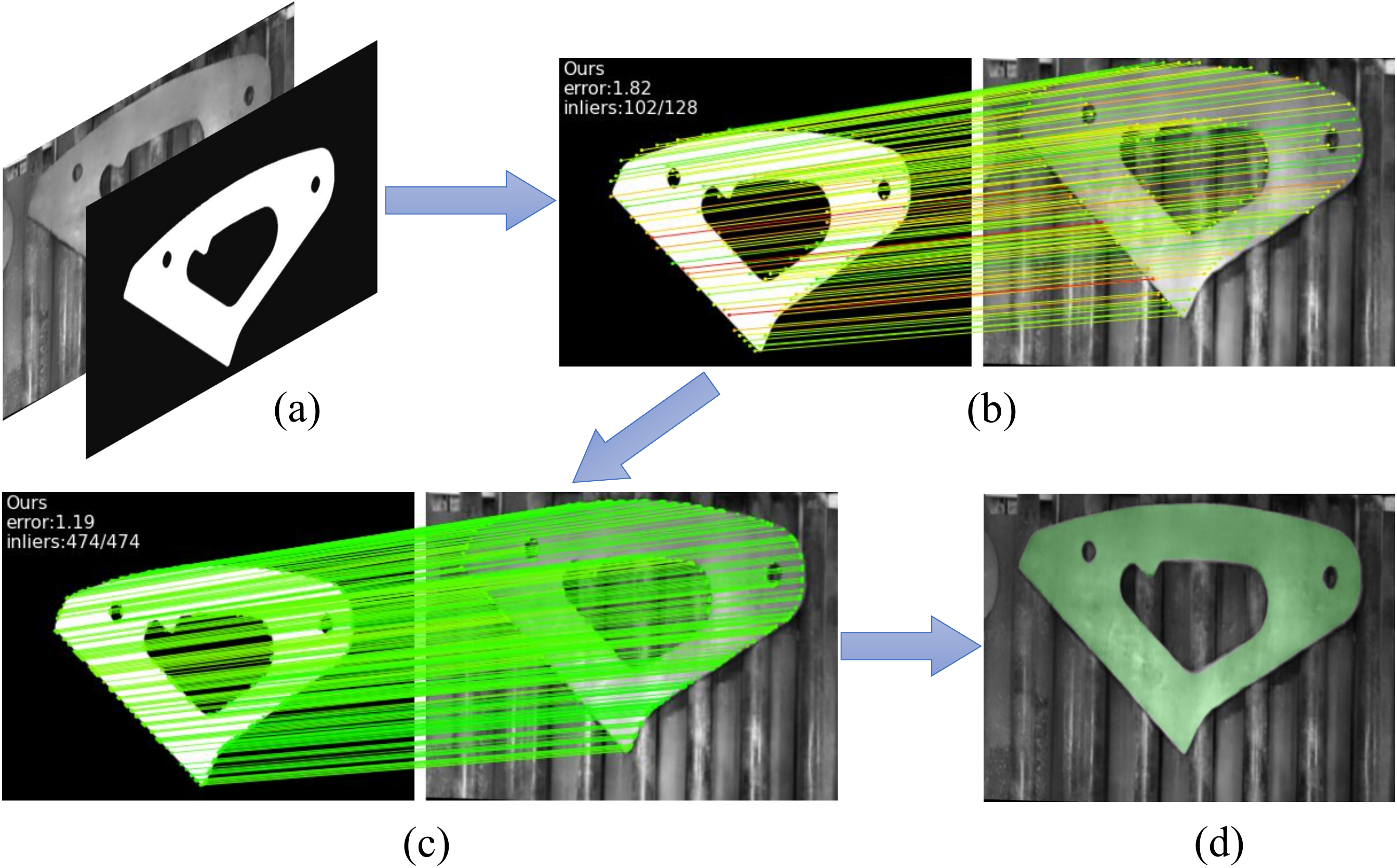}
\end{center}
  
\label{fig:graphical abstract}
\end{figure}

\end{document}